\documentclass[twoside,10pt]{article}

\usepackage{jmlr2e}

\jmlrheading{1}{2000}{1-48}{4/00}{10/00}{Chuanhao Li and Hongning Wang}

\ShortHeadings{Asynchronous Algorithms for Federated Linear Bandit}{Li and Wang}
\firstpageno{1}

\usepackage[utf8]{inputenc} % allow utf-8 input
\usepackage[T1]{fontenc}    % use 8-bit T1 fonts
\usepackage{hyperref}       % hyperlinks
\usepackage{url}            % simple URL typesetting
\usepackage{booktabs}       % professional-quality tables
\usepackage{amsfonts}       % blackboard math symbols
\usepackage{nicefrac}       % compact symbols for 1/2, etc.
\usepackage{microtype}      % microtypography
\usepackage{graphicx}
\usepackage{algorithmic}
\usepackage{algorithm}
\usepackage{wrapfig}

\usepackage{amsmath}
\usepackage{subfigure}
\usepackage{multirow}
\usepackage[section]{placeins}
\usepackage{enumitem}

\newcommand{\modelone}{{Async-LinUCB}}
\newcommand{\modeltwo}{{Async-LinUCB-AM}}
\newcommand{\modelbaseline}{{Sync-LinUCB}}
\newtheorem{assumption}{Assumption}
\DeclareMathOperator*{\argmax}{arg\,max}
\DeclareMathOperator*{\argmin}{arg\,min}
\def \cU {\mathcal{U}}
\def \cD {\mathcal{D}}
\def \cN {\mathcal{N}}

\def \cF {\mathcal{F}}

\def \bx {\mathbf{x}}
\def \bX {\mathbf{X}}

\def \by {\mathbf{y}}

\def \cA {\mathcal{A}}
\def \cI {\mathcal{I}}

\def \bR {\mathbb{R}}
\def \bE {\mathbb{E}}

\def \cE {\mathcal{E}}
\def \cS {\mathcal{S}}

\begin{document}

\title{Asynchronous Upper Confidence Bound Algorithms for Federated Linear Bandits}

\author{\name Chuanhao Li \email cl5ev@virginia.edu \\
       \addr Department of Computer Science\\
       University of Virginia\\
       Charlottesville, VA 22903, USA\\
       \AND
       \name Hongning Wang \email hw5x@virginia.edu \\
       \addr Department of Computer Science\\
       University of Virginia\\
       Charlottesville, VA 22903, USA
       }

% \editor{Leslie Pack Kaelbling}

\maketitle

\begin{abstract}%   <- trailing '%' for backward compatibility of .sty file
Linear contextual bandit is a popular online learning problem. It has been mostly studied in centralized learning settings. With the surging demand of large-scale decentralized model learning, e.g., federated learning, how to retain regret minimization while reducing communication cost becomes an open challenge. 
In this paper, we study linear contextual bandit in a federated learning setting. We propose a general framework with asynchronous model update and communication for a collection of homogeneous clients and heterogeneous clients, respectively. Rigorous theoretical analysis is provided about the regret and communication cost under this distributed learning framework; and extensive empirical evaluations demonstrate the effectiveness of our solution.
\end{abstract}


\section{Introduction}\label{sec:intro}
As a popular online learning problem, linear contextual bandit has been used for a variety of applications, including recommender systems \citep{li2010contextual}, display advertisement \citep{li2010exploitation} and clinical trials \citep{durand2018contextual}. While most existing solutions are designed under a centralized setting (i.e., data is readily available at a central server), in response to the increasing application scale and public concerns of privacy, there is a growing demand to keep data decentralized and push the learning of bandit models to the client side.
% As a classic sequential decision making problem, linear contextual bandit has been widely used for a variety of real-world applications, including recommender systems \citep{li2010contextual}, display advertisement \citep{li2010exploitation} and clinical trials \citep{durand2018contextual}. 
% Most existing solutions are designed under a centralized learning setting, i.e., data is readily available at a central server. However, with the increasing public concerns of privacy, especially the bandit algorithms usually directly learn from user data,
% there is a growing demand to keep data decentralized and push the learning of bandit models to the client side. 
Federated learning has recently emerged as a promising setting for decentralized machine learning \citep{konevcny2016federated}.
% , and its effectiveness was first validated at a large scale by training a global model across all mobile devices via the Google Keyboard Android application \cite{konevcny2016federated}. 
%The term ``federated learning" was first introduced by \citet{mcmahan2017communication} with an emphasis on efficiently training deep models over mobile device applications. As significant amount of later works have applied federated learning to other applications, there may be variations in its meaning for different research communities. 
Since its debut in \citeyear{mcmahan2017communication}, there have been many variations for different applications \citep{yang2019federated}. However, most existing works study offline supervised learning problems \citep{li2019convergence,zhao2018federated}, which only concerns optimization convergence over a fixed dataset. How to perform federated bandit learning remains under-explored, and is the main focus of this paper. 

Analogous to its offline counterpart, the goal of federated bandit learning is to minimize the cumulative regret incurred by $N$ clients during their online interactions with the environment over time horizon $T$,
% $N$ clients in a learning system need to collaborate to minimize the overall cumulative regret over a finite time horizon $T$, 
while keeping each client's raw data local. Take recommender systems as an example, where the clients correspond to the edge devices that directly interact with user by making recommendations and receiving feedbacks. Unlike centralized setting where observations from all clients are immediately transmitted to the server to learn a single model, in federated bandit learning, each client makes recommendations based on its local model, with occasional communication for collaborative model estimation.

% In this paper, we follow the general definition by \citet{kairouz2019advances}: multiple clients collaborate in solving a machine learning problem under the coordination of a central server, while keeping each client's raw data local. 

%Though having potential for wide range of applications, online learning problems like linear bandit in federated learning setting, a.k.a. federated linear bandits \cite{dubey2020differentially}, have not attracted enough attention and still remain an open problem. 

% Therefore, it is a natural idea to study contextual linear bandit in a federated learning paradigm, which is also referred to as federated linear bandits \cite{dubey2020differentially}. In a federated learning paradigm, multiple clients collaborate in solving a machine learning problem, under the coordination of a central server, and each client's raw data is stored locally and not transferred to the server. 
% when linear bandit algorithms are applied to the federated learning paradigm, because these algorithms assume a traditional centralized machine learning system where all the data are collected together and all the computation happens in one machine or data center. 
Several new challenges arise in this problem setting. 
The first is the conflict between the need of timely data/model aggregation for \emph{regret minimization} and the need of \emph{communication efficiency}, since communication is the main bottleneck for many distributed application scenarios, e.g., communication in a network of mobile devices can be slower than local computation by several orders of magnitude \citep{huang2013depth}. A well-designed communication strategy becomes vital to strike the balance. 
In addition, 
% constraints from real-world applications should also be taken into consideration when designing the communication strategy. For example, 
the clients often have various response time and even occasional unavailability in reality, due to the differences in their computational and communication capacities.
% the clients may differ in their computational and communication capacities. This will lead to various response time and even occasional unavailability. 
This hampers global synchronization employed in existing federated bandit solutions \citep{wang2019distributed,dubey2020differentially}, which requires the server to first send a synchronization signal to all clients, wait and collect their returned local updates, and finally send the aggregated update back to every client.
Second, it is very restrictive to only assume homogeneous clients, i.e., they solve the same learning problem. 
% As bandit algorithms are mostly deployed to interact with individual users, studying heterogeneous clients with personalized learning problems has a greater potential.
Studying \emph{heterogeneous clients} with distinct learning problems has a greater potential in practice.
This is referred to as ``\emph{non-IIDness}" of data in the context of federated learning, e.g., the difference in $\mathcal{P}_{i}(\bx,y)=\mathcal{P}_{i}(\bx) \mathcal{P}_{i}(y|\bx)$ is caused by each client $i\in[N]$ serving a particular user or group of users, a particular geographic region, or a particular time period. Apparently, it is also unreasonable to assume every client has equal amount of new observations, which however is assumed in existing works. 

%To be more concrete, due to the time-varying arm set $\cA_{t}$ and the dependence on history data for arm selection in linear bandit, context vector $X$ is non-IID in nature and is not the main concern. 
% It is not a major concern since the performance metric, i.e. regret $r_{t}$, is defined against the best arm in $\cA_{t}$. 

% For example, internet connection and the different computation power of devices.
% \textcolor{red}{reasons we need async algo}

% This naturally leads to the question: how to balance between regret minimization and communication efficiency in the federated linear bandit problem.
To address the first challenge, we propose an asynchronous event-triggered communication framework for federated linear bandit. 
%Our event-triggering mechanism offers a flexible way to balance between the regret-minimization and communication-efficiency dilemma. 
Communication with a client happens only when the last communicated update to the client becomes irrelevant to the latest one; and we prove only by then effective regret reduction can be expected in this client because of the communication. 
Under this asynchronous communication, each client sends local update to and receives aggregated update from the server independently from other clients, with no need for global synchronization. This improves our method's robustness against possible delays and temporary unavailability of clients. It also brings in reduced communication cost when the clients have distinct availability of new observations, because global synchronization requires every client in the learning system to send its local update despite the fact that some clients can have very few new observations since last synchronization.
% make the proposed method more robust and practical against the infrastructure constraints, because the aggregated update sent to each client is asynchronous and  
% This makes our method more robust against possible delays in the communication, and we prove that the client enjoys the same benefit in regret reduction as long as it receives the update before its next interaction with the environment.

To address the second challenge, we design algorithms for federated linear bandit with both ``\emph{IIDness}" and ``\emph{non-IIDness}" based on the proposed communication framework. We consider two different assumptions on the reward functions. First, all the clients share a common reward function i.e., a single model is learned for all clients. Second, each client has a distinct reward function with mutual dependence captured by globally shared components in the unknown parameter, which resembles 
%so one model per client is learned during the interaction with the environment, which in essence is similar to the problem considered in
federated multi-task learning \citep{smith2017federated}.
We rigorously prove the upper bounds of accumulative regret and communication cost for the proposed algorithms in these two settings, and conduct extensive empirical evaluations to demonstrate the effectiveness of our proposed framework.
% especially its flexibility in balancing the trade-off between regret and communication cost.
\section{Related Works}
%Conventionally, the contextual linear bandit problem only involves one learning agent, and 
Most existing bandit solutions assume a centralized learning setting, where data is readily available at a central server. Classical linear bandit algorithms, like LinUCB \citep{li2010contextual,abbasi2011improved} and LinTS \citep{agrawal2013thompson,abeille2017linear} %attain $O(\sqrt{T}\log{T})$ regret upper bound, which matches the $\Omega(\sqrt{T})$ minimax lower bound (up to a logarithmic factor) \citep{lattimore2020bandit}. But they 
only concern a single learning agent.  
Multi-agent bandits mostly focus on customizing algorithms that leverage relationships among the agents for collaborative learning \citep{cesa2013gang,wang2019distributed,gentile2014online,cesa2013gang,wu2016contextual,li2021unifying}, but the data about all agents is still on the central server.
%For example, graph Laplacian is used to capture known agent dependency to regularize each agent's model estimation \cite{cesa2013gang,yang2020laplacian}; and  agents' learned models are clustered for observation sharing \cite{gentile2014online,li2016collaborative,gentile2017context,li2019improved}. However, 
%For example, the series of works following CLUB assume the sub-problems form groups that is unknown to the learner, and different clustering methods are proposed to cluster the learned models during the interaction with the environment for improved model estimation \cite{gentile2014online,li2016collaborative,gentile2017context,li2019improved}. Another line of works assume the learner has information about the sub-problem relationships, e.g. friends in social network applications, and collaborative learning is achieved via applying regularization when optimizing each sub-problem based on the known relationships \cite{cesa2013gang,wu2016contextual,yang2020laplacian}.

Distributed bandit
% The most related work in multi-agent linear bandits to ours is the distributed bandits
\citep{korda2016distributed,wang2019distributed,dubey2020differentially} is the most relevant to ours, where designing an efficient communication strategy is the main focus. Existing algorithms mainly differ in the relations of learning problems solved by the agents (i.e., identical vs., clustered) and the type of communication network (i.e., peer-to-peer (P2P) vs., star-shaped). \citet{korda2016distributed} studied two problem settings with a P2P communication network: 1) all the agents solve a common linear bandit problem, and 2) the problems are clustered. However, they only tried to reduce \textit{per-round} communication, and thus the communication cost is still linear over time.
Two follow-up studies considered the setting where all agents solve a common problem and interact with the environment in a round-robin fashion \citep{wang2019distributed,dubey2020differentially}. Similar to our work, they also used event-triggered communications to obtain a sub-linear communication cost over time.
% In particular, \citet{wang2019distributed} worked under a star-shaped communication network, and designed an event-triggered communication based on the determinant of the covariance matrix, which is similar to the method proposed in our paper.
% However, it requires global synchronization in which all clients exchange their latest observations under the central server's control. 
%that when the determinant of covariance matrix for a client varies too much from that of the most recent synchronization, a new synchronization round will be triggered, so that each client will send its latest sufficient statistics to the server, and the server will send the aggregated sufficient statistics back to all the clients.
% \citet{dubey2020differentially} extended this synchronous protocol to  peer-to-peer network, with differential privacy.
In particular, \citet{wang2019distributed} considered a star-shaped network and proposed a synchronous communication protocol for all clients to exchange their latest observations under the central server's control. 
\citet{dubey2020differentially} extended this synchronous protocol to differentially private LinUCB algorithms under both star-shaped and P2P network.

There is also a rich literature in distributed machine learning/federated learning \citep{mcmahan2017communication,li2019convergence} that studies offline optimization problems. However, as we mentioned earlier, due to the fundamental difference in the learning objectives, they are not applicable to our problem. Specifically, their main focus is to collaboratively learn a good \textit{point estimate} over a fixed dataset, i.e., convergence to the minimizer with fewer communications, while the focus of federated bandit learning is collaborative \textit{confidence interval estimation} for efficient regret reduction, which requires exploration of the unknown data. 
This is also reflected by the difference in the triggering event designs. For distributed offline optimization, triggering event measuring the change in the learned parameters suffices \citep{kia2015distributed,yi2018distributed,george2020distributed}, while for federated bandit learning, triggering event needs to measure change in the volume of the confidence region, i.e., uncertainty in the problem space. This adds serious challenges in the design and analysis of the triggering events. In addition, for linear bandit problems, thanks to the existence of the closed-form solution, there is no need to use gradient-based optimization methods like FedAvg \citep{mcmahan2017communication}, because compared with transmitting the sufficient statistics, it only costs a lot more communication overhead without bringing in any gain in regret minimization.

% Therefore, though there are works in distributed optimization literature that studies event-triggered communications \citep{kia2015distributed,yi2018distributed,george2020distributed}, 

% It is also worth noting that event-triggered communications are commonly used in distributed optimization literature \citep{kia2015distributed,yi2018distributed,george2020distributed}. But we want to emphasize that due to the fundamental difference in the learning objectives, they are not applicable to our problem. As the purpose of distributed optimization is to collaboratively learn a good \textit{point estimate} with fewer communications, a triggering event measuring the change in the learned parameters suffices. However, for federated bandit learning, the purpose is collaborative \textit{confidence interval estimation} for efficient regret reduction, so that the triggering event needs to be designed based on the volume of the confidence region, i.e., the determinant of covariance matrices. This adds serious challenges in the design and analysis of the triggering events.
\section{Methodology}
% \begin{figure*}[h]
% % \vspace{-2mm}
% \centering
% \includegraphics[width=16cm]{imgs/algo.png}
% \caption{Illustration of the proposed Asynchronous LinUCB Algorithm \textcolor{red}{make this diagram flatter? illustrate the second algorithm in this diagram as well?}}
% \label{fig:algo}
% % \vspace{-3mm}
% \end{figure*}
In this section, we establish the asynchronous event-triggered communication framework for federated linear bandit, and propose two UCB-type algorithms under different assumptions about the clients' reward functions, followed by our theoretical analysis of their regret and communication cost.
% In this section,  we first formulate the problem studied in this paper. Then we identify the connection between regret and communication cost for federated linear bandit algorithms, and establish the asynchronous event-triggered communication framework to balance these two objectives. Based on the communication framework, we propose two upper confidence bound algorithms under different assumptions about the reward functions among the clients, followed by our theoretical analysis of their regret and communication cost.

\subsection{Problem Formulation} \label{subsec:problem_formulation}
Consider a learning system with 1) $N$ clients responsible for taking actions and receiving reward feedback from the environment, e.g., each client being an edge device directly interacting with a user, and 2) a central server responsible for coordinating the communication between the clients for collaborative model estimation. 
% The clients can only communicate with the central server, instead of directly communicating with each other.
At each time $t=1,2,...,T$, an arbitrary client $i_{t} \in [N]$ (assume $P(i_{t}=i)>0,\forall i \in [N]$) interacts with the environment by choosing one of the $K$ actions, and receives the corresponding reward. When making the choice, client $i_{t}$ has access to a set $\cA_{t}=\{\bx_{t,1}, \bx_{t,2}, \dots, \bx_{t,K}\}$, where $\bx_{t, a}$ denotes the context vector associated with the $a$-th action for client $i_{t}$ at time $t$. Denote the context vector of the chosen action at time $t$ as $\bx_{t}$,
% an arm $a_{t}$ from a time-varying arm set $\cA_{t}=\{\bx_{t,1}, \bx_{t,2}, \dots, \bx_{t,|\cA_{t}|}\}$, where $\bx_{t, j}$ denotes the context vector associated with the $j$-th arm at time $t$. 
and the corresponding reward received by client $i_{t}$ as $y_{t}$, which is assumed to be generated by an unknown linear reward mapping $y_{t}=f_{i_{t}}(\bx_{t})+\eta_{t}$. As in standard linear bandit, $\eta_{t}$ is zero mean $\sigma$-sub-Gaussian noise conditioning on the $\sigma$-algebra generated by the previously pulled arms and the observed rewards $\cF_{t-1}=\sigma\{\bx_{1},y_{1},\bx_{2},y_{2},\dots,\bx_{t-1},y_{t-1},\bx_{t}\}$. Interaction between the learning system and the environment repeats itself, and the goal of the learning system is to minimize the accumulative (pseudo) regret $R_{T}=\sum_{t=1}^{T}r_{t}$ where $r_{t}=\max_{\bx \in \cA_{t}}f_{i_{t}}(\bx)-f_{i_{t}}(\bx_{t})$. 
% \chuanhao{Does the new assumption require us to change the definition of regret here, i.e., expectation over context as well? It seems not.}

\begin{figure}[t]
% \vspace{-4mm}
\centering
\includegraphics[width=14cm]{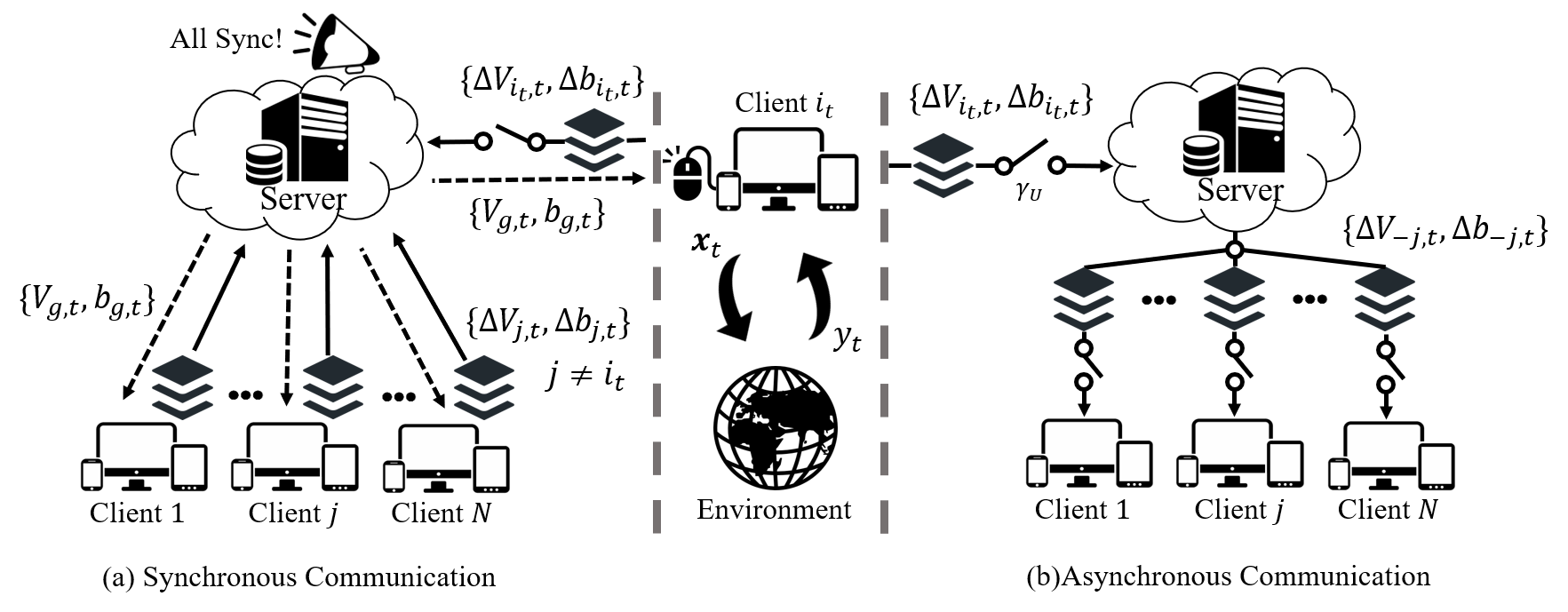}
% \vspace{-3mm}
\caption{Comparison between the synchronous and asynchronous event-triggered communications for federated linear bandit. The former requires all clients to upload their latest data at once and then download the aggregated data, while latter performs both upload and download on a per-client basis.}
\label{fig:algo}
% \vspace{-1mm}
\end{figure}

Denote the set of time steps when client $i$ interacts with the environment up to time $t$ as $\cN_{i}(t)=\left\{1 \leq \tau \leq t:i_{\tau}=i\right\}$. Note that we do not impose any further assumption on the clients' distribution or frequency of its interactions. This makes our setting more general than existing works \citep{wang2019distributed,dubey2020differentially}, where the clients interact with the environment in a round-robin fashion, i.e., $|\cN_{i}(t)|=t/N$. 
In addition, in the federated learning setting, the clients cannot directly communicate with each other, but can only communicate with the central server, i.e., a star-shaped communication network. Raw data collected by each client $\{(\bx_{\tau},y_{\tau})\}_{\tau \in \cN_{i}(t)}$ is stored locally and will not be transferred. Instead, the clients can only collaborate by communicating the parameters of the learning algorithm, e.g., gradients or sufficient statistics; and the communication cost is measured by the amount of parameters transferred across the system up to time $T$, which is denoted by $C_{T}$.
% Raw data sharing is also not allowed, and the clients can only collaborate by communicating the parameters of the learning algorithm, which in this paper, are just the sufficient statistics for the raw data, since analytical solution exists for linear bandit problem. And the communication cost is simply measured by the number of times data is transferred from one agent in the learning system (either client or server) to another.

We consider two different settings about the linear reward mapping function $f_{i}(\cdot)$ for $i=1,\dots,N$:

\paragraph{Homogeneous clients.}
Rewards received by all the clients are generated by a common reward mapping function:
    \begin{equation}\label{eq:homo_theta}
        f_{i}(\bx)=\theta^{\top} \bx, \quad \forall i \in [N]
    \end{equation}
    where $\theta \in \bR^{d}$ is the unknown parameter and we assume $\lVert \theta\rVert \leq 1$ and $\lVert \bx \rVert \leq 1$. Despite its simplicity, this setting is commonly adopted in existing works for federated bandits.
    % , which we refer to as homogeneous clients.
\paragraph{Heterogeneous clients.}
The unknown parameter for each client $i\in [N]$ consists of a globally shared component $\theta^{(g)}\in \bR^{d_{g}}$ and a unique local component $\theta^{(i)}\in \bR^{d_{i}}$: 
    \begin{equation}\label{eq:hetero_theta}
        f_{i}(\bx)= \begin{bmatrix} \theta^{(g)} \\ \theta^{(i)} \end{bmatrix}^{\top} \begin{bmatrix} \bx^{(g)} \\ \bx^{(l)} \end{bmatrix} , \quad \forall i \in [N]
    \end{equation}
    where $\bx^{(g)}\in \bR^{d_{g}}$, $\bx^{(l)}\in \bR^{d_{i}}$ denote the global and local features in $\bx$, and we assume $||\theta^{(g)}||_{2} \leq 1$, $||\theta^{(i)}||_{2} \leq 1, \forall i\in [N]$ and $||\bx^{(g)}||_{2} \leq 1$, $||\bx^{(l)}||_{2} \leq 1$. 
    This setting is more general and fits a larger variety of problems in practice. For example, $\bx^{(g)}$ could be common arm features relevant to all the clients and $\bx^{(l)}$ are those unique to client $i$. And our setting is flexible enough to allow different clients to have varying dimensions of their local features (i.e., $d_{i}\ne d_j$). Alternatively, when $\bx^{(g)}\equiv\bx^{(l)}$, this recovers the multi-task learning setting in \cite{evgeniou2004regularized}. 
    % We refer to this setting as heterogeneous clients, and 

We adopt the context regularity assumption from \cite{gentile2014online,li2019improved,li2021unifying}, which imposes a variance condition on the stochastic process generating $\bx_{t,a}$ (for heterogeneous clients, it is imposed on global features $\bx_{t,a}^{(g)}$). This suggests the informativeness of each observation in expectation. 
% In practice, it means each arm has diverse features over time.
% In practice, it means each arm has diverse features over time, i.e., possible values of $\bx_{t,a}$ for each $a \in [K]$ span $\bR^{d}$. This is slightly stronger than the common assumption that the possible values of $\{\bx_{t,a}\}_{a\in [K]}$ together span $\bR^{d}$.
% But we will see that this assumption does give existing methods \cite{wang2019distributed} any advantage.
% \begin{assumption}[Context regularity] \label{assump:context_diversity}
% At each time $t$, the context vector $\bx_{t,a} \in \cA_{t}$ for each arm $a\in [K]$ is independently generated from a random process, such that $\bE_{t-1}[\bx_{t,a}\bx_{t,a}^{\top}] := \bE[\bx_{t,a}\bx_{t,a}^{\top}|\{i_{s},\cA_{s},\eta_{s}\}_{s\in [t-1]}] =\Sigma_{c} \succeq \lambda_{c} I, \forall t\in [T]$,
% % \begin{align*}
% %     & \bE_{t-1}[\bx_{t,a}\bx_{t,a}^{\top}] := \bE[\bx_{t,a}\bx_{t,a}^{\top}|\{i_{s},\cA_{s},\eta_{s}\}_{s\in [t-1]}] \\
% %     & =\Sigma_{c} \succeq \lambda_{c} I, \forall t\in [T]
% % \end{align*}
% where the constant $\lambda_{x} > 0$.
% % s
% % $\forall a \in [K], \bE_{t-1}[\bx_{t,a} \bx_{t,a}^{\top}]=\bE[\bx_{t,a} \bx_{t,a}^{\top}|\{\eta_{},\bx_{t,a}\}] =\Sigma_{x} \geq \lambda_{min} I$, or equivalently $\text{span}\{\bx_{t,a}|\bx_{t,a} \in \cX_{i_{t}}, i \in [N] \}=\bR^{d}$.
% % $\bE_{t-1}[\bx_{t,a} \bx_{t,a}^{\top}]=\bE[\bx_{t,a} \bx_{t,a}^{\top}|\{\eta_{},\bx_{t,a}\}]$
% \end{assumption}
\begin{assumption}[Context regularity] \label{assump:context_diversity}
At each time $t$, the context vector $\bx_{t,a} \in \cA_{t}$ for each arm $a\in [K]$ is independently generated from a random process, such that $$\bE_{t-1}[\bx_{t,a}\bx_{t,a}^{\top}] := \bE[\bx_{t,a}\bx_{t,a}^{\top}|\{i_{s},\cA_{s},\eta_{s}\}_{s\in [t-1]}] =\Sigma_{c} \succeq \lambda_{c} I, \forall t\in [T]$$ where the constant $\lambda_{c} > 0$. Let also, for any fixed unit vector $z \in \bR^{d}$, the random variable $(z^{\top} \bx_{t,a})^{2}$ be conditionally sub-Gaussian with variance parameter $v^{2} \leq {\lambda_{c}^{2}}/{(8\log{4K})}$.
\end{assumption}

\subsection{Asynchronous Communication Framework}\label{subsec:async_comm}
In order to balance the two conflicting objectives, i.e., regret $R_{T}$ and communication cost $C_{T}$, we introduce an asynchronous event-triggered communication framework as illustrated in Figure \ref{fig:algo}(b). For simplicity, all discussions in this section assume homogeneous clients (Eq \eqref{eq:homo_theta}), and we show in Section \ref{subsec:async_LinUCB_AM} that the result extends to heterogeneous clients (Eq \eqref{eq:hetero_theta}) as well with minor modifications. Also note that in this paper we use LinUCB \citep{abbasi2011improved} with our communication framework as a running example, but our results readily hold for other popular algorithms like LinTS \citep{abeille2017linear} and LinPHE \citep{kveton2019perturbed}  \footnote{Their regret bounds also depend on $\sum_{t=1}^{T}\lVert \bx_{t} \rVert_{V_{t-1}^{-1}}$ (see $R_{\text{TS}}(T)$ in Section 4 of \citet{abeille2017linear} and Theorem 1 in \citet{kveton2019perturbed}). Therefore a similar procedure can be applied, i.e., plug our Algorithm \ref{algo:comm} into LinTS to communicate the $V$ matrix, or into LinPHE to communicate the unperturbed $G$ matrix.}.

We begin our discussion with an important observation about the instantaneous regret of linear bandit algorithms.
% \cite{abbasi2011improved,abeille2017linear,kveton2019perturbed}.
% The design of our asynchronous event-triggered communication framework is based on the following observation.
Denote the sufficient statistics (for $\theta$) collected from all clients by time $t$ as ${V}_{t}=\sum_{\tau=1}^{t}\bx_{\tau}\bx_{\tau}^{\top}$ and ${b}_{t}=\sum_{\tau=1}^{t}\bx_{\tau}y_{\tau}$. In a centralized setting, at each time step $t\in[T]$, $\{{V}_{t-1},{b}_{t-1}\}$ are readily available to make an informed choice of arm $\bx_{t}\in\cA_{t}$. It is known that the instantaneous regret $r_{t}$ incurred by the mentioned linear bandit algorithms is directly related to the width of the confidence ellipsoid in the direction of $\bx_{t}$. Specifically, from Theorem 3 in \citep{abbasi2011improved}, with probability at least $1-\delta$, the instantaneous regret $r_{t}$ incurred by LinUCB can be upper bounded by $r_{t}  \leq 2 \alpha_{t-1} \sqrt{\bx_{t}^{\top}V_{t-1}^{-1}\bx_{t}} $, where $\alpha_{t-1}=O\left(\sqrt{d\log{\frac{T}{\delta}}}\right)$.
% \textcolor{blue}{we can mention this framwork is general because regret analysis of most linear bandit algorithms rely on diameter of the ellipsoid? so similar procedure should still work for other algos, like LinTS?}
%\begin{align*}
%    r_{t} & \leq O\left(\sqrt{d\log{\frac{T}{\delta}}}\right) \sqrt{\bx_{t}^{\top}V_{t-1}^{-1}\bx_{t}} 
%\end{align*}
However, as data is decentralized in our problem, $\{{V}_{t-1},{b}_{t-1}\}$ are not readily available to client $i_{t}$. Instead, the client only has a delayed copy, denoted by  $\{{V}_{i_{t},t-1},{b}_{i_{t},t-1}\}$, which is obtained by its own interactions with the environment on top of the last communication with the server. Therefore, now the instantaneous regret $r_{t} \leq  2 \alpha_{i_{t},t-1} \sqrt{\bx_{t}^{\top}V_{i_{t},t-1}^{-1}\bx_{t}} = 2 \alpha_{i_{t},t-1} \sqrt{\bx_{t}^{\top}{V}_{t-1}^{-1}\bx_{t}}\sqrt{\Gamma_{t-1}}$, where $\Gamma_{t-1} =\frac{\bx_{t}^{\top}{V}_{i_{t},t-1}^{-1}\bx_{t}}{\bx_{t}^{\top}{V}_{t-1}^{-1}\bx_{t}}$ measures how much wider the confidence ellipsoid at client $i_{t}$'s estimation in the direction of $\bx_{t}$ is, compared with that under a centralized setting. The value of $\Gamma_{t-1}$ depends on how frequent local updates are aggregated and shared via the server. Also note that $\Gamma_{t-1} \geq 1$, as ${V}_{t-1} \succeq {V}_{i_{t},t-1},\forall t$, which suggests the regret in the decentralized setting is at best the same as that in the centralized setting. Equality is attained when all the clients are synchronized in every time step.

Based on this observation, we can balance regret and communication cost by controlling the value of $\Gamma_{t-1}$. 
However, in the decentralized setting, neither the server nor the clients has direct access to $\{V_{t-1},b_{t-1}\}$, and the closest thing one can get is the aggregated sufficient statistics managed by the server, which we denote as $\{V_{g,t-1},b_{g,t-1}\}$. Hence, we take an indirect approach by first ensuring $\{V_{g,t-1},b_{g,t-1}\}$ do not deviate too much from $\{V_{t-1},b_{t-1}\}$, and then $\{V_{i,t-1},b_{i,t-1}\}$ do not deviate too much from $\{V_{g,t-1},b_{g,t-1}\}$ for each client $i\in [N]$. The former leads to the `upload' event, i.e., each client decides whether to upload independently, and the latter leads to the `download' event, i.e., the server decides whether to send its latest statistics to each client independently as well. 
In the proposed communication framework shown in Figure \ref{fig:algo}(b), each client $i\in[N]$ stores a local copy of its sufficient statistics $\{V_{i,t-1},b_{i,t-1}\}$, and also an `upload' buffer $\{\Delta V_{i,t-1},\Delta b_{i,t-1}\}$, i.e., client $i$'s local updates that have not been sent to the server. At each time step $t$, client $i_{t}\in [N]$ interacts with the environment, and updates $V_{i_{t},t}=V_{i_{t},t-1}+\bx_{t}\bx_{t}^{\top},b_{i_{t},t}=b_{i_{t},t-1}+\bx_{t}y_{t}$, $\Delta V_{i_{t},t}=\Delta V_{i_{t},t-1}+\bx_{t}\bx_{t}^{\top},\Delta b_{i_{t},t}=\Delta b_{i_{t},t-1}+\bx_{t}y_{t}$ with the new observation $(\bx_{t},y_{t})$. Then it starts executing 
% the communication protocol described in 
Algorithm \ref{algo:comm}, by first checking the following condition (line 2):
% First, it checks the following condition (line 2 in Algorithm \ref{algo:comm}):

\noindent \textbf{`Upload' event}: Client $i_{t}$ sends $\{\Delta V_{i_{t},t},\Delta b_{i_{t},t}\}$ to the server if event:
% \small
    \begin{equation}\label{eq:upload}
        \cU_{t}(\gamma_{U}) = \left\{\frac{\det(V_{i_{t},t})}{\det(V_{i_{t},t}-\Delta V_{i_{t},t})} > \gamma_{U} \right\}
    \end{equation}
% \normalsize
happens, and then sets $\Delta V_{i,t}=\textbf{0},\Delta b_{i,t}=\textbf{0}$. Otherwise, $\{\Delta V_{i,t}, \Delta b_{i,t}\}$ remain unchanged.

The server stores the aggregated sufficient statistics $\{V_{g,t-1},b_{g,t-1}\}$ over the local updates received from the clients, and also maintains `download' buffers $\{\Delta V_{-j,t-1},\Delta b_{-j,t-1}\}$ for each client $j\in [N]$, i.e., the aggregated updates that have not been sent to client $j$. Specifically, after the server receives $\{\Delta V_{i_{t},t},\Delta b_{i_{t},t}\}$ via the `upload' from client $i_t$, it updates $V_{g,t}=V_{g,t-1}+\Delta V_{i_{t},t},b_{g,t}=b_{g,t-1}+\Delta b_{i_{t},t}$, and $\Delta V_{-j,t}=\Delta V_{-j,t-1}+\Delta V_{i_{t},t},\Delta b_{-j,t}=\Delta b_{-j,t-1}+\Delta b_{i_{t},t}$ for all clients $j \neq i_{t}$. Then it checks the following condition for each client $j \neq i_{t}$ (line 7):

\noindent \textbf{`Download' event}: The server sends $\{\Delta V_{j,t},\Delta b_{j,t}\}$ to client $j$ if event:
% \small
    \begin{equation}\label{eq:download}
        \cD_{t,j}(\gamma_{D})=\left\{\frac{\det(V_{g.t})}{\det(V_{g,t}-\Delta V_{-j,t})} > \gamma_{D}\right\}
    \end{equation}
% \normalsize
happens, and then sets $\Delta V_{-j,t}=\textbf{0},\Delta b_{-j,t}=\textbf{0}$. Otherwise, $\{\Delta V_{-j,t}, \Delta b_{-j,t}\}$ remain unchanged.

After client $j$ receives $\{\Delta V_{-j,t},\Delta b_{-j,t}\}$ via the `download' communication, it updates $V_{j,t}=V_{j,t-1}+\Delta V_{-j,t},b_{j,t}=b_{j,t-1}+\Delta b_{-j,t}$.
% \begin{itemize}
%     \item `Upload': At time $t$, after client $i_{t}$ observes the new data point $(\bx_{t},y_{t})$ and locally updates $\tilde{V}_{i_{t},t}$ and $\tilde{b}_{i_{t},t}$, client $i_{t}$ will send $\tilde{V}_{i_{t},t}$ and $\tilde{b}_{i_{t},t}$ to the server if:
%     \begin{equation}\label{eq:upload}
%         \frac{\det(\tilde{V}_{i_{t},t}+\bar{V}_{-i_{t},t-1})}{\det(\bar{V}_{i_{t},t-1}+\bar{V}_{-i_{t},t-1})} \geq \gamma_{U}
%     \end{equation}
%     \item `Download': Upon receiving sufficient statistics from client $i_{t}$, the server will update its aggregated sufficient statistics accordingly. The server will then check the following condition for each client $j\neq i_{t}$, and send $\sum_{i \neq j} \bar{V}_{i,t}$ and $\sum_{i \neq j} \bar{b}_{i,t}$ to client $j$ if:
%     \begin{equation}\label{eq:download}
%         \frac{\det(\bar{V}_{j,t-1}+\sum_{i \neq j} \bar{V}_{i,t})}{\det(\bar{V}_{j,t-1}+\bar{V}_{-j,t-1})} \geq \gamma_{D}
%     \end{equation}
% \end{itemize}
\begin{algorithm}[h]
    \caption{Event-triggered Communication Protocol} \label{algo:comm}
  \begin{algorithmic}[1]
    \STATE \textbf{Input:} thresholds $\gamma_{U}, \gamma_{D} \geq 1$
        \IF{ Event $\cU_{t}(\gamma_{U})$ in Eq \eqref{eq:upload} happens}
            \STATE Upload $\Delta{V}_{i_{t},t},\Delta{b}_{i_{t},t}$ ($\text{client } i_{t} \rightarrow \text{server}$) 
            \STATE Update server:
            % $V_{g,t}=V_{g,t-1}+\Delta V_{i_{t},t},b_{g,t}=b_{g,t-1}+\Delta b_{i_{t},t}$ \newline
            % $\Delta V_{-j,t}=\Delta V_{-j,t-1}+\Delta V_{i_{t},t},\Delta b_{-j,t}=\Delta b_{-j,t-1}+\Delta b_{i_{t},t}$ for $j \neq i_{t}$
            $V_{g,t}\mathrel{+}=\Delta V_{i_{t},t},b_{g,t}\mathrel{+}=\Delta b_{i_{t},t}$, $\Delta V_{-j,t}\mathrel{+}=\Delta V_{i_{t},t},\Delta b_{-j,t}\mathrel{+}=\Delta b_{i_{t},t}$, $\forall j \neq i_{t}$
            \STATE Client $i_{t}$ sets $\Delta{V}_{i_{t},t}=\textbf{0}$, $\Delta{b}_{i_{t},t}=\textbf{0}$
            % and $j \neq i_{t}$
            \FOR{$j = 1, \dots, N$} 
                \IF{ Event $\cD_{t,j}(\gamma_{D})$ in Eq \eqref{eq:download} happens}
                    \STATE Download $\Delta{V}_{-j,t},\Delta{b}_{-j,t}$ ($\text{server} \rightarrow \text{client } j$)
                    \STATE Update client $j$:
                    % $V_{j,t}=V_{j,t-1}+\Delta V_{-j,t},b_{j,t}=b_{j,t-1}+\Delta b_{-j,t}$
                    $V_{j,t}\mathrel{+}=\Delta V_{-j,t},b_{j,t}\mathrel{+}=\Delta b_{-j,t}$
                    \STATE Server sets $\Delta{V}_{-j,t}=\textbf{0},\Delta{b}_{-j,t}=\textbf{0}$
                \ENDIF
            \ENDFOR
        % \ELSE
        %     \STATE All other sufficient statistics remain unchanged
        \ENDIF
  \end{algorithmic}
\end{algorithm}

The following lemma specifies an upper bound of $\Gamma_{t-1}$ by executing Algorithm \ref{algo:comm}, which depends on the thresholds $\{\gamma_{U}, \gamma_{D}\}$ and the number of clients $N$.
\begin{lemma}\label{lem:Gamma_upperbound}
Under Assumption \ref{assump:context_diversity}, the `upload' and `download' triggering events defined in Eq \eqref{eq:upload} and Eq \eqref{eq:download}, then w.h.p. $\Gamma_{t-1} \leq \frac{8\gamma_{D}}{\lambda_{c}} [1+(N-1)(\gamma_{U}-1)],\forall t$.
\end{lemma}

Proof of Lemma \ref{lem:Gamma_upperbound} is given in appendix. The main idea is to use $\det(V_{g,t-1})$ as an intermediate between $\det(V_{i_{t},t-1})$ and $\det(V_{t-1})$, which are separately controlled by the `download' and `upload' events. When setting $\gamma_{D}=\gamma_{U}=1$, $\Gamma_{t-1}=1$, $\forall t \in [T]$, which means global synchronization happens at each time step, it recovers the regret incurred in the centralized setting.

% \begin{remark}[Synchronous vs. asynchronous communication] 
\noindent \textbf{Synchronous vs. asynchronous communication:}
%As illustrated in Figure \ref{fig:algo}, both the synchronous communication from \cite{wang2019distributed,dubey2020differentially} and the proposed asynchronous communication are event-triggered. 
% (based on determinant of covariance matrix). 
%In the former, 
As shown in Figure \ref{fig:algo}(a), in the synchronous protocol (Appendix G in \citep{wang2019distributed}), when a synchronization round is triggered by a client $i_{t}$, the server asks \emph{all} the clients to upload their local updates (illustrated as solid lines), aggregates them, and then sends the aggregated update back (illustrated as dashed lines). This `two-way' communication is vulnerable to delays or unavailability of clients, which are common in a distributed setting.
In comparison, our asynchronous communication, as shown in Figure \ref{fig:algo}(b), is more robust because the server only concerns the clients whose `download' condition has been met, which does not need other clients' acknowledgement. 
In addition, when the clients have distinct availability of new observations, which is usually the case for most applications, synchronizing all $N$ clients leads to inefficient communication as some clients may have very few new observations since last synchronization. We will show later that this unfortunately leads to an increased rate in $N$ in the upper bound of $C_{T}$, compared with our asynchronous communication.
% In addition, global synchronization requires every client in the learning system to send its local update despite the53fact some clients have very few new observations since last synchronization.
% In comparison, in the existing method \cite{wang2019distributed,dubey2020differentially}, synchronous upload from all $N$ clients is triggered if any client in the learning system deviates too much from the last communicated data, and then the server will passively aggregate all the uploaded data and return it to all $N$ clients.
% \end{remark}
% For example, when some client $j$ is temporarily unavailable to receive `download' communication or the packet is lost, the learning system can proceed to next round of interaction as usual, with the server trying to re-send the update in the background. As long as client $j$ receives the update before it interacts with the environment next time, the upper bound of $\Gamma_{t-1}$ in Lemma \ref{lem:Gamma_upperbound} always holds.

% \begin{remark}[Multiple Clients per Time Step]
\noindent \textbf{Multiple clients per time:}
Note that essentially both Algorithm \ref{algo:comm} and the synchronous protocol assumed only one active client per time, i.e., the communication protocol is executed after the current client receives a new observation and completed \textit{before} the next client shows up. 
This setup simplifies the description and also makes the theoretical results compatible with standard linear bandit setting. Otherwise, we will need to introduce additional assumptions to quantify the extent of delay in the communication for regret analysis.
In reality, each client is an independent process interacting with its environment, e.g., serving its user population, so that there could be many active clients at the same time. In the proposed asynchronous protocol, when a client triggers the upload event, it will immediately send the data in its buffer to the server; the server, upon receiving the uploaded data from any client, will immediately aggregate this data and check the download events to see which client needs a download. Thus, different from the synchronous protocol where the server ensures all the clients have the same model after each download, we allow both the server and the clients to update their model at any time without the need of any global synchronization.

\subsection{Learning with Homogeneous Clients}
\label{subsec:async_LinUCB}
Based on the asynchronous event-triggered communication, we design the Asynchronous LinUCB Algorithm (\modelone{}) for homogeneous clients. Detailed steps are explained in Algorithm \ref{algo:AsyncLinUCB}.

\noindent \textbf{Arm selection}:
At each time step $t=1,\dots,T$, 
% when client $i_{t}\in[N]$ interacts with the environment, 
% it has access to sufficient statistics of its local data $\tilde{V}_{i_{t},t-1},\tilde{b}_{i_{t},t-1}$ and the aggregated sufficient statistics $\bar{V}_{-i_{t},t-1},\bar{b}_{-i_{t},t-1}$ recently downloaded from the server. 
client $i_{t}$ selects an arm $\bx_{t} \in \cA_{t}$ using the the UCB strategy. Specifically, client $i_{t}$ pulls arm $\bx_{t}$ that maximizes the UCB score computed as follows (line 8),
\begin{equation}\label{eq:UCB}
    \bx_{t}=\argmax_{\bx \in \cA_{t}}{\bx^{\top}\hat{\theta}_{i_{t},t-1}(\lambda)+\text{CB}_{i_{t},t-1}(\bx)}
\end{equation}
where $\hat{\theta}_{i_{t},t-1}(\lambda)= V_{i_{t},t-1}(\lambda)^{-1}b_{i_{t},t-1}$ is the ridge regression estimator with regularization parameter $\lambda$; $V_{i_{t},t-1}(\lambda)=V_{i_{t},t-1}+\lambda I$; and the confidence bound of reward estimation for arm $\bx$ is $\text{CB}_{i_{t},t-1}(\bx)=\alpha_{i_{t},t-1}||\bx||_{V_{i_{t},t-1}(\lambda)^{-1}}$, where 
$\alpha_{i_{t},t-1}=\sigma \sqrt{\log{\frac{\det{V_{i_{t},t-1}(\lambda) }}{\det{\lambda I}}}+2\log{1/\delta}}+\sqrt{\lambda}$.
% $\alpha_{i_{t},t-1}=\sigma\sqrt{d\log{(1+\frac{t-1}{d\lambda})} + 2\log{\frac{1}{\delta}}}+\sqrt{\lambda}S$.
% where $\hat{\theta}_{i_{t},t-1}= \bigl(\bar{V}_{i_{t},t-1}+\tilde{V}_{-i_{t},t-1}+\lambda I\bigr)^{-1}\bigl(\bar{b}_{i_{t},t-1}+\tilde{b}_{-i_{t},t-1}\bigr)$ is the ridge regression estimator; and the confidence bound of reward estimation for arm $\bx$ is $\text{CB}_{i_{t},t-1}(\bx)=\alpha_{i_{t},t-1}||\bx||_{(\tilde{V}_{i_{t},t-1}+\tilde{V}_{-i_{t},t-1}+\lambda I)^{-1}}$, where $\alpha_{i_{t},t-1}=\sigma\sqrt{d\log{(1+\frac{t-1}{d\lambda})} + 2\log{\frac{1}{\delta}}}+\sqrt{\lambda}S$.
After client $i_{t}$ observes reward $y_{t}$ and updates locally (line 9), it proceeds with the asynchronous event-triggered communication (line 10), and sends updates accordingly.

\begin{algorithm}[h]
    \caption{\modelone{}} \label{algo:AsyncLinUCB}
  \begin{algorithmic}[1]
    \STATE \textbf{Input:} thresholds $\gamma_{U}, \gamma_{D} \geq 1$, $\sigma, \lambda > 0$, $\delta \in (0,1)$
    \STATE Initialize server: ${V}_{g, 0}=\textbf{0}_{d,d}$, ${b}_{g,0}=\textbf{0}_{d}$
        % \hspace*{2em} $\Delta{V}_{-i, 0}=\textbf{0} \in \mathbb{R}^{d \times d}$, $\Delta{b}_{-i,0}=\textbf{0} \in \mathbb{R}^{d}$, $i\in[N]$
    \FOR{ $t=1,2,...,T$}
        \STATE Observe arm set $\mathcal{A}_{t}$ for client $i_{t} \in [N]$
        \IF{client $i_{t}$ is new}
            \STATE Initialize client $i_{t}$: ${V}_{i_{t}, t-1}=\textbf{0}_{d,d}$, ${b}_{i_{t},t-1}=\textbf{0}_{d}$, $\Delta{V}_{i_{t}, t-1}=\textbf{0}_{d,d} $, $\Delta{b}_{i_{t},t-1}=\textbf{0}_{d}$
            \STATE Initialize server's download buffer for client $i_{t}$: $\Delta{V}_{-i_{t}, t-1}=V_{g,t-1}$, $\Delta{b}_{-i_{t},t-1}=b_{g,t-1}$
        \ENDIF
        \STATE Select arm $\bx_{t}\in\cA_{t}$ by Eq \eqref{eq:UCB} and observe reward $y_{t}$
        \STATE Update client $i_{t}$:
            % ${V}_{i_{t},t}={V}_{i_{t},t-1}+x_{t}x_{t}^{T}$, ${b}_{i_{t},t}={b}_{i_{t},t-1}+x_{t}y_{t}$ \newline
            % $\Delta{V}_{i_{t},t}=\Delta{V}_{i_{t},t-1}+x_{t}x_{t}^{T}$, $\Delta{b}_{i_{t},t}=\Delta{b}_{i_{t},t-1}+x_{t}y_{t}$
            ${V}_{i_{t},t} \mathrel{+}= \bx_{t}\bx_{t}^{T}$, ${b}_{i_{t},t} \mathrel{+}= \bx_{t}y_{t}$, $\Delta{V}_{i_{t},t} \mathrel{+}= \bx_{t}\bx_{t}^{T}$, $\Delta{b}_{i_{t},t}+=\bx_{t}y_{t}$
        \STATE Event-triggered Communications (Algorithm \ref{algo:comm})
        % \IF{ Event $\cU_{t}(\gamma_{U})$ (Eq \eqref{eq:upload}) happens}
        %     \STATE Upload $\Delta{V}_{i_{t},t},\Delta{b}_{i_{t},t}$ ($i_{t} \rightarrow \text{server}$) 
        %     \STATE Update server: \newline
        %     % $V_{g,t}=V_{g,t-1}+\Delta V_{i_{t},t},b_{g,t}=b_{g,t-1}+\Delta b_{i_{t},t}$ \newline
        %     % $\Delta V_{-j,t}=\Delta V_{-j,t-1}+\Delta V_{i_{t},t},\Delta b_{-j,t}=\Delta b_{-j,t-1}+\Delta b_{i_{t},t}$ for $j \neq i_{t}$
        %     \hspace*{1em} $V_{g,t}\mathrel{+}=\Delta V_{i_{t},t},b_{g,t}\mathrel{+}=\Delta b_{i_{t},t}$ \newline
        %     \hspace*{1em} $\Delta V_{-j,t}\mathrel{+}=\Delta V_{i_{t},t},\Delta b_{-j,t}\mathrel{+}=\Delta b_{i_{t},t}$, $j \neq i_{t}$
        %     \STATE Client $i_{t}$ reset $\Delta{V}_{i_{t},t}=\textbf{0}$, $\Delta{b}_{i_{t},t}=\textbf{0}$

        %     \FOR{$j = 1, \dots, N$ and $j \neq i_{t}$}
        %         \IF{ Event $\cD_{t,j}(\gamma_{D})$ (Eq \eqref{eq:download}) happens}
        %             \STATE Download $\Delta{V}_{-j,t},\Delta{b}_{-j,t}$ ($\text{server} \rightarrow j$)
        %             \STATE Update client $j$: \newline
        %             % $V_{j,t}=V_{j,t-1}+\Delta V_{-j,t},b_{j,t}=b_{j,t-1}+\Delta b_{-j,t}$
        %             \hspace*{1em} $V_{j,t}\mathrel{+}=\Delta V_{-j,t},b_{j,t}\mathrel{+}=\Delta b_{-j,t}$
        %             \STATE Server reset $\Delta{V}_{-j,t}=\textbf{0},\Delta{b}_{-j,t}=\textbf{0}$
        %         \ENDIF
        %     \ENDFOR
        % % \ELSE
        % %     \STATE All other sufficient statistics remain unchanged
        % \ENDIF
    \ENDFOR
  \end{algorithmic}
\end{algorithm}
The upper bounds of cumulative regret $R_{T}$ and communication cost $C_{T}$ incurred by \modelone{} are given in Theorem \ref{thm:regret_communication_async-linucb} (proof provided in appendix). Note that as discussed in Section \ref{subsec:async_comm}, clients collaborate by transferring updates of the sufficient statistics, i.e., $\{ \Delta V \in \bR^{d \times d},\Delta b \in \bR^{d}\}$. Since our target is not to reduce the size of these parameters, we define the communication cost $C_{T}$ as the number of times $\{ \Delta V,\Delta b\}$ being transferred between agents.
\begin{theorem}[Regret and Communication Cost] \label{thm:regret_communication_async-linucb}
With Assumption \ref{assump:context_diversity}, and the communication thresholds $\gamma_{U},\gamma_{D}$, then w.h.p., the cumulative regret $$R_{T} = O\left(d\sqrt{T\log^{2}{T}}\min(\sqrt{N},\sqrt{\gamma_{D} [1+(N-1)(\gamma_{U}-1)]})\right)$$ and communication $$C_{T} = O\bigl(d N {\log{T}}/{\log{\min{(\gamma_{U},\gamma_{D})}}}\bigr)$$
% $R_{T}$ has upper bound
% \small
% $$R_{T} = O\left(d\sqrt{T\log^{2}{T}}\min(\sqrt{N},\sqrt{\gamma_{D} [1+(N-1)(\gamma_{U}-1)]})\right)$$
% \normalsize
% The communication cost $C_{T}$ has upper bound
% $$C_{T} =\sum_{i=1}^{N} C_{T,i} \leq N \frac{\log{\det({V}_{T-1})}-d \log{\lambda}}{\log{\min{(\gamma_{U},\gamma_{D})}}}$$
\end{theorem}
The thresholds $\gamma_{U},\gamma_{D}$ can be flexibly adjusted to trade-off between $R_{T}$ and $C_{T}$, e.g., interpolate between the two extreme cases: clients never communicate ($R_{T}=O(N^{1/2}d\sqrt{T}\log{T})$); and clients are synchronized in every time step ($R_{T}=O(d\sqrt{T}\log{T})$). Details about threshold selection and the corresponding theoretical results are provided in appendix. For simplicity, we fix $\gamma_{U}=\gamma_{D}=\gamma$ in the following discussions, but one can choose different values to have a finer control especially for applications where the cost of upload and download communication differs. 
Based on Theorem \ref{thm:regret_communication_async-linucb}, to attain $R_{T}=O(N^{1/4}d\sqrt{T}\log{T})$, \modelone{} needs $C_{T}=O(N^{3/2}d\log{T})$ (by setting $\gamma=\exp(N^{-\frac{1}{2}})$). To attain the same $R_{T}$, the corresponding $C_{T}$ of \modelbaseline{} \footnote{\modelbaseline{} refers to DisLinUCB algorithm in Appendix G of \citep{wang2019distributed} adapted to our setting.} is smaller than ours by a factor of $O(N^{1/4})$ only under uniform client distribution ($P(i_{t}=i) = \frac{1}{N},\forall i\in[N]$), while under non-uniform client distribution, which is almost always the case in practice, it is higher than ours by a factor of $O(N^{1/4})$. The description and theoretical analysis of \modelbaseline{} under uniform and non-uniform client distribution are given in appendix.
\subsection{Learning with Heterogeneous Clients}
\label{subsec:async_LinUCB_AM}
In this section, we study the setting of heterogeneous clients as defined in Eq \eqref{eq:hetero_theta}. 
%The main challenge is how to fit a separate reward function for each $f_{i}(x),i\in[N]$, with improved sample efficiency by leveraging the relationship among the clients. 
%As we assume this relation is captured by the globally shared bandit parameter
As the clients only share $\theta^{(g)}$, we need to learn the global component $\theta^{(g)}$ collaboratively by all clients, while learning the personalized local component $\theta^{(i)}$ individually for each client $i$.
% However, the need for  brings extra challenge to model estimation and communication compared with homogeneous clients (Eq \eqref{eq:homo_theta}).
We adopt an Alternating Minimization (AM) method to separately update the two components, and use the asynchronous communication framework in Algorithm \ref{algo:comm} to ensure communication-efficient learning of $\theta^{(g)}$. The resulting algorithm is named Asynchronous LinUCB with Alternating Minimization (\modeltwo{}), and its detailed steps are given in Algorithm \ref{algo:AsyncLinUCB_AM} \footnote{To simplify the description, availability of an unbiased estimate of $\theta^{(g)}$ to initialize the AM steps is assumed here. A slightly modified version that drops this assumption is provided in appendix.}.

\begin{algorithm}[h]
    \caption{\modeltwo{}} \label{algo:AsyncLinUCB_AM}
  \begin{algorithmic}[1]
    \STATE \textbf{Input:} thresholds $\gamma_{U}, \gamma_{D} \geq 1$, $\sigma, \lambda > 0$, $\delta \in (0,1)$
    \STATE Initialize server: ${V}_{g, 0}=\textbf{0}_{d_{g} \times d_{g}}$, ${b}_{g,0}=\textbf{0}_{d_{g}}$
    \FOR{ $t=1,2,...,T$}
        \STATE Observe arm set $\mathcal{A}_{t}$ for client $i_{t} \in [N]$
        \IF{Client $i_{t}$ is new}
            \STATE Initialize client $i_{t}$: ${V}_{i_{t}, t-1}=\textbf{0}_{d_{g} \times d_{g}}$, ${b}_{i_{t},t-1}=\textbf{0}_{d_{g}}$, $\Delta{V}_{i_{t}, t-1}=\textbf{0}_{d_{g} \times d_{g}}$, $\Delta{b}_{i_{t},t-1}=\textbf{0}_{d_{g}}$, ${V}^{(l)}_{i_{t},t-1}=\textbf{0}_{d_{i_{t}},d_{i_{t}}}$, ${b}^{(l)}_{i_{t},t-1}=\textbf{0}_{d_{i_{t}}}$
            \STATE Initialize server's download buffer for client $i_{t}$: $\Delta{V}_{-i_{t}, t-1}=V_{g,t-1}$, $\Delta{b}_{-i_{t},t-1}=b_{g,t-1}$
        \ENDIF
        \STATE Select arm $\bx_{t}\in\cA_{t}$ by Eq \eqref{eq:UCB_AM} and observe reward $y_{t}$
        \STATE Run AM by Eq \eqref{eq:AM} to estimate partial rewards:  $\hat{y}^{(g)}_{t}=y_{t}-{\bx_{t}^{(l)}}^{\top} \hat{\theta}^{(l)}_{i_{t},t}$, $\hat{y}^{(l)}_{t}=y_{t}-{\bx_{t}^{(g)}}^{\top} \hat{\theta}^{(g)}_{i_{t},t}$
        \STATE Update client $i_{t}$: ${V}_{i_{t},t} \mathrel{+}= \bx_{t}^{(g)}{\bx_{t}^{(g)}}^{\top}$, ${b}_{i_{t},t} \mathrel{+}= \bx^{(g)}_{t} \hat{y}^{(g)}_{t}$, $\Delta{V}_{i_{t},t} \mathrel{+}= \bx_{t}^{(g)}{\bx_{t}^{(g)}}^{\top}$, $\Delta{b}_{i_{t},t}+=\bx^{(g)}_{t} \hat{y}^{(g)}_{t}$, ${V}^{(l)}_{i_{t},t} \mathrel{+}= \bx_{t}^{(l)}{\bx_{t}^{(l)}}^{\top}$, ${b}^{(l)}_{i_{t},t} \mathrel{+}= \bx^{(l)}_{t} \hat{y}^{(l)}_{t}$ 
        \STATE Event-triggered Communications (Algorithm \ref{algo:comm})
        % \IF{ Event $\cU_{t}(\gamma_{U})$ (Eq \eqref{eq:upload}) happens}
        %     \STATE Upload $\Delta{V}_{i_{t},t},\Delta{b}_{i_{t},t}$ ($i_{t} \rightarrow \text{server}$) 
        %     \STATE Update server: \newline
        %     % $V_{g,t}=V_{g,t-1}+\Delta V_{i_{t},t},b_{g,t}=b_{g,t-1}+\Delta b_{i_{t},t}$ \newline
        %     % $\Delta V_{-j,t}=\Delta V_{-j,t-1}+\Delta V_{i_{t},t},\Delta b_{-j,t}=\Delta b_{-j,t-1}+\Delta b_{i_{t},t}$ for $j \neq i_{t}$
        %     \hspace*{1em} $V_{g,t}\mathrel{+}=\Delta V_{i_{t},t},b_{g,t}\mathrel{+}=\Delta b_{i_{t},t}$ \newline
        %     \hspace*{1em} $\Delta V_{-j,t}\mathrel{+}=\Delta V_{i_{t},t},\Delta b_{-j,t}\mathrel{+}=\Delta b_{i_{t},t}$, $j \neq i_{t}$
        %     \STATE Client $i_{t}$ reset $\Delta{V}_{i_{t},t}=\textbf{0}$, $\Delta{b}_{i_{t},t}=\textbf{0}$

        %     \FOR{$j = 1, \dots, N$ and $j \neq i_{t}$}
        %         \IF{ Event $\cD_{t,j}(\gamma_{D})$ (Eq \eqref{eq:download}) happens}
        %             \STATE Download $\Delta{V}_{-j,t},\Delta{b}_{-j,t}$ ($\text{server} \rightarrow j$)
        %             \STATE Update client $j$: \newline
        %             % $V_{j,t}=V_{j,t-1}+\Delta V_{-j,t},b_{j,t}=b_{j,t-1}+\Delta b_{-j,t}$
        %             \hspace*{1em} $V_{j,t}\mathrel{+}=\Delta V_{-j,t},b_{j,t}\mathrel{+}=\Delta b_{-j,t}$
        %             \STATE Server reset $\Delta{V}_{-j,t}=\textbf{0},\Delta{b}_{-j,t}=\textbf{0}$
        %         \ENDIF
        %     \ENDFOR
        % \ENDIF
    \ENDFOR
  \end{algorithmic}
\end{algorithm}

% Existing works in bandit learning have explored different relationship models, such as unknown grouping structures \cite{gentile2014online} or smooth signals on a known user relation graph \cite{cesa2013gang}. In this paper, we consider a new relationship model: unknown parameter of client $i\in[N]$ consists of a globally shared component $\theta^{(g)}$ and a unique local component $\theta^{(i)}$, which is suitable for various problems in practice.
% \setlength{\textfloatsep}{0.1cm}
% \setlength{\floatsep}{0.15cm}
\noindent \textbf{Alternating Minimization}:
In a centralized learning setting, applying AM to iteratively update the estimation of the local component and global component is straightforward:
% The AM procedure in Algorithm 2 (line 10) is derived based on the following optimization problem:
% % Specifically, at time step $t$, we adopt AM method for the following optimization problem:
% \begin{align*}
%     & \min_{\theta^{(g)}, \{\theta^{(i)}\}_{i=1}^{N}} \sum_{i=1}^{N} \sum_{\tau \in \cN_{i}(t)} (y_{\tau}-\begin{bmatrix} \theta^{(g)} \\ \theta^{(i)} \end{bmatrix}^{\top} \begin{bmatrix} \bx_{\tau}^{(g)} \\ \bx_{\tau}^{(l)} \end{bmatrix} )^{2} \\
%     % + \frac{\lambda_{g}}{2} ||\theta^{(g)}||_{2}^{2} + \frac{\lambda_{l}}{2} \sum_{i=1}^{N} ||\theta^{(i)}||_{2}^{2} \\
%     & \text{s.t.} \quad ||\theta^{(g)}||_{2} \leq S_{g}, ||\theta^{(i)}||_{2} \leq S_{i}, \forall i\in [N]
% \end{align*}
% % Recall that $\cH_{i}(n)$ denotes the set of time steps up to time $n$ that client $i$ interacts with the environment.
% % Then by taking gradient w.r.t. $\theta^{(g)}$ and $\theta^{(i)}$ and set it to zero, we get the following analytical update rule:
% Assuming a centralized learning setting for a moment, by taking partial derivative w.r.t. each unknown parameter, we get the following analytical steps to alternatively update the estimates:
\begin{equation}\label{eq:centralized_AM}
\begin{split}
    &\hat{\theta}^{(l)}_{i,t} = \textit{proj}_{\mathbb{B}_{2}^{d_{i}}(1)} \bigl(({V}^{(l)}_{i,t})^{-}{b}^{(l)}_{i,t} \bigr), \forall i \in [N]  \\
    &\hat{\theta}^{(g)}_{t} = \textit{proj}_{\mathbb{B}_{2}^{d_{g}}(1)}\bigl(({V}_{t})^{-}{b}_{t}\bigr) 
\end{split}
\end{equation}
where $\mathbb{B}_{2}^{d}(1)=\{\theta \in \bR^{d}:||\theta||_{2} \leq 1\}$ denotes the unit $\ell_{2}$ ball, $(\cdot)^{-}$ denotes generalized matrix inverse, 
$\textit{proj}_{\Theta}(\cdot)$ denotes the Euclidean projection onto $L_{2}$ ball $\Theta$,
% $\textit{proj}_{\{\theta:||\theta||_{2} \leq S\}}(z)={z}/{\max{(||z||_{2}/S,1)}}$ denotes the Euclidean projection onto $L_{2}$ ball with radius $S$,
${V}^{(l)}_{i,t}=\sum_{\tau\in\cN_{i}(t)}\bx_{\tau}^{(l)}{\bx_{\tau}^{(l)}}^{\top}$, ${b}^{(l)}_{i,t}=\sum_{\tau\in \cN_{i}(t)}\bx_{\tau}^{(l)}(y_{\tau}-{\bx_{\tau}^{(g)}}^{\top}\hat{\theta}_{t}^{(g)})$,
${V}_{t}=\sum_{\tau=1}^{t}\bx_{\tau}^{(g)}{\bx_{\tau}^{(g)}}^{\top}$,
and 
${b}_{t}=\sum_{\tau=1}^{t}\bx_{\tau}^{(g)}(y_{\tau}-{\bx_{\tau}^{(l)}}^{\top}\hat{\theta}^{(l)}_{i_{\tau},t})$.
% \begin{itemize}
% \item Fix $\hat{\theta}^{(l)}_{i,t}$, for $i=1,\dots,N$, and update $\hat{\theta}^{(g)}_{t}$
% \begin{equation}\label{eq:am_central_g}
% \hat{\theta}^{(g)}_{t} = \textit{proj}_{\{\theta:||\theta||_{2} \leq S_{g}\}}\bigl(({V}_{t})^{-}{b}_{t}\bigr)
% % \hat{\theta}^{(g)}_{n} & = \textit{proj}_{\{\theta:||\theta||_{2} \leq S_{g}\}} (\hat{\theta}^{(g)}_{n})=  \frac{\hat{\theta}^{(g)}_{n}}{\max{(||\hat{\theta}^{(g)}_{n}||_{2}/S_{g},1)}}
% \end{equation}
% where ${V}_{t}=\sum_{\tau=1}^{t}\bx_{\tau}^{(g)}{\bx_{\tau}^{(g)}}^{\top}$ and ${b}_{t}=\sum_{\tau=1}^{t}\bx_{\tau}^{(g)}(y_{\tau}-{\bx_{\tau}^{(l)}}^{\top}\hat{\theta}^{(l)}_{i_{\tau},t})$. Note that $(\cdot)^{-}$ denotes generalized inverse, and $\textit{proj}$ is the projection onto $L_{2}$ ball $\textit{proj}_{\{\theta:||\theta||_{2} \leq S\}}(z)=\frac{z}{\max{(||z||_{2}/S,1)}}$.
% \item Fix $\hat{\theta}^{(g)}_{t}$ and update $\hat{\theta}^{(i)}_{t}$, for $i=1,\dots,N$
% \begin{equation}\label{eq:am_central_l}
% \hat{\theta}^{(l)}_{i,t} = \textit{proj}_{\{\theta:||\theta||_{2} \leq S_{i}\}} \bigl(({V}^{(l)}_{i,t})^{-}{b}^{(l)}_{i,t} \bigr) 
% % \hat{\theta}^{(i)}_{n} & = \textit{proj}_{\{{\theta}^{(i)}_{n}:||{\theta}^{(i)}_{n}||_{2} \leq S_{l}\}} (\hat{\theta}^{(i)}_{n}) = \frac{\hat{\theta}^{(i)}_{n}}{\max{(||\hat{\theta}^{(i)}_{n}||_{2}/S_{l},1)}}
% \end{equation}
% where ${V}^{(l)}_{i,t}=\sum_{\tau\in\cN_{i}(t)}\bx_{\tau}^{(l)}{\bx_{\tau}^{(l)}}^{\top}$ and ${b}^{(l)}_{i,t}=\sum_{\tau\in \cN_{i}(t)}\bx_{\tau}^{(l)}(y_{\tau}-{\bx_{\tau}^{(g)}}^{\top}\hat{\theta}_{t}^{(g)})$. 
% \end{itemize}

However, iteratively executing Eq \eqref{eq:centralized_AM} is impractical under federated setting: first, $\{{V}_{t},{b}_{t}\}$ are distributed across the clients; second, iteratively updating $b_{t}$ and $b^{(l)}_{i,t}$ requires storage of raw history data. 
This incurs space and communication complexities that are linear in time $T$.
% This makes both the space and communication complexities grow linearly with time $T$.
% Though the sufficient statistics for estimating $\theta^{(i)}$, i.e., $\{{V}^{(l)}_{i,t},{b}^{(l)}_{i,t}\}$, are readily available at client $i$ to execute Eq \eqref{eq:am_central_l}, those for estimating $\theta^{(g)}$, i.e., $\{{V}_{t},{b}_{t}\}$, are distributed across the clients. 
% Therefore, executing Eq \eqref{eq:am_central_g} requires communication. 
% Therefore, the AM steps in Eq \eqref{eq:am_central_g} and \eqref{eq:am_central_l} require not only storage of raw history data, but also repeated communications to alternatively update $b_{t}$ and $b^{(l)}_{i,t}$. This makes both the space and communication complexities grow linearly with time $T$, which is unacceptable for federated linear bandit.
% Taking these into consideration, we use the following AM steps for the local update in each client (line 10 in Algorithm \ref{algo:AsyncLinUCB_AM}). 
Instead, we modify Eq \eqref{eq:centralized_AM} to get the following update rule. At time $t$, after client $i_{t}$ obtains a new data point $(\bx_{t},y_{t})$ from the environment, it alternates between the following two steps (line 9):
\begin{equation}\label{eq:AM}
\begin{split}
    \hat{\theta}^{(l)}_{i_{t},t} &= \textit{proj}_{\mathbb{B}_{2}^{d_{i}}(1)}\left(({V}^{(l)}_{i_{t},t-1}+\bx_{t}^{(l)}{\bx_{t}^{(l)}}^{\top})^{-}({b}^{(l)}_{i_{t},t-1}+\bx_{t}^{(l)}\hat{y}^{(l)}_{t})\right)\\
    \hat{\theta}^{(g)}_{i_{t},t} &= \textit{proj}_{\mathbb{B}_{2}^{d_{g}}(1)}\left((V_{i_{t},t-1}+\bx_{t}^{(g)}{\bx_{t}^{(g)}}^{\top})^{-}({b}_{i_{t},t-1}+\bx_{t}^{(g)}\hat{y}^{(g)}_{t})\right)
\end{split}
\end{equation}
%\small
%\begin{equation}\label{eq:AM}
%\begin{split}
%    &\begin{cases}
%    \hat{\theta}^{(l)}_{i_{t},t} = ({V}^{(l)}_{i_{t},t-1}+\bx_{t}^{(l)}{\bx_{t}^{(l)}}^{\top})^{-}({b}^{(l)}_{i_{t},t-1}+\bx_{t}^{(l)}\hat{y}^{(l)}_{t})  \\
%    \hat{\theta}^{(l)}_{i_{t},t} = \textit{proj}_{\{\theta:||\theta||_{2} \leq S_{i_{t}}\}}(\hat{\theta}^{(l)}_{i_{t},t})
%    \end{cases} \\
%    &\begin{cases}
%    \hat{\theta}^{(g)}_{i_{t},t} = (V_{i_{t},t-1}+\bx_{t}^{(g)}{\bx_{t}^{(g)}}^{\top})^{-}({b}_{i_{t},t-1}+\bx_{t}^{(g)}\hat{y}^{(g)}_{t})  \\
%    \hat{\theta}^{(g)}_{i_{t},t} = \textit{proj}_{\{\theta:||\theta||_{2} \leq S_{g}\}}(\hat{\theta}^{(g)}_{i_{t},t})
%    \end{cases}
%\end{split}
%\end{equation}
%\normalsize
where $\hat{y}^{(l)}_{t}=y_{t}-{\bx_{t}^{(g)}}^{\top} \hat{\theta}^{(g)}_{i_{t},t}$ and $\hat{y}^{(g)}_{t}=y_{t}-{\bx_{t}^{(l)}}^{\top} \hat{\theta}^{(l)}_{i_{t},t}$ denote the estimated `partial' rewards for $\theta^{(i)}$ and $\theta^{(g)}$, and $\{{V}_{i,t-1},{b}_{i,t-1}\}$ denote client $i$'s local copy of the sufficient statistics for $\theta^{(g)}$. Then $\hat{y}^{(g)}_{t}$ and $\hat{y}^{(l)}_{t}$ are used to locally update the sufficient statistics of client $i_{t}$ (line 10 in Algorithm \ref{algo:AsyncLinUCB_AM}). Compared with Eq \eqref{eq:centralized_AM} that iteratively updates the estimated `partial' rewards on all historic data from different clients, Eq \eqref{eq:AM} only updates that on $(\bx_{t},y_{t})$ while keeping the rest fixed. This incurs constant space complexity and no communication cost. Though this comes at a price of slower convergence on the estimators, we later prove that this will not sacrifice the accumulative regret too much.
% , i.e., another trade-off.

Since the local component is unique to each client, only $\{{V}_{i,t-1},{b}_{i,t-1}\}$ for estimating the global component need to be shared among clients. Our asynchronous communication protocol can be directly applied here (line 11) to ensure communication efficiency. 
% and for simplicity, we reuse the previous notations to denote the `upload' and `download' buffers in Algorithm \ref{algo:AsyncLinUCB_AM}, so the descriptions in Algorithm \ref{algo:comm} can be directly used here.

% at time $t$, client $i$'s `upload' buffer is denoted by $\Delta{V}_{i,t-1},\Delta{b}_{i,t-1}$, aggregated sufficient statistics stored on the server is denoted by ${V}_{g,t-1},{b}_{g,t-1}$, and the `download' buffer for each client $j\in[N]$ is denoted by $\Delta{V}_{-j,t-1},\Delta{b}_{-j,t-1}$.

% To facilitate discussion about the communication of the sufficient statistics for estimating $\theta^{(g)}$, we slightly modify our previous notations.
% Specifically, at time $t$, client $i$'s local copy of the sufficient statistics for estimating the global component are denoted by: ${V}^{(g)}_{i,t-1},{b}_{i,t-1}^{(g)}$, and its `upload' buffer is denoted by $\Delta{V}^{(g)}_{i,t-1},\Delta{b}_{i,t-1}^{(g)}$. The aggregated sufficient statistics stored on the server is denoted by ${V}^{(g)}_{g,t-1},{b}_{g,t-1}^{(g)}$, and the `download' buffer for each client $j\in[N]$ is denoted by $\Delta{V}^{(g)}_{-j,t-1},\Delta{b}_{-j,t-1}^{(g)}$.

\noindent \textbf{Arm selection}:
Client $i_{t}$ selects arm $\bx_{t} \in \cA_{t}$ via the UCB strategy (line 8 in Algorithm \ref{algo:AsyncLinUCB_AM}). Confidence ellipsoids for the estimations obtained by Eq \eqref{eq:AM} are given in Lemma \ref{lem:confidence_ellipsoid}. Now the UCB score consists of two terms corresponding to the global and local component estimations, respectively:
\begin{equation}\label{eq:UCB_AM}
    \bx_{t}=\argmax_{\bx \in \cA_{t}}{\text{UCB}^{(g)}_{i_{t},t-1}({\bx^{(g)}})+\text{UCB}^{(l)}_{i_{t},t-1}({\bx^{(l)}})}
\end{equation}
where 
$\text{UCB}^{(g)}_{i_{t},t-1}({\bx^{(g)}}) = {\bx^{(g)}}^{\top}\hat{\theta}^{(g)}_{i_{t},t-1}(\lambda)+\alpha^{(g)}_{i_{t},t-1}||\bx^{(g)}||_{V_{i_{t},t-1}(\lambda)^{-1}}$,
$\text{UCB}^{(l)}_{i_{t},t-1}({\bx^{(l)}}) = {\bx^{(l)}}^{\top}\hat{\theta}^{(l)}_{i_{t},t-1}(\lambda)+\alpha^{(l)}_{i_{t},t-1}||\bx^{(l)}||_{V^{(l)}_{i_{t},t-1}(\lambda)^{-1}}$. $\hat{\theta}^{(g)}_{i_{t},t-1}(\lambda)=V_{i_{t},t-1}(\lambda)^{-1}b_{i_{t},t-1}$ is the ridge regression estimator for the global component with the regularization parameter $\lambda$. $\hat{\theta}^{(l)}_{i_{t},t-1}(\lambda)= V^{(l)}_{i_{t},t-1}(\lambda)^{-1}b^{(l)}_{i_{t},t-1}$ is the ridge regression estimator for the local component with the regularization parameter $\lambda$.
$\alpha^{(g)}_{i_{t},t-1}$ and $\alpha^{(l)}_{i_{t},t-1}$ are given in Lemma \ref{lem:confidence_ellipsoid} (proof given in appendix).

\begin{lemma}[Confidence ellipsoids]\label{lem:confidence_ellipsoid}
With probability at least $1-\delta$, $||\hat{\theta}^{(g)}_{i,t}(\lambda) - \theta^{(g)}||_{V_{i,t}(\lambda)} \leq \alpha^{(g)}_{i,t}$, 
where $\alpha^{(g)}_{i,t} = (\sigma+2) \sqrt{\log{\frac{\det{V_{i,t}(\lambda)}}{\det{\lambda I}}}+2\log{1/\delta}}+\sqrt{\lambda}$. 
% where $\alpha^{(g)}_{i,t} = (\sigma+2L_{l}S_{l}) \sqrt{2\ln{\frac{\det{(V_{i,t}(\lambda) )^{1/2}}}{\det{(\lambda I)^{1/2}}\delta}}}+\sqrt{\lambda}S_{g}$. 
And with probability at least $1-\delta$, $||\hat{\theta}^{(l)}_{i,t}(\lambda) - \theta^{(i)}||_{V^{(l)}_{i,t}(\lambda)} \leq \alpha^{(l)}_{i,t}$, 
where $\alpha^{(l)}_{i,t} =  (\sigma+2) \sqrt{\log{\frac{\det{V^{(l)}_{i,t}(\lambda)}}{\det{\lambda I}}+2\log{1/\delta}}}+\sqrt{\lambda}$.
% where $\alpha^{(l)}_{i,t} =  (\sigma+2L_{g}S_{g}) \sqrt{2\ln{\frac{\det{(V^{(l)}_{i,t}(\lambda))^{1/2}}}{\det{(\lambda I)^{1/2}}\delta}}}+\sqrt{\lambda}S_{l}$.
% \small
% \begin{align*}
%     \alpha^{(i)}_{i,t} =  (\sigma+\frac{L_{l}S_{l}}{2}) \sqrt{2\ln{\frac{\det{(V^{(i)}_{i,t})^{1/2}}\det{(\lambda_{l} I)^{-1/2}}}{\delta}}}+\sqrt{\lambda_{l}}S_{l}
% \end{align*}
% \normalsize
% where $V^{(i)}_{n}=\sum_{t\in\cH_{i}(n)}\bx_{t}^{(l)}\bx_{t}^{(l)}^{\top} + \lambda_{l} \bI_{d_{l}}$, we denote this confidence set as $C_{n}^{(i)}$.
\end{lemma}

Then based on the constructed confidence ellipsoid, we can prove Theorem \ref{thm:regret_communication_async-linucb-am} (proof given in appendix), which provides the upper bounds of cumulative regret $R_{T}$ and communication cost $C_{T}$ incurred by \modeltwo{}.
\begin{theorem}[Regret and Communication Cost] \label{thm:regret_communication_async-linucb-am}
With Assumption \ref{assump:context_diversity}, and the communication thresholds $\gamma_{U},\gamma_{D}$, then w.h.p., the cumulative regret $$R_{T} = O\bigl(d_{g}\sqrt{T\log^{2}{T}}\min(\sqrt{N},\sqrt{\gamma_{D} [1+(N-1)(\gamma_{U}-1)]} + \sum_{i=1}^{N} d_{i} \sqrt{|\cN_{i}(T)|\log^{2}{|\cN_{i}(T)|}}\bigr)$$ and communication cost $$C_{T}=O(d_{g} N\log{T}/\log{\min(\gamma_{U},\gamma_{D})})$$
% $R_{T}$ has upper bound
% \small
% $$R_{T} = O\left(d\sqrt{T\log^{2}{T}}\min(\sqrt{N},\sqrt{\gamma_{D} [1+(N-1)(\gamma_{U}-1)]})\right)$$
% \normalsize
% The communication cost $C_{T}$ has upper bound
% $$C_{T} =\sum_{i=1}^{N} C_{T,i} \leq N \frac{\log{\det({V}_{T-1})}-d \log{\lambda}}{\log{\min{(\gamma_{U},\gamma_{D})}}}$$
\end{theorem}
Note that the regret upper bound of \modeltwo{} consists of two terms: the first term corresponds to the global components $\theta^{(g)}$, which enjoys the benefit of communication; and the second term corresponds to the unique local components $\theta^{(i)}$ of each client, which essentially matches the regret upper bound for running $N$ LinUCB independently for $\theta^{(i)}$ of each client. 
Intuitively, when the problems solved by different clients become more similar, then the first term dominates as $d_{g}$ becomes larger compared with $d_{i}$.

\section{Experiments}\label{sec:exp}
We performed extensive empirical evaluations of \modelone{} and \modeltwo{} on both synthetic and real-world datasets (we set $\gamma_{U}=\gamma_{D}=\gamma$ in all experiments), and included \modelbaseline{} \citep{wang2019distributed} as baseline. 
% Due to the space limit, we provide a brief summary of the experiment setup and results. Detailed descriptions and discussions are presented in the appendix.
\subsection{Experiments on Synthetic Dataset}
To validate our theoretical analysis in Section \ref{subsec:async_LinUCB} and Section \ref{subsec:async_LinUCB_AM}, two sets of simulation experiments were conducted.
We first conducted simulation experiment in homogeneous client setting to
% compare how well the algorithms can balance regret $R_{T}$ and communication cost $C_{T}$ under uniform and non-uniform client distributions, as well as 
validate our theoretical comparison between \modelone{} and \modelbaseline{} (see Section \ref{subsec:async_LinUCB} and Section \ref{sec:tb_theretical} in appendix), i.e., how well the algorithms balance regret $R_{T}$ and communication cost $C_{T}$ under uniform and non-uniform client distributions. Then we conducted simulation experiment in heterogeneous setting to validate our regret upper bound for \modeltwo{} (see Section \ref{subsec:async_LinUCB_AM}), i.e., how the portion of global components $\frac{d_{g}}{d_{g}+d_{i}}$ in the bandit parameter affects the regret of \modeltwo{}.
\subsubsection{Synthetic dataset.} We simulated the federated linear bandit problem setting in Section \ref{subsec:problem_formulation}, with $T=30000, N=1000$, and $\cA_{t}$ ($K=25$) uniformly sampled from a $\ell_2$ ball.
% is sampled from a pool of $1000$ randomly generated arms. 
% Two sets of simulation experiments were performed, with 
(1) Homogeneous clients: 
To compare how the algorithms balance $R_{T}$ and $C_{T}$ under uniform ($P(i_{t}=i)=\frac{1}{N},\forall i,t$) and non-uniform client distributions ($P(i_{t})$ is an arbitrary point on probability simplex), we fixed $d=25$, and ran \modelone{} and \modelbaseline{} with a large range of threshold values (logarithmically spaced between $10^{-2}$ and $10^{3}$).
% To compare algorithms' regret and communication cost under different trade-off settings over time, we fix $d=25$, $L=S=1$, and run \modelone{} with $\gamma \in \{1,2,5,8,+\infty\}$, \modelbaseline{} with threshold $D \in \{\frac{T}{d N^{2} \log{T}}, \frac{T}{d N^{1.5} \log{T}}\}$ (see Remark \ref{rmk:regret}) with both uniformly and non-uniformly sampled clients.
(2) Heterogeneous clients: To see how $R_{T}$ and $C_{T}$ of \modeltwo{} changes as the portion of global components change, we set the local components for all clients to have equal dimension, i.e., $d_{i}=d_{l},\forall i \in [N]$, fixed $d_{g}+d_{l}=25$, $\forall i\in [N]$, and then ran \modeltwo{} (with $\gamma=5$) under varying $d_{g} \in \{4,8,12,16,20,24\}$.

\subsubsection{Experiment results.}
\begin{figure}
\centering     %%% not \center
\subfigure[Homogeneous (uniform client distribution)]{\label{fig:a}\includegraphics[width=0.55\textwidth]{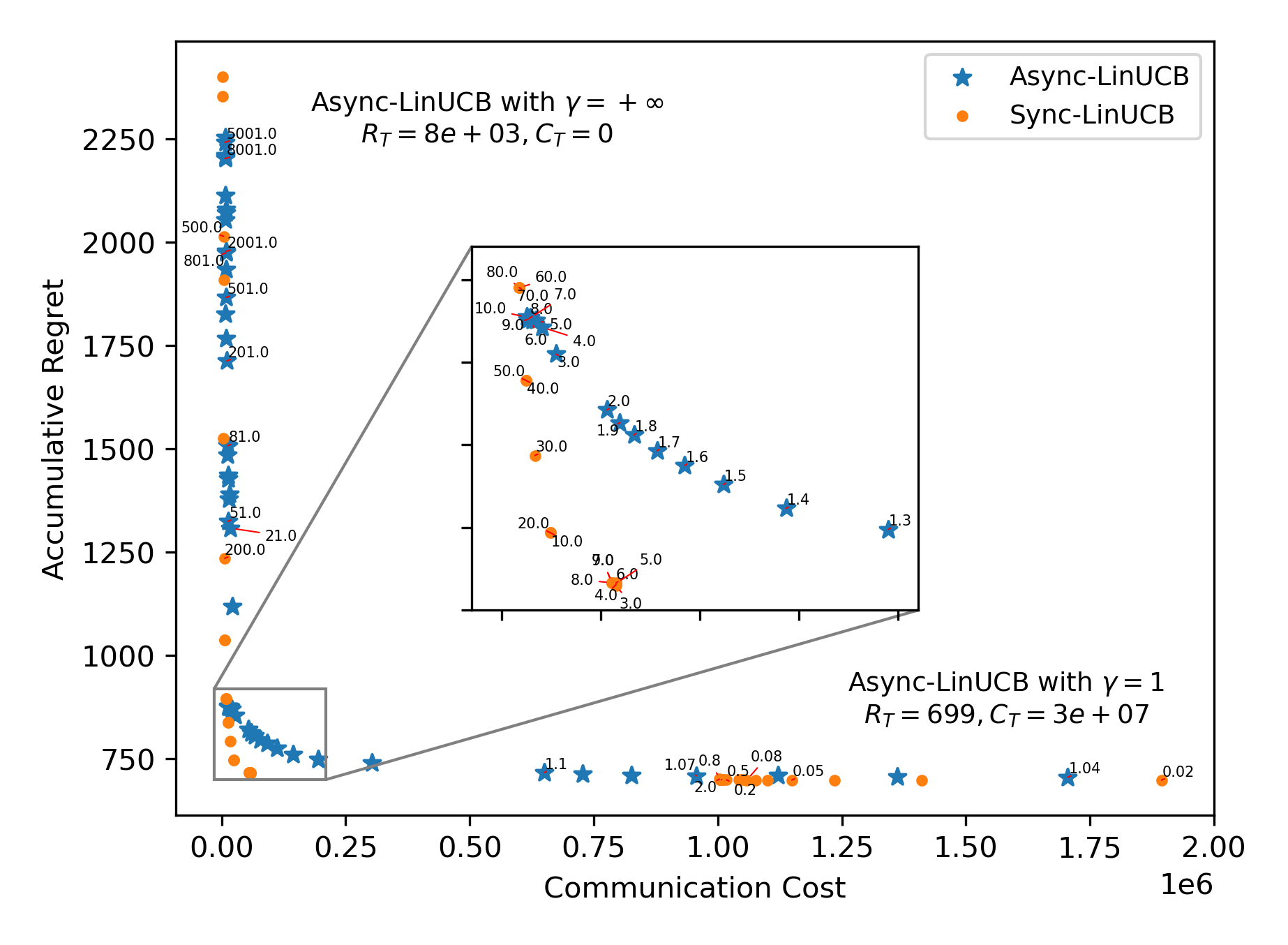}}
\vspace{-1mm}
\subfigure[Homogeneous (non-uniform client distribution)]{\label{fig:b}\includegraphics[width=0.55\textwidth]{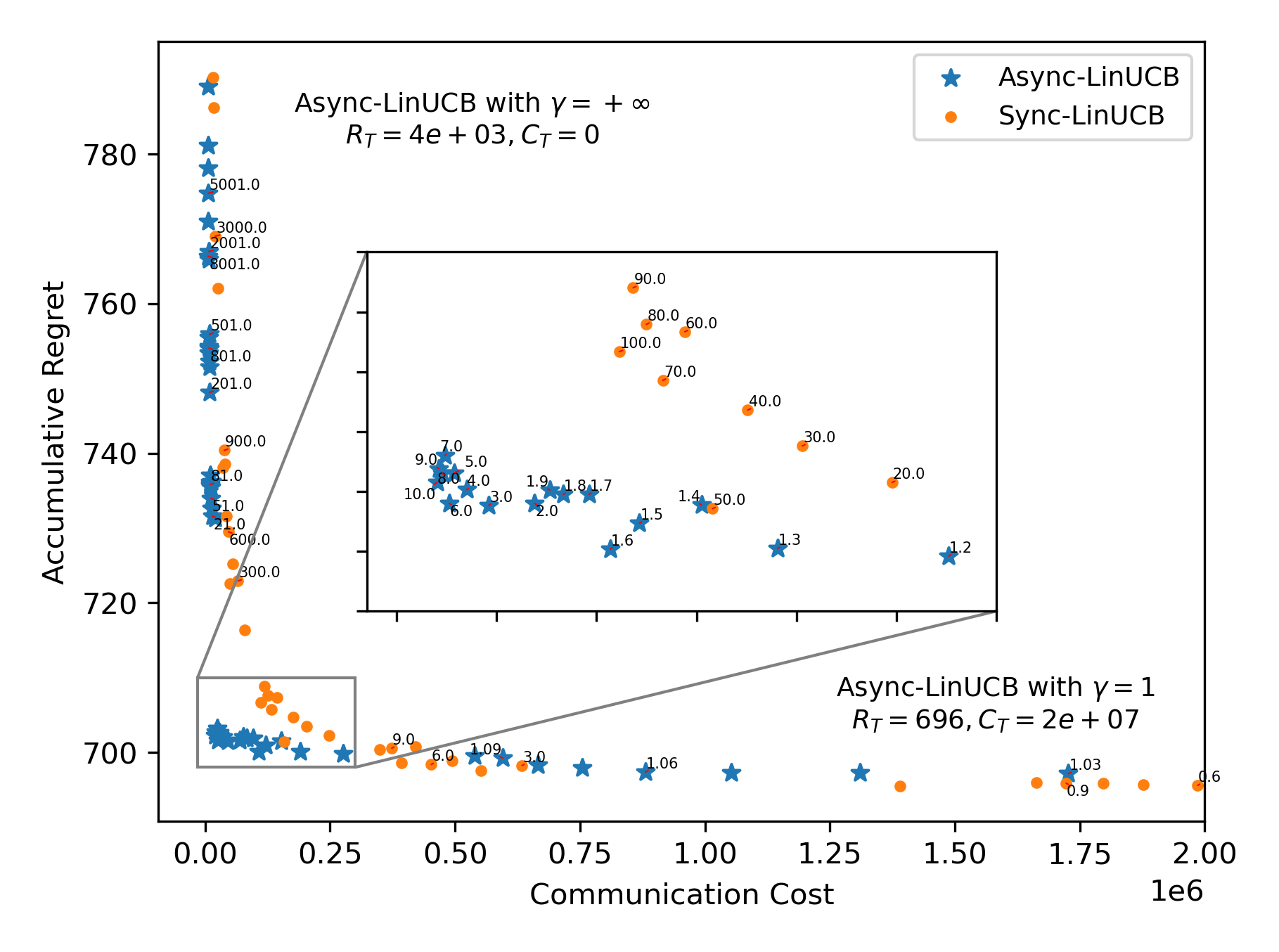}}
\vspace{-1mm}
%\medskip
\subfigure[Heterogeneous clients]{\label{fig:c}\includegraphics[width=0.55\textwidth]{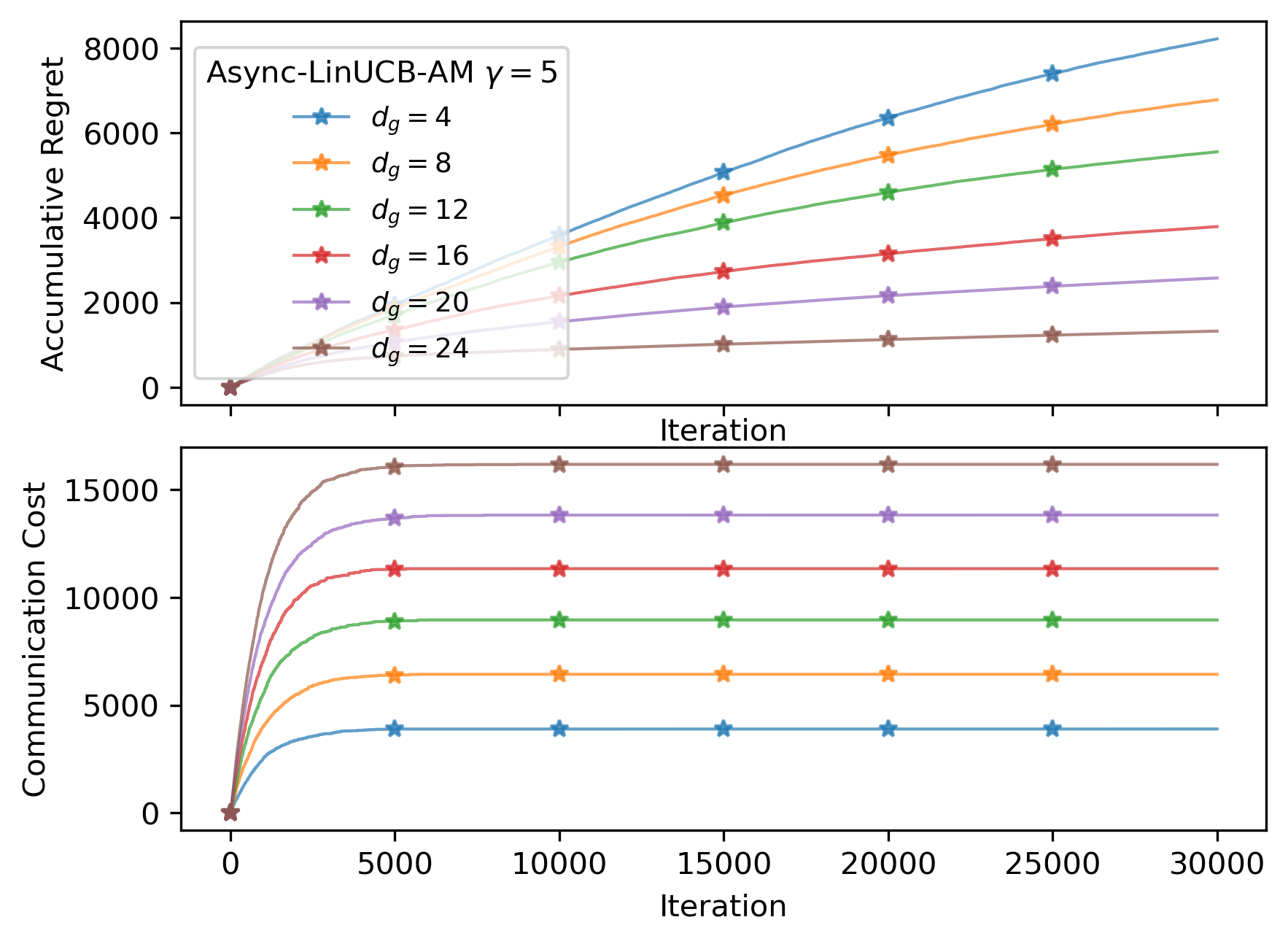}}
% \vspace{-1mm}
% \subfigure[LastFM ($N=1892$)]{\label{fig:d}\includegraphics[width=0.48\textwidth]{imgs/regretVScommCost_lastfm_noclutter.png}}
% \medskip
% \subfigure[Delicious ($N=1867$)]{\label{fig:e}\includegraphics[width=0.48\textwidth]{imgs/regretVScommCost_delicious_noclutter.png}}
% \subfigure[MovieLens ($N=54$)]{\label{fig:f}\includegraphics[width=0.48\textwidth]{imgs/regretVScommCost_movielens_noclutter.png}}
\vspace{-1mm}
\caption{Experiment results on synthetic dataset.}
\end{figure}

Experiment results (averaged over $10$ runs) on synthetic dataset are shown in Figure
% \ref{fig:exp_result}(a)-(c):
\ref{fig:a}-\ref{fig:c}. 
Note that in the scatter plots, 
% the x-axis is the total communication cost at iteration $T$ and the y-axis is the corresponding accumulative regret/reward at iteration $T$. 
each dot denotes the cumulative communication cost (x-axis) and regret (y-axis) that an algorithm (\modelone{} or \modelbaseline{}) with certain threshold value (labeled next to the dot) has obtained at iteration $T$.

(1) Homogeneous clients (Figure \ref{fig:a}-\ref{fig:b}): 
From both Figure \ref{fig:a} and Figure \ref{fig:b}, we can see that as the threshold value increases, $C_{T}$ decreases and $R_{T}$ increases, and that the use of event-triggered communication significantly reduces $C_{T}$ while attaining low $R_{T}$, compared with synchronizing all the clients at each time step (\modelone{} with $\gamma=1$). In Figure \ref{fig:a}, \modelbaseline{} has lower $C_{T}$ than \modelone{} under the same $R_{T}$, and in Figure \ref{fig:b}, \modelone{} has lower $C_{T}$ than \modelbaseline{} under the same $R_{T}$, which conform with our theoretical results that \modelbaseline{} has inefficient communication under non-uniform client distribution.
% Among all the algorithms compared, \modelone{} with $\gamma=5$ and $\gamma=8$ and \modelbaseline{} with $D=\frac{T}{d \log{T}}$ strike a good balance between regret and communication cost. The other algorithms either incur a very high regret, i.e. \modelone{} with $\gamma=+\infty$ (as no communication can be triggered), or incur too much communication cost, i.e., \modelone{} with $\gamma=1$ and \modelbaseline{} with $D=\frac{T}{d N \log{T}}$ (as they always communicate). And almost in all results, both synthetic and real-world datasets, always communicating (i.e., \modelone{} with $\gamma=1$) costs significant overhead in communication, while it does not necessarily lead to an obvious advantage in regret. This suggests the necessity and benefit of our event-triggered communication control. In addition, the results for \modelbaseline{} with $D=\frac{T}{d N \log{T}}$ and $D=\frac{T}{d \log{T}}$ conform with our theoretical analysis in Remark \ref{rmk:regret_comm}, e.g., it incurs higher regret when matching the communication cost with \modelone{} or higher communication cost when matching the regret with \modelone{}. 
% \modelone{} with $\gamma=1$ and \modelbaseline{} with $D=\frac{T}{d N \log{T}}$ attain the lowest regret, but also incur much higher communication costs compared with other algorithms.

(2) Heterogeneous clients (Figure \ref{fig:c}): By increasing the portion of global components $\theta^{g}$ in the bandit parameter, we can observe a clear trend in both regret and communication cost, i.e., the regret keeps decreasing while the communication cost keeps increasing. This validates our theoretical analysis about $R_{T}$ and $C_{T}$ in Section \ref{subsec:async_LinUCB_AM}. With $d_{g}$ increases and $d_{l}$ decreases, the first term in the upper bound of $R_{T}$ dominates (which grows slower w.r.t. $N$ compared with the second term), leading to the decreased regret, but the communication cost would increase since $C_{T}=O(d_{g} N \log{T})$.

\subsection{Experiments on Real-world Dataset}
% From the experiments on synthetic dataset, we have validated our upper bounds on the regret and communication cost of \modelone{} and \modeltwo{}. 
%However, we should note that our theoretical results are based on a slightly stronger assumption on the context vectors (see Assumption \ref{assump:context_diversity}), compared with \modelbaseline{}. 
%Therefore, we want to validate whether such an assumption is reasonable in practice, i.e., whether our algorithms can still work properly on the real-world datasets that contain TF-IDF feature vectors extracted from the tags and metadata of the items. 
We continue investigating the effectiveness of our proposed solution on real-world datasets. Note that these real-world datasets do not necessarily satisfy the assumption that all the clients are homogeneous, in other words not all the users have the same preference, we pay special attention to \modeltwo{} in the comparison, by setting $x_{g} \equiv x_{i}, \forall i \in [N]$, as mentioned in Section \ref{subsec:problem_formulation}. This allows the clients to learn a global model collaboratively, and in the meantime each learns a personalized model independently. Intuitively, this should make \modeltwo{} more robust to different settings, i.e., the clients are either homogeneous or heterogeneous.
\subsubsection{Real-world dataset.}
We compared \modelone{}, \modeltwo{} and \modelbaseline{} on three public recommendation datasets: LastFM, Delicious and MovieLens \citep{Cantador:RecSys2011,harper2015movielens}, with various threshold values (logarithmically spaced between $10^{-2}$ and $10^{3}$). 
The LastFM dataset contains $N=1892$ users, 17632 items (artists), and $T=96733$ interactions. We consider the ``\textit{listened artists}'' in each user as positive feedback. The Delicious dataset contains $N=1861$ users, 69226 items (URLs), and $T=104799$ interactions. We treat the bookmarked URLs in each user as positive feedback. 
The MovieLens dataset used in the experiment is extracted from the MovieLens 20M dataset by keeping users with over $3000$ observations, which results in a dataset with $N=54$ users, 26567 items (movies), and $T=214729$ interactions. We consider all items with non-zero ratings as positive feedback. The datasets were preprocessed following the procedure in \cite{cesa2013gang} to fit the linear bandit setting (with TF-IDF feature $d=25$ and arm set $K=25$).

\subsubsection{Experiment results.}
% \textbf{Results on real-world datasets}:
%First, from the results on all three datasets (Figure \ref{fig:d}-\ref{fig:f}), the performance of \modelone{} is as good as that of \modelbaseline{}, which indicates Assumption \ref{assump:context_diversity} is still reasonable with the TF-IDF feature vectors. Especially in the results on MovieLens dataset (Figure \ref{fig:f}), whose data conforms with our homogeneous clients assumption as we will see below, we can see the dots corresponding to \modelone{} and \modelbaseline{} have very similar patterns as that in the ideal simulation environment (Figure \ref{fig:a}).

\begin{figure}
\centering     %%% not \center
% \subfigure[Homogeneous (uniform)]{\label{fig:a}\includegraphics[width=0.48\textwidth]{imgs/regretVScommCost_uniform_noclutter.png}}
% \subfigure[Homogeneous (non-uniform)]{\label{fig:b}\includegraphics[width=0.48\textwidth]{imgs/regretVScommCost_nonuniform_noclutter.png}}
% % \vspace{-1mm}
% \medskip
% \subfigure[Heterogeneous clients]{\label{fig:c}\includegraphics[width=0.48\textwidth]{imgs/sim_hetero_30000.png}}
\vspace{-2mm}
\subfigure[LastFM ($N=1892$)]{\label{fig:d}\includegraphics[width=0.55\textwidth]{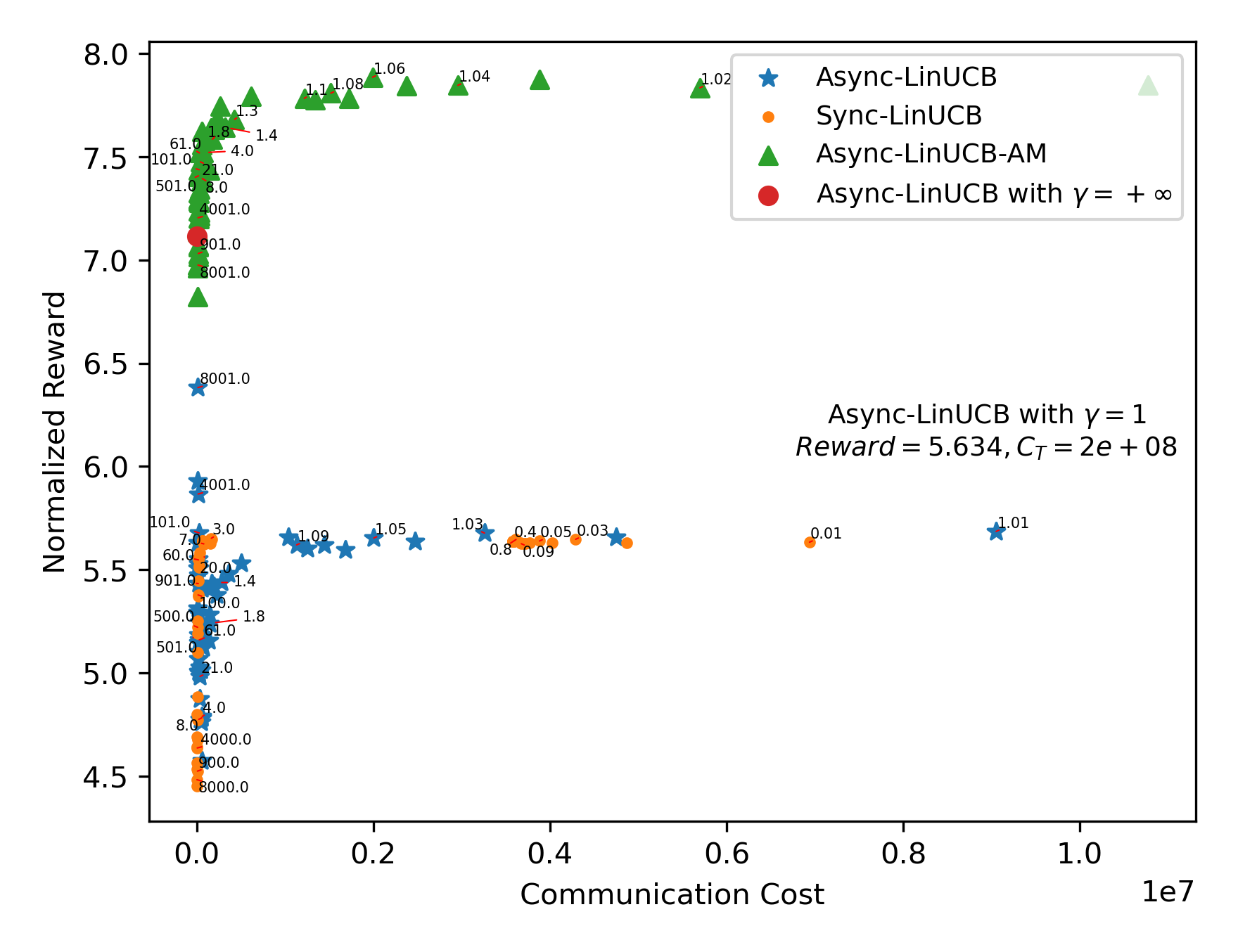}}
\vspace{-2mm}
\subfigure[Delicious ($N=1867$)]{\label{fig:e}\includegraphics[width=0.55\textwidth]{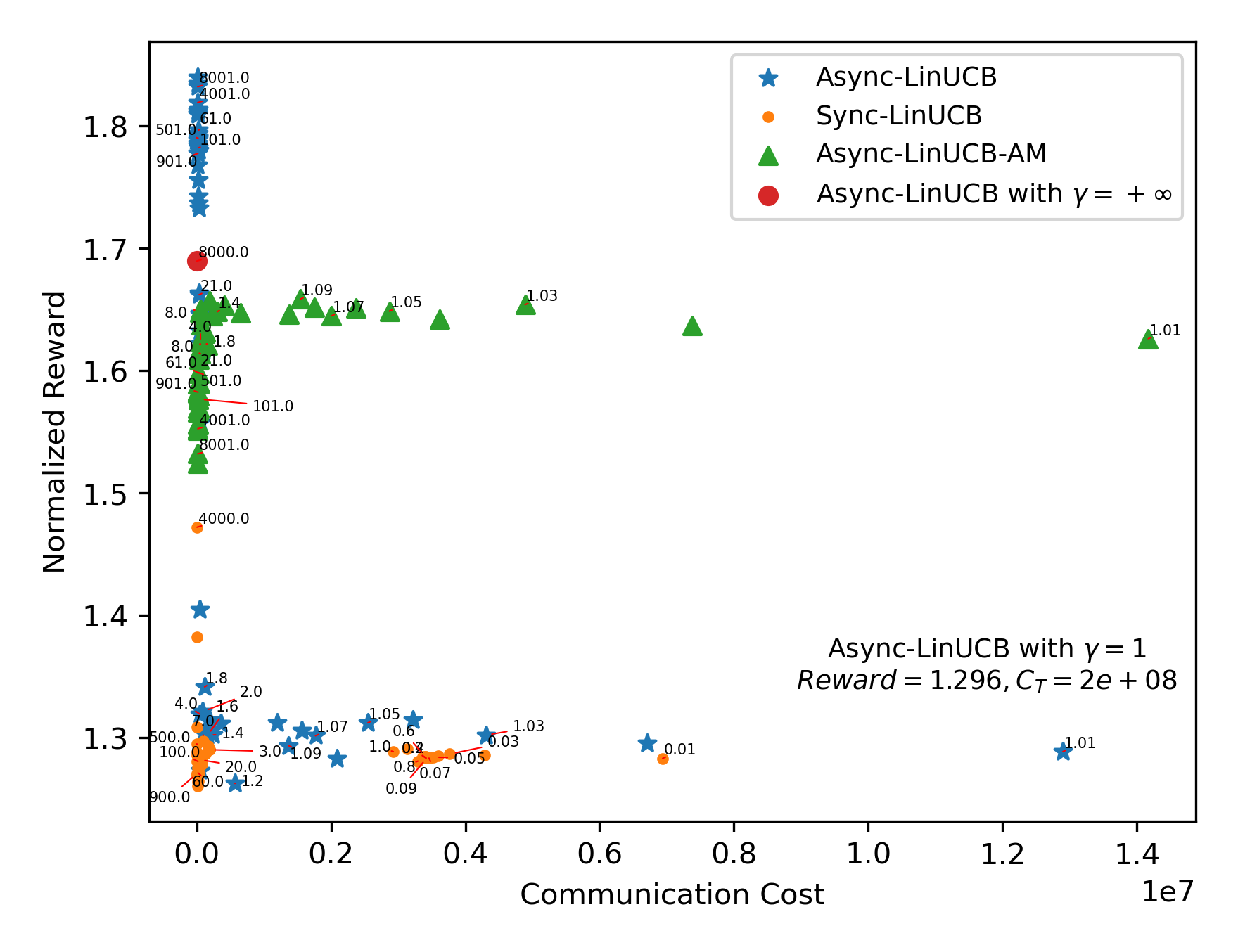}}
\vspace{-2mm}
\subfigure[MovieLens ($N=54$)]{\label{fig:f}\includegraphics[width=0.55\textwidth]{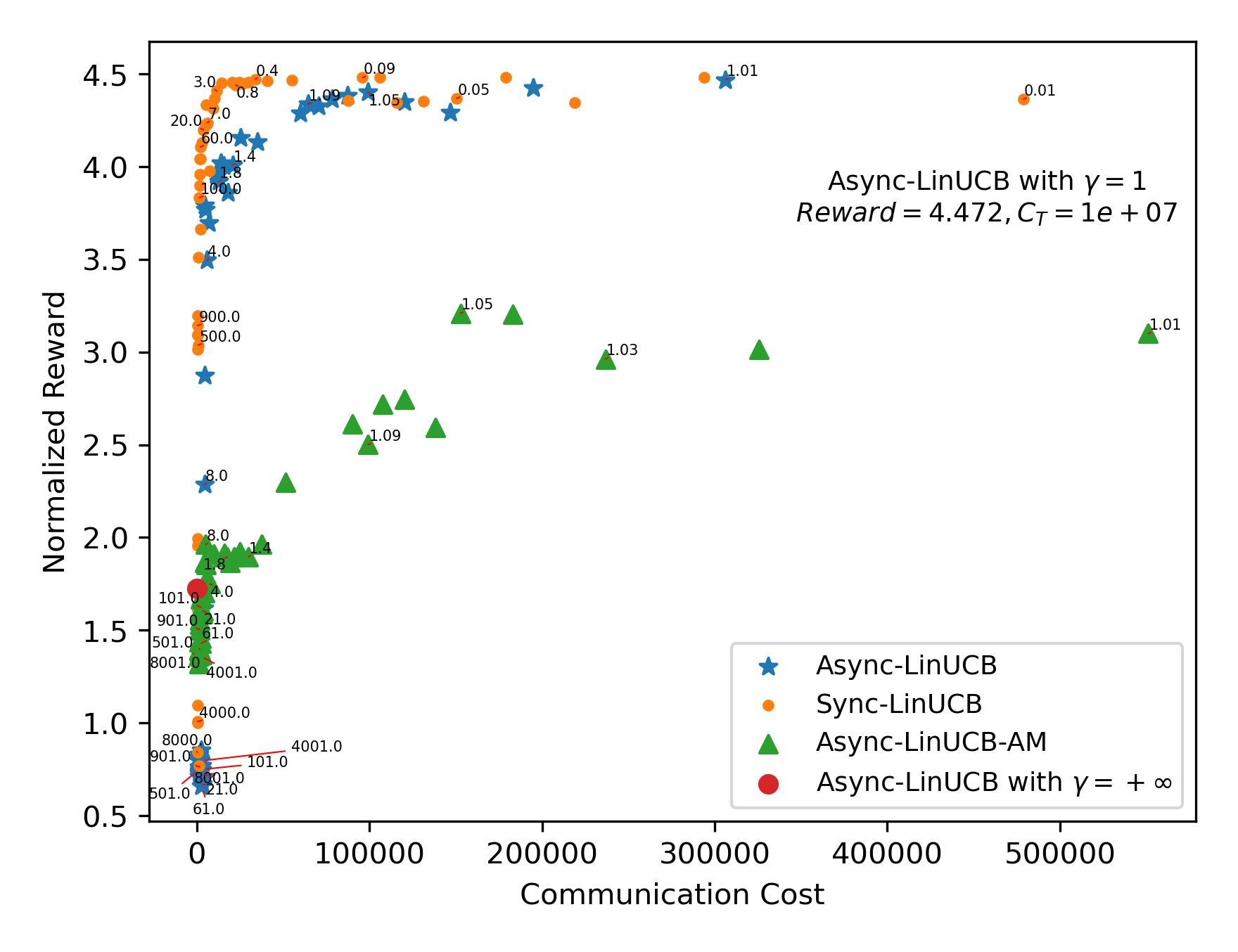}}
% \vspace{-1mm}
\caption{Experiment results on real-world recommendation datasets.}
\end{figure}

Experiment results on the three real-world datasets are shown in Figure \ref{fig:d}-\ref{fig:f}. 
In the scatter plots, each dot denotes the cumulative communication cost (x-axis) and normalized reward by a random strategy (y-axis) that an algorithm (\modelone{}, \modeltwo{}, or \modelbaseline{}) with certain threshold value (labeled next to the dot) has obtained at iteration $T$.
To understand the results of these algorithms on the three real-world datasets, we can first look at how well the two extreme cases, \modelone{} with $\gamma=1$ (as the communication cost of this algorithm is outside of the figure, its result is illustrated as text label) and \modelone{} with $\gamma=+\infty$ perform. 

(1) LastFM \& Delicious (Figure \ref{fig:d}-\ref{fig:e}): 
On both LastFM and Delicious datasets, \modelone{} with $\gamma=+\infty$ (illustrated as the red dot) attains very high reward, which suggests users in these two datasets have very diverse preferences, such that aggregating their data has a negative impact on the performance. 
Since the homogeneous clients assumption does not hold in this case, both \modelone{} and \modelbaseline{} perform as badly as the extreme case of \modelone{} with $\gamma=1$, which is especially true when the clients frequently communicate with each other, i.e., with lower threshold values.
% , and the more the clients communicate with each other, the lower rewards they will obtain, due to the mistakenly aggregated heterogeneous data. 
In comparison, \modeltwo{} attains relatively good performance even when the clients frequently communicate with each other, as it allows personalized models to be learned on each client. Note that on Delicious dataset, in the low communication/high threshold region (top left corner of Figure \ref{fig:e}), the reward of \modelone{} actually increases as communication increases. Our hypothesis is that, with high threshold, only the most active users contribute to global data sharing, and when the other less active clients download these data, the benefit from reduced variance outweighs the harm caused by the increased bias (due to user heterogeneity). However, with the threshold further reduced, many more clients are able to contribute to global data sharing, such that the global data would become so heterogeneous that it starts to hurt the overall performance. Additional experiment and visualization are given in appendix (Section \ref{sec:additional_exp}) to validate this hypothesis.

(2) MovieLens (Figure \ref{fig:f}): 
Note that on this dataset, \modelone{} with $\gamma=1$ attains very high reward, which indicates that the users share similar preferences, so that data aggregation over different users becomes vital for good performance. 
In this case, learning a personalized model on each client becomes unnecessary and slows down the convergence of model estimation, which leads to the lower accumulative reward of \modeltwo{} compared with the other two algorithms.
% And as \modeltwo{} requires each client to learn a personalized model, which becomes unnecessary in this case, 
% it slows down convergence of the model estimation and thus leads a lower reward compared with the other two algorithms.
However, we can see that \modeltwo{} can still benefit from collaborative model estimation, as it has a much higher accumulative reward than the extreme case of \modelone{} with $\gamma=+\infty$.

\section{Conclusion}
%more and more people are reluctant to provide their own data and strict regulations on data usage like GDPR have also went into effect \cite{voigt2017eu}, which makes 
% The omnipresence of intelligent system in modern life has brought convenience but also increasing public awareness about data privacy. Coupled with the growing computaional power of mobile devices, decentralized learning method like federated learning has emerged as a promising solution to this dilemma.
In this paper, we propose an asynchronous event-triggered communication framework for federated linear bandit problem, which offers a flexible way to balance regret and communication cost. 
Based on this communication framework, two UCB-type algorithms are proposed for homogeneous clients and the more challenging heterogeneous clients, respectively.
From a theoretical aspect, we prove rigorously that our algorithm strikes a better tradeoff between regret and communication cost than existing works in the general case when the distribution over clients is non-uniform. From a practical aspect, 
%compared with existing works that all require synchronization of the clients, 
ours is the first asynchronous method for federated linear bandit. It is more robust against lagging communications, which are often inevitable in reality, and handles heterogeneity in different clients' learning tasks. Hence, it has greater potential in large-scale decentralized applications.
% we identified three challenges in applying bandit learning under the federated learning setting: 1. the trade-off between regret and communication cost; 2. 

% From the discussion in Section \ref{subsec:async_comm}, we know 
% The $\Omega(d\sqrt{T})$ minimax lower regret bound of standard linear bandit also holds in federated setting. However, the optimal trade-off between regret and communication cost is still unknown, e.g., lower bound on communication cost for consistent algorithms (with subpolynomial regret). 
The optimal trade-off between regret and communication cost is still unknown for this problem, e.g., lower bound on communication cost for a certain rate of regret.
Another interesting direction is a differential-private version of the proposed asynchronous algorithms, e.g., trading off regret and communication cost under given privacy budget. 
%e.g., carefully perturbing the transferred parameters, with the tree-based mechanism. 

% \bibliographystyle{plain}
\bibliography{bibfile}

% \newpage
\appendix
\section{Notations and Technical Lemmas}\label{lem:technical_lem}
Let $V \in \bR^{d \times d}$ be a positive semi-definite matrix. We denote the norm of vector $\bx \in \bR^{d}$ induced by $V$ as $||\bx||_{V}=\sqrt{\bx^{\top} V \bx}$. And we denote the operator norm of $V$ as $||V||_{op}=\max_{\bx \neq 0} \frac{||V \bx||_{p}}{||\bx||_{p}}$, where $||\cdot||_{p}$ denotes the $\ell_p$ norm.

Given $\by \in \bR^{d}$, the Euclidean projection of $\by$ onto a (non-empty and compact) set $\Theta \subseteq \bR^{d}$ is denoted as
\begin{align*}
    \textit{proj}_{\Theta}(\by)=\argmin_{\bx \in \Theta} ||\bx-\by||_{2}
\end{align*}
In particular, when $\Theta=\{\bx:||\bx||_{2} \leq S\}$, i.e., an $\ell_2$ ball with radius $S$, $\textit{proj}_{\{\bx:||\bx||_{2} \leq S\}}(\by)=\frac{\by}{\max{(||\by||_{2}/S,1)}}$.

% \begin{lemma}[Sum of quadratic forms]
% Let $A$, $B$ be positive semi-definite matrices, then $\bx^{\top}(A+B)\bx=\bx^{\top}A\bx+\bx^{\top}B\bx$.
% \end{lemma}

\begin{lemma}[Lemma 12 of \cite{abbasi2011improved}] \label{lem:quadratic_det_inequality}
Let $A$, $B$ and $C$ be positive semi-definite matrices such that $A=B+C$. Then, we have that:
\begin{align*}
    \sup_{\bx \neq \textbf{0}} \frac{\bx^{\top} A \bx}{\bx^{\top} B \bx} \leq \frac{\det(A)}{\det(B)}
\end{align*}
\end{lemma}

\begin{lemma}[Theorem 1 of \cite{abbasi2011improved}] \label{lem:self_normalized_bound}
Let $\{\cF_{t}\}_{t=0}^{\infty}$ be a filtration. Let $\{\eta_{t}\}_{t=1}^{\infty}$ be a real-valued stochastic process such that $\eta_{t}$ is $\cF_{t}$-measurable, and $\eta_{t}$ is conditionally zero mean $R$-sub-Gaussian for some $R \geq 0$.
Let $\{X_{t}\}_{t=1}^{\infty}$ be a $\bR^{d}$-valued stochastic process such that $X_{t}$ is $\cF_{t-1}$-measurable. Assume that $V$ is a $d \times d$ positive definite matrix. For any $t > 0$, define
\begin{align*}
    V_{t}=V+\sum_{\tau=1}^{t} X_{\tau} X_{\tau}^{\top} \quad \cS_{t}=\sum_{\tau=1}^{t}\eta_{\tau} X_{\tau} 
\end{align*}
Then for any $\delta >0$, with probability at least $1-\delta$,
\begin{align*}
    ||\cS_{t}||_{V_{t}^{-1}} \leq  R \sqrt{2\log{\frac{\det(V_{t})^{1/2}}{\det(V)^{1/2}\delta}}}, \quad \forall t\geq 0
\end{align*}
\end{lemma}

\begin{lemma}[Bounded random variable] \label{lem:bounded_rv}
Let $X$ be a real random variable such that $X \in [a,b]$ almost surely. Then $$\bE[\exp(s X)]\leq \exp( \frac{s^{2}(b-a)^{2}}{8})$$ for any $s \in \bR$, or equivalently, $X$ is $\frac{b-a}{2}$-sub-Gaussian.
\end{lemma}

\begin{lemma} \label{lem:quadratic_product} 
For a symmetric positive definite matrix $A\in \bR^{d \times d}$ and any vector $\bx\in \bR^{d}$, we have the following inequality
\begin{align*}
    \bx^{\top} \bx\leq \bx^{\top} A \bx \cdot \bx^{\top} A^{-1} \bx \leq \frac{||\bx||_{2}^{4}\lambda_{max}(A)}{\lambda_{min}(A)}
\end{align*}
\end{lemma}

\begin{lemma}[Matrix Freedman’s inequality \citep{tropp2011freedman}] \label{lem:freedman}
Consider a matrix martingale $\{Y_{s}\}_{s=1,2,\dots}$ whose values are matrices with dimension $d_{1} \times d_{2}$, and let $\{Z_{s}\}_{s=1,2,\dots}$ be the corresponding martingale difference sequence. Assume that the difference sequence is almost surely uniformly bounded, i.e., $||Z_{s}||_{op} \leq R$, for $s=1,2,\dots$. 

Define two predictable quadratic variation processes of the martingale:
\begin{align*}
    W_{col,t} & := \sum_{s=1}^{t} \bE_{s-1}[Z_{s} Z_{s}^{\top}] \quad \text{and} \\
    W_{row,t} & := \sum_{s=1}^{t} \bE_{s-1}[Z_{s}^{\top} Z_{s}] \quad \text{for} \hspace{0.25em} t=1,2,\dots
\end{align*}
Then for all $u \geq 0$ and $\omega^{2} \geq 0$, we have
\small
\begin{align*}
    P(\exists t \geq 0: ||Y_{t}||_{op}\geq u, \hspace{0.25em} \text{and} \hspace{0.25em} \max\{||W_{col,t}||_{op}, ||W_{row,t}||_{op}\} \leq \omega^{2}) \leq (d_{1}+d_{2}) \exp{\left(-\frac{u^{2}/2}{\omega^{2}+R u/3}\right)}
\end{align*}
\normalsize
\end{lemma}

\section{Proof of Lemma \ref{lem:Gamma_upperbound}}

% Recall that $r_{t} \leq 2 \alpha_{i_{t},t-1} \sqrt{\bx_{t}^{\top}V_{i_{t},t-1}^{-1}\bx_{t}} = 2 \alpha_{i_{t},t-1} \sqrt{\bx_{t}^{\top}{V}_{t-1}^{-1}\bx_{t}}\sqrt{\Gamma_{t-1}}$, where $\Gamma_{t-1} =\frac{\bx_{t}^{\top}{V}_{i_{t},t-1}^{-1}\bx_{t}}{\bx_{t}^{\top}{V}_{t-1}^{-1}\bx_{t}}$. 
% To show that $r_{t} \leq O\left(\sqrt{d\log{\frac{T}{\delta}}}\right)\sqrt{\bx_{t}^{\top}{V}_{t-1}^{-1}\bx_{t}}\sqrt{\gamma_{D} [1+(N-1)(\gamma_{U}-1)]}$, 
To show that $\Gamma_{t-1} \leq \frac{8 \gamma_{D}}{\lambda_{c}} [1+(N-1)(\gamma_{U}-1)]$, we first need the following lemma.

\begin{lemma} \label{lem:min_eig}
Denote the number of observations that have been used to update $\{V_{i,t}, b_{i,t}\}$ as $\tau_{i}$, i.e., $V_{i,t}=\lambda I + \sum_{s=1}^{\tau_{i}}\bx_{s} \bx_{s}^{\top}$.
Then under Assumption \ref{assump:context_diversity}, with probability at least $1-\delta^{'}$, we have:
\begin{align*}
    \lambda_{\min}(V_{i,t}) \geq \lambda + \frac{\lambda_{c} \tau_{i}}{8}
\end{align*}
$\forall \tau_{i} \in\{\tau_{min},\tau_{min}+1,\dots,T\}$, where $\tau_{min}=\lceil \frac{32(1+L^{2})}{3 \lambda_{c}}\log(\frac{2Td}{\delta^{\prime}}) \rceil$.
\end{lemma}

\noindent \textit{Proof of Lemma \ref{lem:min_eig}.}
This proof is based on standard matrix martingale arguments, and is included here for the sake of completeness. 

Consider the random variable $(z^{\top} \bx_{s,a})^{2}$, where $z \in \bR^{d}$ is an arbitrary vector such that $\lVert z \rVert_{2}\leq 1$ and $\bx_{s,a} \in \cA_{s}=\{\bx_{s,1}, \bx_{s,2}, \dots, \bx_{s,K}\}$. 
Then by Assumption \ref{assump:context_diversity}, $(z^{\top} \bx_{s,a})^{2}$ is sub-Gaussian with variance parameter $v^{2}$. Now we follow the same argument as Claim 1 of \citet{gentile2014online} to derive a lower bound for $\lambda_{\min}(\Sigma_{s})$. First we construct $Z_{a}=(z^{\top} \bx_{s,a})^{2}-\bE_{s-1}[(z^{\top} \bx_{s,a})^{2}]$, for $a \in [K]$. Due to (conditional) sub-Gaussianity, we have
$$P_{s-1}(Z_{a} < -h) \leq P_{s-1}(|Z_{a}| > h) \leq 2e^{-\frac{h^{2}}{2v^{2}}}$$
Then by union bound, and the fact that $\bE_{s-1}[(z^{\top} \bx_{s,a})^{2}]= z^{\top}\Sigma_{c}z \geq \lambda_{c}$, we have:
$$P_{s-1}\bigl( \min_{a\in[K]} (z^{\top} \bx_{s,a}) \geq \lambda_{c} - h \bigr) \geq (1-2e^{-\frac{h^{2}}{2v^{2}}})^{K}$$
Therefore,
$$\bE_{s-1}((z^{\top} \bx_{s})^{2}) \geq \bE_{s-1}(\min_{a\in[K]}(z^{\top} \bx_{s,a})^{2}) \geq (\lambda_{c}-h) (1-2e^{-\frac{h^{2}}{2v^{2}}})^{K}$$
Then by seting $h=\sqrt{2 v^{2}\log{(4K)}}$, we have $(1-2e^{-\frac{h^{2}}{2v^{2}}})^{K} = (1-\frac{1}{2K})^{K} \geq \frac{1}{2}$ because $K \geq 1$, and $(\lambda_{c}-h)\geq \frac{\lambda_{c}}{2}$ because of the assumption on $v^{2}$. Now we have $z^{\top} \Sigma_{s} z = \bE_{s-1}((z^{\top} x_{s})^{2}) \geq \frac{1}{4}\lambda_{c},\forall z$, so $\lambda_{\min}(\Sigma_{s}) \geq \frac{1}{4}\lambda_{c}$.

Then we are ready to lower bound $\lambda_{\min}(V_{i,t})$ as shown below. Specifically, consider the sequence $Y_{\tau_{i}}:= \sum_{s=1}^{\tau_{i}}[\bx_{s} \bx_{s}^{\top} - \Sigma_{s}]$, for $\tau_{i}=1,2,\dots$. And $\{Y_{\tau_{i}}\}_{\tau_{i}=1,2,\dots}$ is a matrix martingale, because $\bE[\lVert Y_{\tau_{i}} \rVert_{op}] < +\infty$ and $\bE_{\tau_{i}-1}[Y_{\tau_{i}}]=\sum_{s=1}^{\tau_{i}-1}[\bx_{s}\bx_{s}-\Sigma_{s}]+\bE_{\tau_{i}-1}[\bx_{\tau_{i}}\bx_{\tau_{i}}^{\top}-\Sigma_{\tau_{i}}]=Y_{\tau_{i}-1}$. Then with the Matrix Freedman inequality (Lemma \ref{lem:freedman}), we have
\begin{equation}
    P( \lVert \sum_{s=1}^{\tau_{i}} (x_{s} x_{s}^{\top} - \Sigma_{s}) \rVert_{op} \geq u) \leq 2d \exp(\frac{-u^{2}/2}{w^{2}+2 u L^{2}/3})
\end{equation}
where $\lVert \cdot \rVert_{op}$ denotes the operator norm. This can be rewritten as $P(- \lVert \sum_{s=1}^{\tau_{i}}\Sigma_{s} - \sum_{s=1}^{\tau_{i}} x_{s} x_{s}^{\top} \rVert_{op} > -u) \geq 1-2d \exp(\frac{-u^{2}/2}{w^{2}+2 u L^{2}/3})$. Then, we have
\begin{align*}
    & 1-2d \exp(\frac{-u^{2}/2}{w^{2}+2 u L^{2}/3}) \leq P(-\lVert \sum_{s=1}^{\tau_{i}}\Sigma_{s} - \sum_{s=1}^{\tau_{i}} x_{s} x_{s}^{\top} \rVert_{op} > -u) \leq P(-\lambda_{\min}(\sum_{s=1}^{\tau_{i}}\Sigma_{s} - \sum_{s=1}^{\tau_{i}} x_{s} x_{s}^{\top} ) > -u) \\
    & \leq P(-\lambda_{\min}(\sum_{s=1}^{\tau_{i}}\Sigma_{s}) + \lambda_{\min}( \sum_{s=1}^{\tau_{i}} x_{s} x_{s}^{\top} ) > -u) \leq P(-\sum_{s=1}^{\tau_{i}}\lambda_{\min}(\Sigma_{s}) + \lambda_{\min}( \sum_{s=1}^{\tau_{i}} x_{s} x_{s}^{\top} ) > -u) \\
    & \leq P(\lambda_{\min}( \sum_{s=1}^{\tau_{i}} x_{s} x_{s}^{\top} ) > \frac{\tau_{i}\lambda_{c}}{4} -u)
\end{align*}
where the third and forth inequalities are due to Weyl's inequality, i.e., $\lambda_{\min}(A+B) \geq \lambda_{\min}(A)+\lambda_{\min}(B)$ for symmetric matrices $A$ and $B$, and the fifth inequality is due to $\lambda_{\min}(\Sigma_{s}) \geq \frac{1}{4}\lambda_{c}$.

By setting $u=\frac{\lambda_{c}\tau_{i}}{8}$ and $w^{2}=\frac{\tau_{i}}{12}$, we have $P(\lambda_{\min}( \sum_{s=1}^{\tau_{i}} x_{s} x_{s}^{\top} ) > \frac{\lambda_{c}\tau_{i}}{8}) \geq 1-2d\exp(\frac{-\lambda_{c}\tau_{i}}{32(1+L^{2})/3})$. Then when $\tau_{i} \geq \frac{32(1+L^{2})}{3 \lambda_{c}}\log(\frac{2Td}{\delta^{\prime}}):=\tau_{\min}$, we have $P(\lambda_{\min}( \sum_{s=1}^{\tau_{i}} x_{s} x_{s}^{\top} ) > \frac{\lambda_{c}\tau_{i}}{8}) \geq 1-\frac{\delta^{\prime}}{T}$. By taking a union bound over all 
$\tau_{i} \in\{\tau_{min},\tau_{min}+1,\dots,T\}$, we have $P(\lambda_{\min}( V_{i,t} ) > \lambda + \frac{\lambda_{c}\tau_{i}}{8}) \geq 1-\delta^{\prime}$, which finishes the proof.

\noindent \textit{Proof of Lemma \ref{lem:Gamma_upperbound}.}
% \textit{Proof of Lemma \ref{lem:Gamma_upperbound}.}
% $\Gamma_{t-1}$ measures the ``delayedness" of $\tilde{V}_{i_{t},t-1}+\bar{V}_{-i_{t},t-1}$ compared with $\tilde{V}_{t}$. Note that no individual agent in the learning system has access to $\tilde{V}_{t}$, so it is not possible to design a triggering event that can directly upper bound $\Gamma_{t-1}$. However, we can use the aggregated sufficient statistics stored on the server, i.e. $\sum_{i=1}^{N}\bar{V}_{i,t-1}$ as an intermediate. For example, the `download' triggering event guarantees $\tilde{V}_{i_{t},t-1}+\bar{V}_{-i_{t},t-1}$ will not be too ``delayed" compared with $\sum_{i=1}^{N}\bar{V}_{i,t-1}$, and the upload triggering event guarantees $\sum_{i=1}^{N}\bar{V}_{i,t-1}$ will not be too delayed compared with $\tilde{V}_{t}$.

\noindent Under Lemma \ref{lem:quadratic_product}, we have $$\Gamma_{t-1}=\frac{\bx_{t}^{\top}{V}_{i_{t},t-1}^{-1}\bx_{t}}{\bx_{t}^{\top}{V}_{t-1}^{-1}\bx_{t}} \leq \frac{\lambda_{max}(V_{i_{t},t-1})}{\lambda_{min}(V_{i_{t},t-1})}\frac{\bx_{t}^{\top}{V}_{t-1}\bx_{t}}{\bx_{t}^{\top}{V}_{i_{t},t-1}\bx_{t}}$$
Then when $\tau_{i_{t}} \geq \tau_{min}$, with Lemma \ref{lem:min_eig} and the fact that $\lambda_{max}(V_{i_{t},t-1})\leq \lambda + \tau_{i_{t}}$, w.h.p. we have $$\Gamma_{t-1} \leq \frac{\lambda + \tau_{i_{t}}}{\lambda+{\tau_{i_{t}}\lambda_{c}}/{8}} \cdot \frac{\bx_{t}^{\top}{V}_{t-1}\bx_{t}}{\bx_{t}^{\top}{V}_{i_{t},t-1}\bx_{t}} \leq \frac{8}{\lambda_{c}}\cdot \frac{\bx_{t}^{\top}{V}_{t-1}\bx_{t}}{\bx_{t}^{\top}{V}_{i_{t},t-1}\bx_{t}}$$ where the second inequality is because, for bounded context vector ($\lVert \bx_{t,a} \rVert_{2} \leq 1$), $\lambda_{c} \leq \frac{1}{d} < 8$, so $\frac{\lambda_{c}}{8}<1$. In this case $r_{t} \leq 
2 \alpha_{i_{t},t-1}\sqrt{\bx_{t}^{\top}{V}_{t-1}^{-1}\bx_{t}}\sqrt{ \frac{8}{\lambda_{c}} \frac{\bx_{t}^{\top}{V}_{t-1}\bx_{t}}{\bx_{t}^{\top}{V}_{i_{t},t-1}\bx_{t}}}$. Note that when $\tau_{i_{t}} < \tau_{min}$, we can simply bound $r_{t}$ by the constant $2 L S$, and in total this added regret is only $O(\log{T})$, which is negligible compared with the $O(\sqrt{T})$ term in the upper bound of $R_{T}$.

Now we need to show that $$\frac{\bx_{t}^{\top}{V}_{t-1}\bx_{t}}{\bx_{t}^{\top}{V}_{i_{t},t-1}\bx_{t}} \leq \gamma_{D} [1+(N-1)(\gamma_{U}-1)]$$
In order to do this, we need the following two facts: 
\begin{itemize}
    \item $V_{i_{t},t-1}-\Delta{V}_{i_{t},t-1}=V_{g,t-1}-\Delta{V}_{-i_{t},t-1}$, because they both equal to the copy of sufficient statistics in the most recent communication between the client $i_{t}$ and the server.
    \item Due to Lemma \ref{lem:quadratic_det_inequality}, and our design of the `upload' and `download' triggering events in Eq \eqref{eq:upload} and Eq \eqref{eq:download}, at the beginning of time $t\in[T]$, the inequalities
\small
\begin{equation} \label{eq:upload_inequailty}
    \sup_{\bx}\frac{\bx^{\top}({V}_{j,t-1})\bx}{\bx^{\top} (V_{j,t-1}-\Delta{V}_{j,t-1}) \bx} \leq \frac{\det({V}_{j,t-1})}{\det(V_{j,t-1}-\Delta{V}_{j,t-1})} \leq \gamma_{U}
\end{equation}
\normalsize
and  
\small
\begin{equation}\label{eq:download_inequailty}
    \sup_{\bx}\frac{\bx^{\top}({V}_{g,t-1})\bx}{\bx^{\top} (V_{g,t-1}-\Delta{V}_{-j,t-1}) \bx} \leq \frac{\det({V}_{g,t-1})}{\det(V_{g,t-1}-\Delta{V}_{-j,t-1})} \leq \gamma_{D}
\end{equation}
\normalsize
hold $\forall j \in [N], \forall t\in [T]$.
\end{itemize}
Then by decomposing $\frac{\bx_{t}^{\top}V_{t-1}\bx_{t}}{\bx_{t}^{\top}V_{i_{t},t-1}\bx_{t}}$, we have:
\begin{equation*}
\begin{split}
& \frac{\bx_{t}^{\top}(V_{t-1})\bx_{t}}{\bx_{t}^{\top}(V_{i_{t},t-1})\bx_{t}} = \frac{\bx_{t}^{\top}(V_{g,t-1}+\sum_{j=1}^{N}\Delta{V}_{j,t-1})\bx_{t}}{x_{t}^{\top}(V_{i_{t},t-1}-\Delta{V}_{i_{t},t-1}+\Delta{V}_{i_{t},t-1})\bx_{t}} \\
& \leq \frac{\bx_{t}^{\top}(V_{g,t-1}+\sum_{j\neq 1}\Delta{V}_{j,t-1})\bx_{t}}{\bx_{t}^{\top}(V_{i_{t},t-1}-\Delta{V}_{i_{t},t-1})\bx_{t}} = \frac{\bx_{t}^{\top}(V_{g,t-1})\bx_{t} +\sum_{j\neq 1} \bx_{t}^{\top}(\Delta{V}_{j,t-1})\bx_{t}}{\bx_{t}^{\top}(V_{g,t-1}-\Delta{V}_{-i_{t},t-1})\bx_{t}} \\
\end{split}
\end{equation*}
% \begin{equation*}
% \begin{split}
% \Gamma_{t-1} & = \frac{x_{t}^{\top}(\sum_{i=1}^{N} [\bar{V}_{i,t-1} + \Delta V_{i,t-1}])x_{t}}{x_{t}^{\top}(\bar{V}_{i_{t},t-1}+\Delta V_{i_{t},t-1}+\bar{V}_{-i_{t},t-1})x_{t}} \\
% & \leq \frac{x_{t}^{\top}(\sum_{i=1}^{N}\bar{V}_{i,t-1}+\sum_{i \neq i_{t}} \Delta V_{i,t-1})x_{t}}{x_{t}^{\top}(\bar{V}_{i_{t},t-1}+\bar{V}_{-i_{t},t-1})x_{t}} \\
% & = \frac{x_{t}^{\top}(\sum_{i=1}^{N}\bar{V}_{i,t-1})x_{t} +\sum_{i \neq i_{t}} x_{t}^{\top} (\Delta V_{i,t-1} ) x_{t} }{x_{t}^{\top}(\bar{V}_{i_{t},t-1}+\bar{V}_{-i_{t},t-1})x_{t}} \\
% % = & \frac{x_{t}^{\top}(\bar{V}_{g,t-1})x_{t} +\sum_{i \neq i_{t}} x_{t}^{\top} (\bar{V}_{g,t-1}) x_{t} \cdot \frac{x_{t}^{\top}\Delta V_{i,t-1} x_{t}}{x_{t}^{\top}\bar{V}_{g,t-1} x_{t}} }{x_{t}^{\top}(\bar{V}_{i_{t},t-1}+\bar{V}_{-i_{t},t-1})x_{t}} \\
% % \leq & \frac{x_{t}^{\top}(\bar{V}_{g,t-1})x_{t} +\sum_{i \neq i_{t}} x_{t}^{\top} (\bar{V}_{g,t-1}) x_{t} \cdot \frac{x_{t}^{\top}\Delta V_{i,t-1} x_{t}}{x_{t}^{\top} (\bar{V}_{i,t-1}+\bar{V}_{-i,t-1}) x_{t}} }{x_{t}^{\top}(\bar{V}_{i_{t},t-1}+\bar{V}_{-i_{t},t-1})x_{t}} \\
% % = & \frac{x_{t}^{\top}(\bar{V}_{g,t-1})x_{t} +\sum_{i \neq i_{t}} x_{t}^{\top} (\bar{V}_{g,t-1}) x_{t} \cdot [\frac{x_{t}^{\top}(\Delta V_{i,t-1}+\bar{V}_{i,t-1}+\bar{V}_{-i,t-1}) x_{t} }{x_{t}^{\top} (\bar{V}_{i,t-1}+\bar{V}_{-i,t-1}) x_{t}} -1] }{x_{t}^{\top}(\bar{V}_{i_{t},t-1}+\bar{V}_{-i_{t},t-1})x_{t}} \\
% \end{split}
% \end{equation*}
And the term $\sum_{j\neq 1} \bx_{t}^{\top}(\Delta{V}_{j,t-1})\bx_{t}$ can be further upper bounded by:
\begin{align*}
    & \sum_{j\neq 1} \bx_{t}^{\top}(\Delta{V}_{j,t-1})\bx_{t} = \bx_{t}^{\top} V_{g,t-1} \bx_{t} \cdot \sum_{j \neq i_{t}}  \frac{\bx_{t}^{\top}(\Delta{V}_{j,t-1})\bx_{t}}{\bx_{t}^{\top} V_{g,t-1} \bx_{t}}  \\
    & \leq \bx_{t}^{\top} V_{g,t-1} \bx_{t}\cdot  \sum_{j \neq i_{t}} \frac{\bx_{t}^{\top}(\Delta{V}_{j,t-1})\bx_{t}}{\bx_{t}^{\top} (V_{g,t-1}-\Delta{V}_{-j,t-1}) \bx_{t}}  = \bx_{t}^{\top} V_{g,t-1} \bx_{t}\cdot  \sum_{j \neq i_{t}} \frac{\bx_{t}^{\top}(\Delta{V}_{j,t-1})\bx_{t}}{\bx_{t}^{\top} (V_{j,t-1}-\Delta{V}_{j,t-1}) \bx_{t}}  \\
    & = \bx_{t}^{\top} V_{g,t-1} \bx_{t} \cdot  \sum_{j \neq i_{t}} \bigl[ \frac{\bx_{t}^{\top}({V}_{j,t-1})\bx_{t}}{\bx_{t}^{\top} (V_{j,t-1}-\Delta{V}_{j,t-1}) \bx_{t}} -1 \bigr] \leq \bx_{t}^{\top} V_{g,t-1} \bx_{t} \cdot (N-1)(\gamma_{U}-1)
\end{align*}
where the last inequality is due to Eq \eqref{eq:upload_inequailty}.
% Note that according to our current communication strategy, $\frac{x_{t}^{\top}(\Delta V_{i,t-1}+\bar{V}_{i,t-1}+\bar{V}_{-i,t-1}) x_{t} }{x_{t}^{\top} (\bar{V}_{i,t-1}+\bar{V}_{-i,t-1}) x_{t}} \leq \frac{\det{(\Delta V_{i,t-1}+\bar{V}_{i,t-1}+\bar{V}_{-i,t-1})}}{\det{(\bar{V}_{i,t-1}+\bar{V}_{-i,t-1})}} \leq \gamma^{U}$ and $\frac{x_{t}^{\top}(\bar{V}_{g,t-1})x_{t}}{x_{t}^{\top}(\bar{V}_{i_{t},t-1}+\bar{V}_{-i_{t},t-1})x_{t}} \leq \frac{\det{(\bar{V}_{g,t-1})}}{\det{(\bar{V}_{i_{t},t-1}+\bar{V}_{-i_{t},t-1})}} \leq \gamma^{D}$.
Then by substituting this back, and using Eq \eqref{eq:download_inequailty}, we have
\begin{align*}
    \frac{\bx_{t}^{\top}(V_{t-1})\bx_{t}}{\bx_{t}^{\top}(V_{i_{t},t-1})\bx_{t}} \leq \frac{\bx_{t}^{\top}({V}_{g,t-1})\bx_{t} [1+(N-1)(\gamma_{U}-1)] }{\bx_{t}^{\top}(V_{g,t-1}-\Delta{V}_{-i_{t},t-1})\bx_{t}} \leq \gamma_{D} [1+(N-1)(\gamma_{U}-1)]
\end{align*}
which finishes the proof.
% Therefore, with probability at least $1-\delta^{\prime}$, $\Gamma_{t-1} = O(\gamma_{D} [1+(N-1)(\gamma_{U}-1)])$.
% When setting $\gamma^{D}=\gamma^{U}=1$, global synchronization happens in each time step and $\Gamma_{t-1}=1$ holds $\forall t \in [T]$, which means our algorithm essentially reduces to LinUCB under the centralized learning setting.

\section{Proof of Theorem \ref{thm:regret_communication_async-linucb} (Regret and Communication Upper Bound for \modelone{})}
\noindent \textbf{Regret}:
Based on the discussion in Section \ref{subsec:async_comm} that the instantaneous regret $r_{t}$ directly depends on $\Gamma_{t-1}$, we can upper bound the accumulative regret of \modelone{} by 
\begin{align*}
& R_{T} = \sum_{t=1}^{T}r_{t} \leq \sum_{t=1}^{T}O\left(\sqrt{d\log{\frac{T}{\delta}}}\right)\sqrt{\bx_{t}^{\top}{V}_{t-1}^{-1}\bx_{t}}\sqrt{\Gamma_{t-1}} \\
& \leq O\left(\sqrt{d\log{\frac{T}{\delta}}}\right)\sqrt{\sum_{t=1}^{T}x^{\top}{V}_{t-1}^{-1}x} \sqrt{\sum_{t=1}^{T}\Gamma_{t-1}} \\
& \leq O\left(\sqrt{d\log{\frac{T}{\delta}}}\right)  \sqrt{\log{\frac{det({V}_{T-1})}{det(\lambda I)}}}\sqrt{\sum_{t=1}^{T}\Gamma_{t-1}}
\end{align*}
where the second inequality is by the Cauchy–Schwarz inequality, and the third is based on Lemma 11 in \cite{abbasi2011improved}.
Then with the upper bound of $\Gamma_{t-1}$ given in Lemma \ref{lem:Gamma_upperbound}, the accumulative regret is upper bounded by $R_{T} = O\left(d\sqrt{T\log^{2}{T}}\min(\sqrt{N},\sqrt{\gamma_{D} [1+(N-1)(\gamma_{U}-1)]})\right)$. 

\noindent \textbf{Communication cost}:
As discussed in Section \ref{subsec:async_comm}, clients collaborate by transferring updates of the sufficient statistics, i.e., $\{ \Delta V \in \bR^{d \times d},\Delta b \in \bR^{d}\}$. Since our target is not to reduce the size of these parameters, for all following discussions, we define the communication cost $C_{T}$ as the number of times $\{ \Delta V,\Delta b\}$ being transferred between agents.
To analyze $C_{T}$, we denote the sequence of time steps when either `upload' or `download' is triggered up to time $T$ as $\{t_{1},t_{2},\dots,t_{C_{T,i}}\}$, where $C_{T,i}$ is the total number of communications between client $i$ and the server.
Then the corresponding sequence of local covariance matrices is $\{\lambda I,V_{i,t_{1}},V_{i,t_{2}},\dots,V_{i,t_{C_{T,i}}}\}$. 
We can decompose $$\log{\frac{\det{V_{i,t{C_{T,i}}}}}{\det{\lambda I}}}=\log{\frac{\det{V_{i,t_{1}}}}{\det{\lambda I}}} + \log{\frac{\det{V_{i,t_{2}}}}{\det{V_{i,t_{1}}}}}+\dots\log{\frac{\det{V_{i,t_{C_{T,i}}}}}{\det{V_{i,t_{C_{T,i}-1}}}}} \leq \log{\frac{\det{{V}_{T-1}}}{\det{\lambda I}}}$$
Since the matrices in the sequence trigger either Eq \eqref{eq:upload} or Eq \eqref{eq:download}, each term in this summation is lower bounded by $\log{\min(\gamma_{U},\gamma_{D})}$. When $\min(\gamma_{U},\gamma_{D})>1$, by the pigeonhole principle, $C_{T,i} \leq \frac{\log \det({V}_{T-1})-d \log \lambda}{\log \min(\gamma_{U},\gamma_{D})}$; as a result, the communication cost for $N$ clients is 
$C_{T} =\sum_{i=1}^{N} C_{T,i} \leq N \frac{\log{\det({V}_{T-1})}-d \log{\lambda}}{\log{\min{(\gamma_{U},\gamma_{D})}}}$.

\section{Synchronous Communication Method} \label{sec:sync_method}
The synchronous method DisLinUCB (Appendix G in \cite{wang2019distributed}) imposes a stronger assumption about the appearance of clients: i.e., they assume all $N$ clients interact with the environment in a round-robin fashion (so $\cN_{i}(T)=\frac{T}{N}$ \footnote{It is worth noting the difference in the meaning of $T$ between our paper and \cite{wang2019distributed}. In our paper, $T$ is the total number of interactions for all $N$ clients, while for \cite{wang2019distributed}, $T$ is the total number of interactions for each client.}). 
For the sake of completeness, we present the formal description of this algorithm adapted to our problem setting in Algorithm \ref{algo:SyncLinUCB} (which is referred to as Synchronous LinUCB algorithm, or \modelbaseline{} for short), and provide the corresponding theoretical analysis about its regret $R_{T}$ and communication cost $C_{T}$ under both uniform and non-uniform client distribution. In particular, in this setting we no longer assume uniform appearance of clients. 

% In order to compare with their theoretical results for regret $R_{T}$ and communication cost $C_{T}$ and compare the empirical performance in experiments, we slightly modified this algorithm. Here we provide the formal description of the resulting algorithm, which we will refer to as synchronous LinUCB algorithm (Sync-LinUCB), and the corresponding theoretical results in our problem setting.

\begin{algorithm}
    \caption{Synchronous LinUCB Algorithm} \label{algo:SyncLinUCB}
  \begin{algorithmic}[4]
    \STATE \textbf{Input:} threshold $D$, $\sigma, \lambda > 0$, $\delta \in (0,1)$
    \STATE Initialize server: ${V}_{g, 0}=\textbf{0}_{d \times d} \in \mathbb{R}^{d \times d}$, ${b}_{g,0}=\textbf{0}_{d} \in \mathbb{R}^{d}$
        % \hspace*{2em} $\Delta{V}_{-i, 0}=\textbf{0} \in \mathbb{R}^{d \times d}$, $\Delta{b}_{-i,0}=\textbf{0} \in \mathbb{R}^{d}$, $i\in[N]$
    \FOR{ $t=1,2,...,T$}
        \STATE Observe arm set $\mathcal{A}_{t}$ for client $i_{t} \in [N]$
        \IF{client $i_{t}$ is new}
            \STATE Initialize client $i_{t}$: ${V}_{i_{t}, t-1}=\textbf{0}_{d \times d}$, ${b}_{i_{t},t-1}=\textbf{0}_{d}$, $\Delta{V}_{i_{t}, t-1}=\textbf{0}_{d \times d}$, $\Delta{b}_{i_{t},t-1}=\textbf{0}_{d}$, $\Delta t_{i_{t},t-1}=0$
        \ENDIF
        \STATE Select arm $\bx_{t}\in\cA_{t}$ by Eq \eqref{eq:UCB} and observe reward $y_{t}$
        \STATE Update client $i_{t}$:
            % ${V}_{i_{t},t}={V}_{i_{t},t-1}+x_{t}x_{t}^{T}$, ${b}_{i_{t},t}={b}_{i_{t},t-1}+x_{t}y_{t}$ \newline
            % $\Delta{V}_{i_{t},t}=\Delta{V}_{i_{t},t-1}+x_{t}x_{t}^{T}$, $\Delta{b}_{i_{t},t}=\Delta{b}_{i_{t},t-1}+x_{t}y_{t}$
            ${V}_{i_{t},t} \mathrel{+}= \bx_{t}\bx_{t}^{T}$, ${b}_{i_{t},t} \mathrel{+}= \bx_{t}y_{t}$, $\Delta{V}_{i_{t},t} \mathrel{+}= \bx_{t}\bx_{t}^{T}$, $\Delta{b}_{i_{t},t}+=\bx_{t}y_{t}$, $\Delta t_{i_{t},t} \mathrel{+}= 1$  \newline
            \textit{\# Check whether global synchronization is triggered}
        \IF{$\Delta t_{i_{t},t}\log{\frac{\det(V_{i_{t},t}+\lambda I)}{\det(V_{i_{t},t}-\Delta{V}_{i_{t},t}+\lambda I)}} > D$}
            % \STATE Upload $\Delta{V}_{i_{t},t},\Delta{b}_{i_{t},t}$ ($i_{t} \rightarrow \text{server}$) 
            \FOR{$i = 1, \dots, N$}
                \STATE Upload $\Delta{V}_{i,t},\Delta{b}_{i,t}$ ($i \rightarrow \text{server}$)
                \STATE Client $i$ reset $\Delta{V}_{i,t}=\textbf{0}$, $\Delta{b}_{i,t}=\textbf{0}$, $\Delta t_{i,t}=0$
                \STATE Update server: $V_{g,t}\mathrel{+}=\Delta V_{i,t},b_{g,t}\mathrel{+}=\Delta b_{i,t}$
            \ENDFOR
            \FOR{$i = 1, \dots, N$}
                \STATE Download ${V}_{g,t},{b}_{g,t}$ ($\text{server} \rightarrow i$)
                \STATE Update client $i$: $V_{i,t}=V_{g,t},b_{j,t}=b_{g,t}$
            \ENDFOR
        % \ELSE
        %     \STATE All other sufficient statistics remain unchanged
        \ENDIF
    \ENDFOR
  \end{algorithmic}
\end{algorithm}

In our problem setting (Section \ref{subsec:problem_formulation}), other than assuming each client has a nonzero probability to appear in each time step, we do not impose any further assumption on the clients’ distribution or its frequency of interactions with the environment. This is more general than the setting considered in \cite{wang2019distributed}, since the clients now may have distinct availability of new observations. We will see below that this will cause additional communication cost for \modelbaseline{}, compared with the case where all the clients interact with the environment in a round-robin fashion, i.e., all $N$ clients have equal number of observations.
% However, when users show up following a non-uniform distribution, additional communication cost will be incurred. 
Intuitively, when one single client accounts for the majority of the interactions with the environment and always triggers the global synchronization, all the other $N-1$ clients are forced to upload their local data despite the fact that they have very few new observations since the last synchronization. This directly leads to a waste of communication.
Below we give the analysis of $R_{T}$ and $C_{T}$ of sync-LinUCB considering both uniform and non-uniform client distribution.

\noindent \textbf{Regret of \modelbaseline{}: }
Most part of the proof for Theorem 4 in \cite{wang2019distributed} extends to the problem setting considered in this paper (with slight modifications due to the difference in the meaning of $T$ as mentioned in the footnote). Since now only one client interacts with the environment in each time step, the accumulative regret for the `good epochs' is $REG_{good}=O(d\sqrt{T}\log(T))$. Denote the first time step of a certain `bad epoch' as $t_{s}$ and the last as $t_{e}$. The accumulative regret for this `bad epoch' can be upper bounded by: $O(\sqrt{d\log{T}})\sum_{i=1}^{N}\sum_{\tau \in \cN_{i}(t_{e})\setminus \cN_{i}(t_{s})}\min(1,||\bx_{\tau}||_{V_{i,\tau-1}^{-1}})\leq  O(\sqrt{d\log{T}})\sum_{i=1}^{N} \sqrt{\Delta t_{i,t_{e}}\log{\frac{\det(V_{i,t_{e}-1}+\lambda I)}{\det(V_{i,t_{e}-1}-\Delta{V}_{i,t_{e}-1}+\lambda I)}}} \leq O(\sqrt{d \log{T}} N \sqrt{D})$. And using the same argument as in the original proof, there can be at most $R=O(d \log{T})$ `bad epochs', so that accumulative regret for the `bad epochs' is upper bounded by $REG_{bad}=O(d^{1.5}\log^{1.5}{(T)}N\sqrt{D})$. Therefore, with the threshold $D$, the accumulative regret is $R_{T}=O(d\sqrt{T}\log(T))+O(d^{1.5}\log^{1.5}{(T)}N\sqrt{D})$.

For the analysis of communication cost $C_{T}$, we consider the settings of uniform and non-uniform client distributions separately in the following two paragraphs.

% \subsection{Communication cost under uniform user distribution}
\noindent \textbf{Communication cost of \modelbaseline{} under uniform client distribution: }
Denote the length of an epoch as $\alpha$, so that there can be at most $\lceil \frac{T}{\alpha}\rceil$ epochs with length longer than $\alpha$.
For an epoch with less than $\alpha$ time steps, similarly, we denote the first time step of this epoch as $t_{s}$ and the last as $t_{e}$, i.e., $t_{e}-t_{s} < \alpha$. Then since the users appear in a uniform manner, the number of interactions for any user $i\in[N]$ satisfies $\Delta t_{i,t_{e}} < \frac{\alpha}{N}$. Therefore, $\log{\frac{\det(V_{t_{e}})}{\det(V_{t_{s}})}} > \frac{D N}{\alpha}$. Following the same argument as in the original proof, the number of epochs with less than $\alpha$ time steps is at most $\lceil \frac{R \alpha}{D N}\rceil$. Then $C_{T}=N \cdot (\lceil \frac{T}{\alpha}\rceil+\lceil \frac{R \alpha}{D N}\rceil)$, because at the end of each epoch, the synchronization round incurs $2N$ communication cost. We minimize $C_{T}$ by choosing $\alpha=\sqrt{\frac{D T N}{R}}$, so that $C_{T}=O(N \cdot \sqrt{\frac{T R}{D N}})$. Note that this result is the same as \cite{wang2019distributed} (we can see this by simply substituting $T$ in our result with $TN$), because $T$ in our paper denotes the total number of iterations for all $N$ clients.

% \subsection{Non-uniform user distribution}
\noindent \textbf{Communication cost of \modelbaseline{} under non-uniform client distribution: }
However, for most applications in reality, the client distribution can hardly be uniform, i.e., the clients have distinct availability of new observations. Then the global synchronization of \modelbaseline{} leads to a waste of communication in this more common situation.
Specifically, when considering epochs with less than $\alpha$ time steps, the number of interactions for any client $i\in[N]$ can be equal to $t_{e}-t_{s}$ in the worst case, i.e., all the interactions with the environment in this epoch are done by this single client. In this case, $\Delta t_{i,t_{e}} < \alpha$, which is different from the case of uniform client distribution.
Therefore, $\log{\frac{\det(V_{t_{e}})}{\det(V_{t_{s}})}} > \frac{D}{\alpha}$. The number of epochs with less than $\alpha$ time steps is at most $\lceil \frac{R \alpha}{D}\rceil$. Then $C_{T}=N \cdot (\lceil \frac{T}{\alpha}\rceil+\lceil \frac{R \alpha}{D}\rceil)$. Similarly, we choose $\alpha=\sqrt{\frac{D T}{R}}$ to minimize $C_{T}$, so that $C_{T}= O(N \cdot\sqrt{\frac{T R}{D}})$. We can see that this is larger than the communication cost under a uniform client distribution by a factor of $\sqrt{N}$.

\section{Comparison between \modelone{} and \modelbaseline{}} \label{sec:tb_theretical}
In this section, we provide more details about the theoretical results of \modelone{}, and add the corresponding results of \modelbaseline{} for comparison (see Table \ref{tb:theoretical_comparison}).
% , i.e., the comparison between \modelone{} and \modelbaseline{} under different trade-off settings between regret $R_{T}$ and communication cost $C_{T}$. 
Depending on the application, the thresholds $\gamma_{U}$ and $\gamma_{D}$ of \modelone{} can be flexibly adjusted to get various trade-off between $R_{T}$ and $C_{T}$. For all the discussions below, we constrain $\gamma_{U}=\gamma_{D}=\gamma$ for simplicity. However, when necessary, different values can be chosen for $\gamma_{U}$ and $\gamma_{D}$ for different clients. This gives our algorithm much more flexibility in practice, i.e., allows for a fine-grained control of every single edge in the communication network, compared with \modelbaseline{}.
For example, for users who are less willing to participate in frequent uploads and downloads, a higher threshold can be chosen for their corresponding clients to reduce communication, and vice versa. 
% Or for users who are less active in interacting with the learning system, higher threshold value can be chosen for the download event of their clients to save unnecessary downloads. 
% and summarize some choices of $\gamma$ and the corresponding results in Table \ref{tb:theoretical_comparison}

% We present some choices of the thresholds for \modelone{} and \modelbaseline{}, and the corresponding upper bounds for $R_{T}$ and $C_{T}$ in Table \ref{tb:theoretical_comparison}.

% In the case where communication cost is not a concern, we can chose $\gamma=1$, such that all the clients are synchronized in each time step. This incurs communication cost $C_{T}=NT$, and regret $R_{T}=d\sqrt{T}\log{T}$, which is the same as 

\begin{table*}[h]
% \vspace{-4mm}
\centering
\caption{Upper bounds for $R_{T}$ and $C_{T}$ under different thresholds.}
\begin{tabular}{ c c c c c}
\toprule
%  \multirow{2}{*}{Algorithm} & \multirow{2}{*}{Threshold} & \multirow{2}{*}{$R_{T}$} & \multicolumn{2}{c}{$C_{T}$} \\
%  & & & uniform user & non-uniform user \\
 Algorithm & Threshold & $R_{T}$ & $C_{T}$ (uniform) & $C_{T}$ (non-uniform) \\
 \hline
 \multirow{4}{*}{\modelone{}} 
 & $\gamma=1$ & $d \sqrt{T} \log{T}$ & $N T$ & $N T$ \\  
 & $\gamma=\exp(N^{-1})$ & $ d \sqrt{T} \log{T}$ & $N^{2} d \log{T}$ & $N^{2} d \log{T}$ \\
 & $\gamma=\exp(N^{-\frac{1}{2}})$ & $ N^{\frac{1}{4}} d \sqrt{T} \log{T}$ & $N^{\frac{3}{2}} d \log{T}$  & $N^{\frac{3}{2}} d \log{T}$ \\
%  & constant $\gamma \in (1, +\infty)$ & $N^{\frac{1}{2}} d \sqrt{T} \log{T}$ & $N d \log{T}$ & $N d \log{T}$\\ 
 & $\gamma=+\infty$ & $ N^{\frac{1}{2}} d \sqrt{T} \log{T}$ & $0$ & $0$ \\
 \hline
 \multirow{2}{*}{\modelbaseline{}} & $D={T}/{(N^{2} d \log{T})}$ & $d \sqrt{T} \log{T}$ & $N^{\frac{3}{2}} d \log{T}$ & $N^{2} d \log{T}$ \\
  & $D={T}/{(N^{\frac{3}{2}} d \log{T})}$ & $N^{\frac{1}{4}} d \sqrt{T} \log{T}$ & $N^{\frac{5}{4}} d \log{T}$ & $N^{\frac{7}{4}} d \log{T}$\\ 
%   & $D={T}/{(N d \log{T})}$ & $N^{\frac{1}{2}} d \sqrt{T} \log{T}$ & $N d \log{T}$ & $N^{\frac{3}{2}} d \log{T}$\\ 
\bottomrule
\end{tabular}
\label{tb:theoretical_comparison}
% \vspace{-4mm}
\end{table*}

When setting $\gamma=+\infty$, all communications in the learning system are blocked; and in this case, $C_{T}=0$ and $R_{T}=O(N^{\frac{1}{2}}d\sqrt{T}\log{T})$, which recovers the regret of running an instance of LinUCB for each client independently. When setting $\gamma=1$, the upload and download events are always triggered, i.e., synchronize all $N$ clients in each time step. And in this case $C_{T}=NT$ and $R_{T}=O(d\sqrt{T}\log{T})$, which recovers the regret in the centralized setting. 

What we prefer is to strike a balance between these two extreme cases, i.e., reduce the communication cost without sacrificing too much on regret. Specifically, we should note that $T$ is the dominating variable for almost all applications instead of $N$ or $d$. For example, in the three real-world datasets used in our experiments (Section \ref{sec:exp}), $d$ has an order of $10^{1}$, $N$ has an order of $10^{1}-10^{3}$, but $T$ has an order of $10^{5}$. Since even without communication $R_{T}=O(N^{\frac{1}{2}}d\sqrt{T}\log{T})$ already matches the minimax lower bound $\Omega(d\sqrt{T})$ in $T$ (up to a logarithmic factor) and $d$, we are most interested in the case where $C_{T}$'s rate in $T$ is improved from $O(T)$ to $O(\log{T})$. 

For example, we can set \modelone{}'s upper bound of the communication cost $C_{T} \leq N d \frac{\log{T}}{\log{\gamma}}$ to be $N^{\frac{3}{2}} d \log{T}$, and thus $\gamma=\exp(N^{-\frac{1}{2}})$. Then by substituting $\gamma$ into the upper bound of $R_{T}$, we have 
\begin{align*}
R_{T} &=O\left(\sqrt{(N-1)\gamma^{2}+(2-N)\gamma}d \sqrt{T} \log{T}\right)  = O\left(\sqrt{(N-1)e^{2 N^{-\frac{1}{2}}}+(2-N)e^{N^{-\frac{1}{2}}}}d \sqrt{T} \log{T}\right) 
\end{align*}
% \small
% \begin{align*}
%     R_{T} & =O\left(\sqrt{(N-1)\gamma^{2}+(2-N)\gamma}d \sqrt{T} \log{T}\right)  = O\left(\sqrt{(N-1)e^{2 N^{-\frac{1}{2}}}+(2-N)e^{N^{-\frac{1}{2}}}}d \sqrt{T} \log{T}\right)  \\
% \end{align*}
% \normalsize
Since $\lim_{N \rightarrow \infty} \frac{\sqrt{(N-1)e^{2 N^{-\frac{1}{2}}}+(2-N)e^{N^{-\frac{1}{2}}}}}{N^{\frac{1}{4}}}=1$, we know $\sqrt{(N-1)e^{2 N^{-\frac{1}{2}}}+(2-N)e^{N^{-\frac{1}{2}}}}=O(N^{\frac{1}{4}})$. Therefore, $R_{T}=O(N^{\frac{1}{4}}d \sqrt{T} \log{T})$. And similarly, by setting $\gamma=\exp(N^{-1})$, \modelone{} has $C_{T}=N^{2}d\log{T}$ and $R_{T}=O(d\sqrt{T}\log{T})$. For both choices of $\gamma$, at the cost of an increased rate in $N$, we have improved $C_{T}$'s rate in the dominating variable $T$ from $O(T)$ to $O(\log{T})$.
% \modelone{}'s other trade-off settings between $R_{T}$ and $C_{T}$ in Table \ref{tb:theoretical_comparison} are obtained using the same procedure.

% In particular, to obtain the results for \modelone{}, we first fix the upper bound for $C_{T}$, get the corresponding threshold $\gamma$, and then substitute the threshold $\gamma$ into the upper bound of $R_{T}$. For example, when setting the upper bound $C_{T} \leq N d \frac{\log{T}}{\log{\gamma}}$ to be $N^{\frac{3}{2}} d \log{T}$, we have $\frac{1}{\log{\gamma}}=N^{\frac{1}{2}}$, and thus $\gamma=\exp(N^{-\frac{1}{2}})$. Then by substituting $\gamma$ into the upper bound of $R_{T}$, we have:
% \small
% \begin{align*}
%     R_{T} & =O\left(\sqrt{(N-1)\gamma^{2}+(2-N)\gamma}d \sqrt{T} \log{T}\right)  = O\left(\sqrt{(N-1)e^{2 N^{-\frac{1}{2}}}+(2-N)e^{N^{-\frac{1}{2}}}}d \sqrt{T} \log{T}\right)  \\
% \end{align*}
% \normalsize
% Since $\lim_{N \rightarrow \infty} \frac{\sqrt{(N-1)e^{2 N^{-\frac{1}{2}}}+(2-N)e^{N^{-\frac{1}{2}}}}}{N^{\frac{1}{4}}}=1$, we know $\sqrt{(N-1)e^{2 N^{-\frac{1}{2}}}+(2-N)e^{N^{-\frac{1}{2}}}}=O(N^{\frac{1}{4}})$. Therefore, $R_{T}=O(N^{\frac{1}{4}}d \sqrt{T} \log{T})$. \modelone{}'s other trade-off settings between $R_{T}$ and $C_{T}$ in Table \ref{tb:theoretical_comparison} are obtained using the same procedure.

For comparison, we choose the threshold $D$ for \modelbaseline{} such that its upper bound of $R_{T}$ matches that of \modelone{}; and we include the corresponding results in Table \ref{tb:theoretical_comparison} as well.
We can see that \modelone{}'s upper bound of $C_{T}$ is not influenced by whether the client distribution is uniform or not, while \modelbaseline{} is, as we have shown in Section \ref{sec:sync_method}.
% In the ideal case where the client distribution is uniform, \modelbaseline{} is better than \modelone{} in the rate of $C_{T}$ by a factor of $O(N^{\frac{1}{4}})$.
Specifically, under the same regret $R_{T}=O(N^{\frac{1}{4}}d\sqrt{T}\log{T})$, in terms of $C_{T}$'s rate in $N$, \modelbaseline{} is slightly better than \modelone{} (by a factor of $O(N^{\frac{1}{4}})$) under the ideal case of uniform client distribution, and slightly worse than \modelone{} (by a factor of $O(N^{\frac{1}{4}})$) under non-uniform client distribution.

\section{Proof of Lemma \ref{lem:confidence_ellipsoid}}
Recall that the set of time steps corresponding to the observations used to compute $\{V_{i,t},b_{i,t}\}$ is denoted as $\cN^{(g)}_{i}(t)$.
By substituting $y_{\tau}={\bx_{\tau}^{(g)}}^{\top}\theta^{(g)}+{\bx_{\tau}^{(l)}}^{\top}\theta^{(i_{\tau})}+\eta_{\tau}$ into $\hat{\theta}^{(g)}_{i,t}(\lambda)=V_{i,t}(\lambda)^{-1}b_{i,t}$, for $\tau \in \cN^{(g)}_{i}(t)$, we get $\hat{\theta}^{(g)}_{i,t}(\lambda)=V_{i,t}(\lambda)^{-1} ( V_{i,t} \theta^{(g)} + \cS^{(g)}_{t} + \cE^{(g)}_{t})$, where $\cS^{(g)}_{t}=\sum_{\tau \in \cN^{(g)}_{i}(t)}\bx_{\tau}^{(g)}\eta_{\tau}$, $\cE^{(g)}_{t}=\sum_{\tau \in \cN^{(g)}_{i}(t)}\bx_{\tau}^{(g)}e^{(l)}_{\tau}$, and $e^{(l)}_{\tau}={\bx_{\tau}^{(l)}}^{\top}(\theta^{(i_{\tau})}-\hat{\theta}^{(l)}_{i_{\tau},t})$.
Therefore, we have:
\begin{equation*}
\begin{split}
     ||\hat{\theta}^{(g)}_{i,t}(\lambda) - \theta^{(g)}||_{V_{i,t}(\lambda)} & \leq ||V_{i,t}(\lambda)^{-1} ( V_{i,t} \theta^{(g)} + \cS^{(g)}_{t} + \cE^{(g)}_{t}) - \theta^{(g)}||_{V_{i,t}(\lambda)} \\
    & \leq ||\cS^{(g)}_{t}||_{V_{i,t}(\lambda)^{-1}} + ||\cE^{(g)}_{t}||_{V_{i,t}(\lambda)^{-1}} + \sqrt{\lambda} ||\theta^{(g)}||_{2}
\end{split}
\end{equation*}
where the third term $\sqrt{\lambda} ||\theta^{(g)}|| \leq \sqrt{\lambda}$.
To further bound the first two terms, we rely on the self-normalized bound in Theorem 1 of \cite{abbasi2011improved}, which we included in Lemma \ref{lem:self_normalized_bound} for the sake of completeness.

Since $\eta_{\tau}$ in $\cS^{(g)}_{t}$ is zero mean $\sigma$-sub-Gaussian conditioning on $\cF_{\tau-1}$, by Lemma \ref{lem:self_normalized_bound}, $||\cS^{(g)}_{t}||_{V_{i,t}(\lambda)^{-1}} \leq \sigma \sqrt{2\ln{\frac{\det{(V_{i,t}+\lambda I)^{1/2}}}{\det{(\lambda I)^{1/2}}\delta}}}$, with probability at least $1-\delta$. Now it remains to bound the term $||\cE^{(g)}_{t}||_{V_{i,t}(\lambda)^{-1}}$ that depends on $e^{(l)}_{\tau}={\bx_{\tau}^{(l)}}^{\top}(\theta^{(i_{\tau})}-\hat{\theta}^{(l)}_{i_{\tau},t})$, the estimation error of `partial' reward for $\theta^{l}$, due to the AM steps in Eq \eqref{eq:AM}.

In the following lemma, we show that $e^{(l)}_{\tau}$, for $\tau \in [t]$ is also zero mean conditionally sub-Gaussian, if the AM steps in Eq \eqref{eq:AM} is properly initialized, i.e., when executing Eq \eqref{eq:AM} for the first time, the initial value of $\hat{\theta}_{i_{t},t}^{(g)}$ is an unbiased estimator of $\theta^{(g)}$.
\begin{lemma}\label{lem:e_subgaussian}
When the AM steps in Eq \eqref{eq:AM} is properly initialized, $e^{(l)}_{t}$ is zero mean $2$-sub-Gaussian, $e^{(g)}_{t}$ is zero mean $2$-sub-Gaussian, conditioning on $\cF_{t-1}$, $\forall t$.
\end{lemma}

Then, similarly,  by Lemma \ref{lem:self_normalized_bound}, $||\cE^{(g)}_{t}||_{V_{i,t}(\lambda)^{-1}} \leq 2 \sqrt{2\ln{\frac{\det{(V_{i,t}+\lambda I)^{1/2}}}{\det{(\lambda I)^{1/2}}\delta}}}$, which shows that the errors caused by AM steps in Eq \eqref{eq:AM} only contribute a constant factor compared with the standard result. Putting everything together, we have $||\hat{\theta}^{(g)}_{i,t}(\lambda) - \theta^{(g)}||_{V_{i,t}(\lambda)} \leq (\sigma+2) \sqrt{2\ln{\frac{\det{(V_{i,t}(\lambda) )^{1/2}}}{\det{(\lambda I)^{1/2}}\delta}}}+\sqrt{\lambda}$. Following the same procedure, we can show that, $||\hat{\theta}^{(l)}_{i,t}(\lambda) - \theta^{(i)}||_{V^{(l)}_{i,t}(\lambda)} \leq (\sigma+2) \sqrt{2\ln{\frac{\det{(V^{(l)}_{i,t}(\lambda))^{1/2}}}{\det{(\lambda I)^{1/2}}\delta}}}+\sqrt{\lambda}$, with probability at least $1-\delta$.

\noindent \textit{Proof of Lemma \ref{lem:e_subgaussian}}

Recall that $e^{(l)}_{t}=(\theta^{(i_{t})}-\hat{\theta}^{(l)}_{i_{t},t})^{\top}{\bx_{t}^{(l)}}$ and $e^{(g)}_{t}=(\theta^{(g)}-\hat{\theta}^{(g)}_{i_{t},t})^{\top}{\bx_{t}^{(g)}}$. And the two estimators $\hat{\theta}^{(l)}_{i_{t},t}$ and $\hat{\theta}^{(g)}_{i_{t},t}$ are obtained from running the AM steps in Eq \eqref{eq:AM} on new data point $(\bx_{t},y_{t})$. When conditioning on $\cF_{t-1}=\{X_{1},Y_{1},\dots,X_{t-1},Y_{t-1},X_{t}\}$, $e^{(l)}_{t}$ and $e^{(g)}_{t}$ are random variables. In addition, they are bounded in $[ -2, 2 ]$ and $[-2, 2]$ respectively, because $|e^{(l)}_{t}| \leq ||\bx_{t}^{(l)}||_{2} \cdot ||\theta^{(i_{t})}-\hat{\theta}^{(l)}_{i_{t},t}||_{2} \leq 2 $ and $|e^{(g)}_{t}| \leq ||\bx_{t}^{(g)}||_{2} \cdot ||\theta^{(g)}-\hat{\theta}^{(g)}_{i_{t},t}||_{2} \leq 2 $. Therefore, by Lemma \ref{lem:bounded_rv}, $e^{(l)}_{t}$ is $2 $-sub-Gaussian, and $e^{(g)}_{t}$ is $2$-sub-Gaussian.

Now we look at the mean $\bE[e^{(l)}_{t}]={\bx_{t}^{(l)}}^{\top}(\theta^{(i_{t})}-\bE[\hat{\theta}^{(l)}_{i_{t},t}])$ and $\bE[e^{(g)}_{t}]={\bx_{t}^{(g)}}^{\top}(\theta^{(g)}-\bE[\hat{\theta}^{(g)}_{i_{t},t}])$. Note that $\bE[e^{(l)}_{t}]$ and $\bE[e^{(g)}_{t}]$ have an recursive dependence on each other as we iteratively update them using Eq \eqref{eq:AM}. For example,
% $\bE[e^{(l)}_{t}]= {\bx_{t}^{(l)}}^{\top}(\theta^{(i_{t})}-\bE[\hat{\theta}^{(l)}_{i_{t},t}])$ depends on
$\bE[\hat{\theta}^{(l)}_{i_{t},t}]=(\sum_{\tau\in\cN_{i_{t}}(t)}\bx_{\tau}^{(l)}{\bx_{\tau}^{(l)}}^{\top})^{-1}[\sum_{\tau\in \cN_{i_{t}}(t)}\bx_{\tau}^{(l)}({\bx_{\tau}^{(l)}}^{\top}\theta^{(i_{t})}+\eta_{\tau}+\bE[e^{(g)}_{\tau}])]$. In order to make $\bE[e^{(l)}_{t}]=0,\bE[e^{(g)}_{t}]=0,\forall t$, we need to initialize AM steps with an unbiased estimate of $\theta^{(g)}$, such that $\bE[e^{(l)}_{0}]=0$, and then all subsequent $e^{(l)}_{t},e^{(g)}_{t}$ will have zero mean.
% ${V}^{(l)}_{i,t}=\sum_{\tau\in\cN_{i}(t)}\bx_{\tau}^{(l)}{\bx_{\tau}^{(l)}}^{\top}$, ${b}^{(l)}_{i,t}=\sum_{\tau\in \cN_{i}(t)}\bx_{\tau}^{(l)}(y_{\tau}-{\bx_{\tau}^{(g)}}^{\top}\hat{\theta}_{t}^{(g)})$,

\begin{remark}
In order to simplify the description in Algorithm \ref{algo:AsyncLinUCB_AM}, we assumed such an unbiased estimate is readily available to initialize the AM steps. 
Note that an unbiased estimate of $\theta^{(g)}$ can be obtained by taking the global component of an MLE estimator of $\theta$, because $\bE[\hat{\theta}_{\text{MLE}}]=\begin{bmatrix}\bE[\hat{\theta}^{(g)}_{\text{MLE}}]\\\bE[\hat{\theta}^{(i)}_{\text{MLE}}]\end{bmatrix}=\begin{bmatrix} \theta^{(g)}\\ \theta^{(i)} \end{bmatrix}$.
If the learning system has access to a rank-sufficient dataset on any of the client before running Algorithm \ref{algo:AsyncLinUCB_AM}, then it can construct such a MLE estimator for initialization.
% This is a reasonable assumption in most cases, since it only requires the learning system to have access to a rank-sufficient dataset on any of the client before running Algorithm \ref{algo:AsyncLinUCB_AM}, such that it can construct a MLE estimator for initialization.
\end{remark}

However, for situations where this does not hold, i.e., the learning system does not have any history data before running Algorithm \ref{algo:AsyncLinUCB_AM}. 
We can slightly modify Algorithm \ref{algo:AsyncLinUCB_AM} as described in Algorithm \ref{algo:AsyncLinUCB_AM_v2}. Now, each client $i\in [N]$ will run standard LinUCB algorithm (line 9-10 in Algorithm \ref{algo:AsyncLinUCB_AM_v2}), until it collects enough data to construct an unbiased estimate of $\theta^{(g)}$ (line 12 in Algorithm \ref{algo:AsyncLinUCB_AM_v2}): either by using aggregated updates it has received from the server; or by collecting enough history data locally. 
Our Assumption \ref{assump:context_diversity} guarantees the clients are able to collect a rank-sufficient dataset locally, and how long it takes for the first client to do so is determined by the constant $\lambda_{c}$, i.e., the lower bound for the minimum eigenvalue of the covariance matrix $\Sigma_{c}$.
Then after using the unbiased estimate of $\theta^{(g)}$ to initialize AM steps (line 13 in Algorithm \ref{algo:AsyncLinUCB_AM_v2}), which we mark as $\text{State}(i)=1$, the client will proceed with the same steps as in Algorithm \ref{algo:AsyncLinUCB_AM} (line 18-21 in Algorithm \ref{algo:AsyncLinUCB_AM_v2}).

\begin{algorithm}[h]
    \caption{Asynchronous LinUCB Algorithm with Alternating Minimization} \label{algo:AsyncLinUCB_AM_v2}
  \begin{algorithmic}[1]
    \STATE \textbf{Input:} thresholds $\gamma_{U}, \gamma_{D} \geq 1$, $d_{g},d_{i}$ for $i\in [N]$, $\sigma, \lambda > 0$, $\delta \in (0,1)$
    \STATE Initialize server: ${V}_{g, 0}=\textbf{0}_{d_{g} \times d_{g}}$, ${b}_{g,0}=\textbf{0}_{d_{g}}$
    \FOR{ $t=1,2,...,T$}
        \STATE Observe arm set $\mathcal{A}_{t}$ for client $i_{t} \in [N]$
        \IF{Client $i_{t}$ is new}
            \STATE Initialize client $i_{t}$: $\text{State}(i_{t})=0$, $\cI_{i_{t},t-1}=\emptyset$, ${V}_{i_{t}, t-1}=\textbf{0}_{d_{g} \times d_{g}}$, ${b}_{i_{t},t-1}=\textbf{0}_{d_{g}}$
            \STATE Initialize server's download buffer for client $i_{t}$: $\Delta{V}_{-i_{t}, t-1}=V_{g,t-1}$, $\Delta{b}_{-i_{t},t-1}=b_{g,t-1}$
        \ENDIF
        \IF{$\text{State}(i_{t})=0$}
            \STATE Select arm $\bx_{t}\in\cA_{t}$ by Eq \eqref{eq:UCB} and observe $y_{t}$
            \STATE $\cI_{i_{t},t}=\cI_{i_{t},t-1} \cup (t)$
            \IF{$\sum_{\tau \in \cI_{i_{t},t}}\bx_{\tau} \bx_{\tau}^{\top}$ or $V_{i_{t},t-1}$ is full rank}
                \STATE Initialize AM on local data $\{(\bx_{\tau},y_{\tau})\}_{\tau \in \cI_{i_{t},t}}$ to get the estimated partial reward vectors $\hat{\by}^{(g)},\hat{\by}^{(l)}$
                \STATE Update client $i_{t}$:
                % $\bX^{(g)}=\sum_{\tau \in \cI_{i_{t},t}}\bx_{\tau}^{(g)} {\bx_{\tau}^{(g)}}^{\top}$, $\bX^{(l)}=\sum_{\tau \in \cI_{i_{t},t}}\bx_{\tau}^{(l)} {\bx_{\tau}^{(l)}}^{\top}$ \newline
                ${V}_{i_{t}, t} \mathrel{+}= {\bX^{(g)}}^{\top}\bX^{(g)} $, ${b}_{i_{t},t} \mathrel{+}= {\bX^{g}}^{\top} \hat{\by}^{(g)}$, $\Delta{V}_{i_{t}, t-1}={\bX^{(g)}}^{\top}\bX^{(g)}$, $\Delta{b}_{i_{t},t-1}={\bX^{g}}^{\top} \hat{\by}^{(g)}$, ${V}^{(l)}_{i_{t},t-1}={\bX^{(l)}}^{\top}\bX^{(l)} $, ${b}^{(l)}_{i_{t},t-1}={\bX^{l}}^{\top} \hat{\by}^{(l)}$
                % \STATE Run AM on $\{(\bx_{\tau},y_{\tau})\}_{\tau \in \cI_{i_{t},t}}$ to get estimated partial rewards $\{(\hat{y}_{\tau}^{(g)},\hat{y}_{\tau}^{(l)})\}_{\tau \in \cI_{i_{t},t}}$
                % \STATE Update client $i_{t}$: \newline
                % \hspace*{1em} ${V}_{i_{t}, t} \mathrel{+}= \sum_{\tau \in \cI_{i_{t},t}}\bx_{\tau}^{(g)} {\bx_{\tau}^{(g)}}^{\top}$, ${b}_{i_{t},t} \mathrel{+}= \sum_{\tau \in \cI_{i_{t},t}}\bx_{\tau}^{(g)} \hat{y}_{\tau}^{(g)}$ \newline
                % \hspace*{1em} $\Delta{V}_{i_{t}, t-1}= \sum_{\tau \in \cI_{i_{t},t}}\bx_{\tau}^{(g)} {\bx_{\tau}^{(g)}}^{\top}$, $\Delta{b}_{i_{t},t-1}=\sum_{\tau \in \cI_{i_{t},t}}\bx_{\tau}^{(g)} \hat{y}_{\tau}^{(g)}$ \newline
                % \hspace*{1em} ${V}^{(l)}_{i_{t},t-1}=\sum_{\tau \in \cI_{i_{t},t}}\bx_{\tau}^{(l)} {\bx_{\tau}^{(l)}}^{\top}$, ${b}^{(l)}_{i_{t},t-1}=\sum_{\tau \in \cI_{i_{t},t}}\bx_{\tau}^{(l)} \hat{y}_{\tau}^{(l)}$
                \STATE Set $\text{State}(i_{t})=1$
            \ENDIF
        \ELSE
            \STATE Select arm $\bx_{t}\in\cA_{t}$ by Eq \eqref{eq:UCB_AM} and observe $y_{t}$
            \STATE Run AM by Eq \eqref{eq:AM} to get the estimated partial rewards: $\hat{y}^{(g)}_{t}=y_{t}-{\bx_{t}^{(l)}}^{\top} \hat{\theta}^{(l)}_{i_{t},t}$, $\hat{y}^{(l)}_{t}=y_{t}-{\bx_{t}^{(g)}}^{\top} \hat{\theta}^{(g)}_{i_{t},t}$
            \STATE Update client $i_{t}$: ${V}_{i_{t},t} \mathrel{+}= \bx_{t}^{(g)}{\bx_{t}^{(g)}}^{\top}$, ${b}_{i_{t},t} \mathrel{+}= \bx^{(g)}_{t} \hat{y}^{(g)}_{t}$, $\Delta{V}_{i_{t},t} \mathrel{+}= \bx_{t}^{(g)}{\bx_{t}^{(g)}}^{\top}$, $\Delta{b}_{i_{t},t}+=\bx^{(g)}_{t} \hat{y}^{(g)}_{t}$, ${V}^{(l)}_{i_{t},t} \mathrel{+}= \bx_{t}^{(l)}{\bx_{t}^{(l)}}^{\top}$, ${b}^{(l)}_{i_{t},t} \mathrel{+}= \bx^{(l)}_{t} \hat{y}^{(l)}_{t}$ 
        \ENDIF
        \STATE Event-triggered Communications (Algorithm \ref{algo:comm})
        % \IF{ Event $\cU_{t}(\gamma_{U})$ (Eq \eqref{eq:upload}) happens}
        %     \STATE Upload $\Delta{V}_{i_{t},t},\Delta{b}_{i_{t},t}$ ($i_{t} \rightarrow \text{server}$) 
        %     \STATE Update server: \newline
        %     % $V_{g,t}=V_{g,t-1}+\Delta V_{i_{t},t},b_{g,t}=b_{g,t-1}+\Delta b_{i_{t},t}$ \newline
        %     % $\Delta V_{-j,t}=\Delta V_{-j,t-1}+\Delta V_{i_{t},t},\Delta b_{-j,t}=\Delta b_{-j,t-1}+\Delta b_{i_{t},t}$ for $j \neq i_{t}$
        %     \hspace*{1em} $V_{g,t}\mathrel{+}=\Delta V_{i_{t},t},b_{g,t}\mathrel{+}=\Delta b_{i_{t},t}$ \newline
        %     \hspace*{1em} $\Delta V_{-j,t}\mathrel{+}=\Delta V_{i_{t},t},\Delta b_{-j,t}\mathrel{+}=\Delta b_{i_{t},t}$, $j \neq i_{t}$
        %     \STATE Client $i_{t}$ reset $\Delta{V}_{i_{t},t}=\textbf{0}$, $\Delta{b}_{i_{t},t}=\textbf{0}$

        %     \FOR{$j = 1, \dots, N$ and $j \neq i_{t}$}
        %         \IF{ Event $\cD_{t,j}(\gamma_{D})$ (Eq \eqref{eq:download}) happens}
        %             \STATE Download $\Delta{V}_{-j,t},\Delta{b}_{-j,t}$ ($\text{server} \rightarrow j$)
        %             \STATE Update client $j$: \newline
        %             % $V_{j,t}=V_{j,t-1}+\Delta V_{-j,t},b_{j,t}=b_{j,t-1}+\Delta b_{-j,t}$
        %             \hspace*{1em} $V_{j,t}\mathrel{+}=\Delta V_{-j,t},b_{j,t}\mathrel{+}=\Delta b_{-j,t}$
        %             \STATE Server reset $\Delta{V}_{-j,t}=\textbf{0},\Delta{b}_{-j,t}=\textbf{0}$
        %         \ENDIF
        %     \ENDFOR
        % \ENDIF
    \ENDFOR
  \end{algorithmic}
\end{algorithm}

\section{Proof of Theorem \ref{thm:regret_communication_async-linucb-am} (Regret and Communication Upper Bound for \modeltwo{})}
\noindent \textbf{Regret and communication cost}:
The instantaneous regret $r_{t}$ can be upper bounded w.r.t. the confidence bounds for global component and local component: 
% $ r_{t}  = \theta^\top \bx^{*} - \theta^\top \bx_{t}  \leq \tilde{\theta}^\top_{t-1} \bx_{t} - \theta^\top \bx_{t}  = (\tilde{\theta}_{t-1}-\theta)^\top \bx_{t}  = (\tilde{\theta}^{(g)}_{t-1}-\theta^{(g)})^\top \bx_{t}^{(g)} + (\tilde{\theta}^{(i_{t})}_{t-1}-\theta^{(i_{t})})^\top \bx_{t}^{(l)}  \leq 2 \text{CB}^{(g)}_{i_{t},t-1}(\bx_{t}^{(g)}) + 2 \text{CB}^{(l)}_{i_{t},t-1}(\bx_{t}^{(l)})$,
\begin{align*}
    & r_{t}  = \theta^\top \bx^{*} - \theta^\top \bx_{t}  \leq \tilde{\theta}^\top_{t-1} \bx_{t} - \theta^\top \bx_{t} = (\tilde{\theta}_{t-1}-\theta)^\top \bx_{t} \\
    & = (\tilde{\theta}^{(g)}_{t-1}-\theta^{(g)})^\top \bx_{t}^{(g)} + (\tilde{\theta}^{(i_{t})}_{t-1}-\theta^{(i_{t})})^\top \bx_{t}^{(l)} \\
    & \leq 2 \text{CB}^{(g)}_{i_{t},t-1}(\bx_{t}^{(g)}) + 2 \text{CB}^{(l)}_{i_{t},t-1}(\bx_{t}^{(l)})
\end{align*}
% \begin{align*}
%     r_{t} & = \theta^\top \bx^{*} - \theta^\top \bx_{t}  \leq \tilde{\theta}^\top_{t-1} \bx_{t} - \theta^\top \bx_{t}  = (\tilde{\theta}_{t-1}-\theta)^\top \bx_{t}  \\
%     & = (\tilde{\theta}^{(g)}_{t-1}-\theta^{(g)})^\top \bx_{t}^{(g)} + (\tilde{\theta}^{(i_{t})}_{t-1}-\theta^{(i_{t})})^\top \bx_{t}^{(l)}  \leq 2 \text{CB}^{(g)}_{i_{t},t-1}(\bx_{t}^{(g)}) + 2 \text{CB}^{(l)}_{i_{t},t-1}(\bx_{t}^{(l)})
% \end{align*}
where $\tilde{\theta}_{t-1}$ denotes the optimistic estimate used in UCB strategy, and $\{\tilde{\theta}^{(i_{t})}_{t-1},\tilde{\theta}^{(g)}_{t-1}\}$ denote its global and local components, respectively.
Then the accumulative regret can be upper bounded by: 
% $R_{T} \leq 2 \sum_{t=1}^{T}\text{CB}^{(g)}_{i_{t},t-1}(\bx_{t}^{(g)}) + 2 \sum_{t=1}^{T} \text{CB}^{(l)}_{i_{t},t-1}(\bx_{t}^{(l)}) \leq 2 \sum_{t=1}^{T}\text{CB}^{(g)}_{i_{t},t-1}(\bx_{t}^{(g)}) + 2 \sum_{i=1}^{N} \sum_{t \in \cN_{i}(T)} \text{CB}^{(l)}_{i,t-1}(\bx_{t}^{(l)}) = O\left(d_{g}\sqrt{T\log^{2}{T}})\min(\sqrt{N},\sqrt{\gamma_{D} [1+(N-1)(\gamma_{U}-1)]}\right) \quad + \sum_{i=1}^{N} O\left(d_{l}\sqrt{|\cN_{i}(T)|\log^{2}{|\cN_{i}(T)|}}\right)$,
% \small
\begin{align*}
    R_{T} & \leq 2 \sum_{t=1}^{T}\text{CB}^{(g)}_{i_{t},t-1}(\bx_{t}^{(g)}) + 2 \sum_{t=1}^{T} \text{CB}^{(l)}_{i_{t},t-1}(\bx_{t}^{(l)}) \\
    & \leq 2 \sum_{t=1}^{T}\text{CB}^{(g)}_{i_{t},t-1}(\bx_{t}^{(g)}) + 2 \sum_{i=1}^{N} \sum_{t \in \cN_{i}(T)} \text{CB}^{(l)}_{i,t-1}(\bx_{t}^{(l)}) \\
    & = O\bigl(d_{g}\sqrt{T\log^{2}{T}}\min(\sqrt{N},\sqrt{\gamma_{D} [1+(N-1)(\gamma_{U}-1)]} + \sum_{i=1}^{N} d_{i} \sqrt{|\cN_{i}(T)|\log^{2}{|\cN_{i}(T)|}}\bigr)
\end{align*}
% \normalsize
where the first term is upper bounded following the same procedure as in Section \ref{subsec:async_LinUCB}, and the second is essentially the regret upper bound for running $N$ independent LinUCB algorithms in each client $i$ for $\theta^{(i)}$. Intuitively, when the problems solved by different clients become more similar, the first term dominates as $d_{g}$ becomes larger compared with $d_{i}$.

In addition, as the clients only communicate sufficient statistics for $\theta^{(g)}$, following the same steps for upper bounding the communication cost in Section \ref{subsec:async_LinUCB}, we can show that the communication cost for \modeltwo{} is $C_{T}=O(d_{g} N\log{T}/\log{\min(\gamma_{U},\gamma_{D})})$.

% \begin{align*}
%     C_{T} &=\sum_{i=1}^{N} C_{T,i} \leq N \frac{\log \det({V}_{T-1})-d_{g} \log \lambda}{\log \min(\gamma_{U},\gamma_{D})}\\
%     &  = O(N d_{g} \log{T})
% \end{align*}

\section{Addition Experiments on Delicious Dataset}  \label{sec:additional_exp}
In Figure \ref{fig:e}, the blue stars illustrate the results of \modelone{} with its threshold $\gamma$ ranging from $9001$ to $1.01$. And within this range, we observe that the reward decreases from $1.8322$ to $1.2887$ as the communication increases from $14599$ to $12901135$. However, we can see from the figure that when $\gamma$ is set in the interval between $\infty$ and $9001$, the reward seems to increase as communication increases, which implies a changing point for the relationship between communication and reward on this dataset. To validate this, we have run some additional experiments on \modelone{} with $\gamma$ ranging from $10^{4}$ to $10^{20}$. We observed that, in this low-communication region (e.g. with $\gamma>10^{4}$), the reward indeed increases when communication increases. Specifically, the reward increased from $1.6891$ to $1.8348$, as the communication increased from $0$ to $14230$.
 
Our hypothesis for this observation is: as the threshold is high in the low-communication region, only the most active users are able to contribute to global data sharing. The observation that this boosts the overall performance indicates that, as the less active clients download this data, the benefit from reduced variance outweighs the harm caused by the increased bias (due to user heterogeneity). However, with the threshold further reduced, many more clients are able to contribute to global data sharing, such that the global data would become so heterogeneous that it starts to hurt the overall performance. 
 
To verify this hypothesis, we split all the users into ten groups based on their number of interactions, and then include the following statistics about the results for \modelone{} with $\gamma=+\infty$ ($C_{T}=0$), $\gamma=6001$ ($C_{T}=14230$), and $\gamma=3$ ($C_{T}=54006$), respectively in Table \ref{tb:additional_exp} below. 
\begin{table}[ht]
\centering
\caption{Statistics about experiment results split into ten groups.}
% \begin{tabular}{@{}p{0.08\textwidth}*{12}{L{\dimexpr0.22\textwidth-2\tabcolsep\relax}}@{}}
\begin{tabular}{p{1cm} p{1.2cm} p{1.3cm} p{1.6cm} p{1cm} p{1.6cm} p{1cm} p{1.6cm} p{1cm}}
\toprule
% & \multicolumn{6}{c}{Environment settings} &
% \multicolumn{6}{c}{Accumulated regret} \\
% \cmidrule(r{4pt}){2-7} \cmidrule(l){8-13}
 &  &  & \multicolumn{2}{c}{$C_T=0$} & \multicolumn{2}{c}{$C_T=14230$} & \multicolumn{2}{c}{$C_T=54006$} \\
group & user no. & data no. & upload no. & reward & upload no. & reward & upload no. & reward \\
\midrule
% \cmidrule(r{4pt}){2-7} \cmidrule(l){8-14}
0-10 & 161 & 680 & 0 & 0 & 118 & 21 & 24 & 22\\
10-20 & 102 & 1514 & 0 & 1 & 75 & 123 & 117 & 61 \\
20-30 & 103 & 2575 & 0 & 0 & 79 & 214 & 202 & 103\\
30-40 & 114 & 3988 & 0 & 0 & 92 & 290 & 308 & 124\\
40-50 & 185 & 8253 & 0 & 0 & 144 & 704 & 764 & 298\\
50-60 & 243 & 13330 & 0 & 0 & 185 & 680 & 751 & 424\\
60-70 & 352 & 22693 & 0 & 1 & 280 & 1419 & 1576 & 978\\
70-80 & 336 & 24972 & 0 & 1 & 252 & 1628 & 1701 & 1351\\
80-90 & 213 & 17785 & 0 & 6 & 167 & 1185 & 1369 & 1341\\
90-100 & 38 & 3456 & 0 & 2 & 31 & 414 & 448 & 389 \\
sum & 1847 & 99246 & 0 & 11 & 1423 & 6678 & 7251 & 5091 \\
\bottomrule
\label{tb:additional_exp}
\end{tabular}
\end{table}

\noindent Note that abbreviations used in the table header means:
\begin{itemize}[noitemsep]
    \item User No: number of users in each group
    \item Data No. total number of data points users in each group have
    \item Upload No: total number of uploads that users in each group have triggered
    \item Reward: cumulative reward obtained by users in each group
\end{itemize}
First, we can see that, with $\gamma=6001$ ($C_{T}=14230$), most upload came from the active users, and with $\gamma=3.0$ ($C_{T}=54006$), every group contributed considerable amount to the upload. Second, when $C_{T}=14230$, almost all the groups have improved performance, which suggests that data uploaded by users in groups 80-100 can help improve the performance of other users. However, when $C_{T}$ increased to $54006$, the cumulative reward for the less active groups (10-80) dropped dramatically, while that for the most active groups (80-100) received much less negative impact. 

\begin{figure}[ht]
    \centering
    \includegraphics[width=0.75\textwidth]{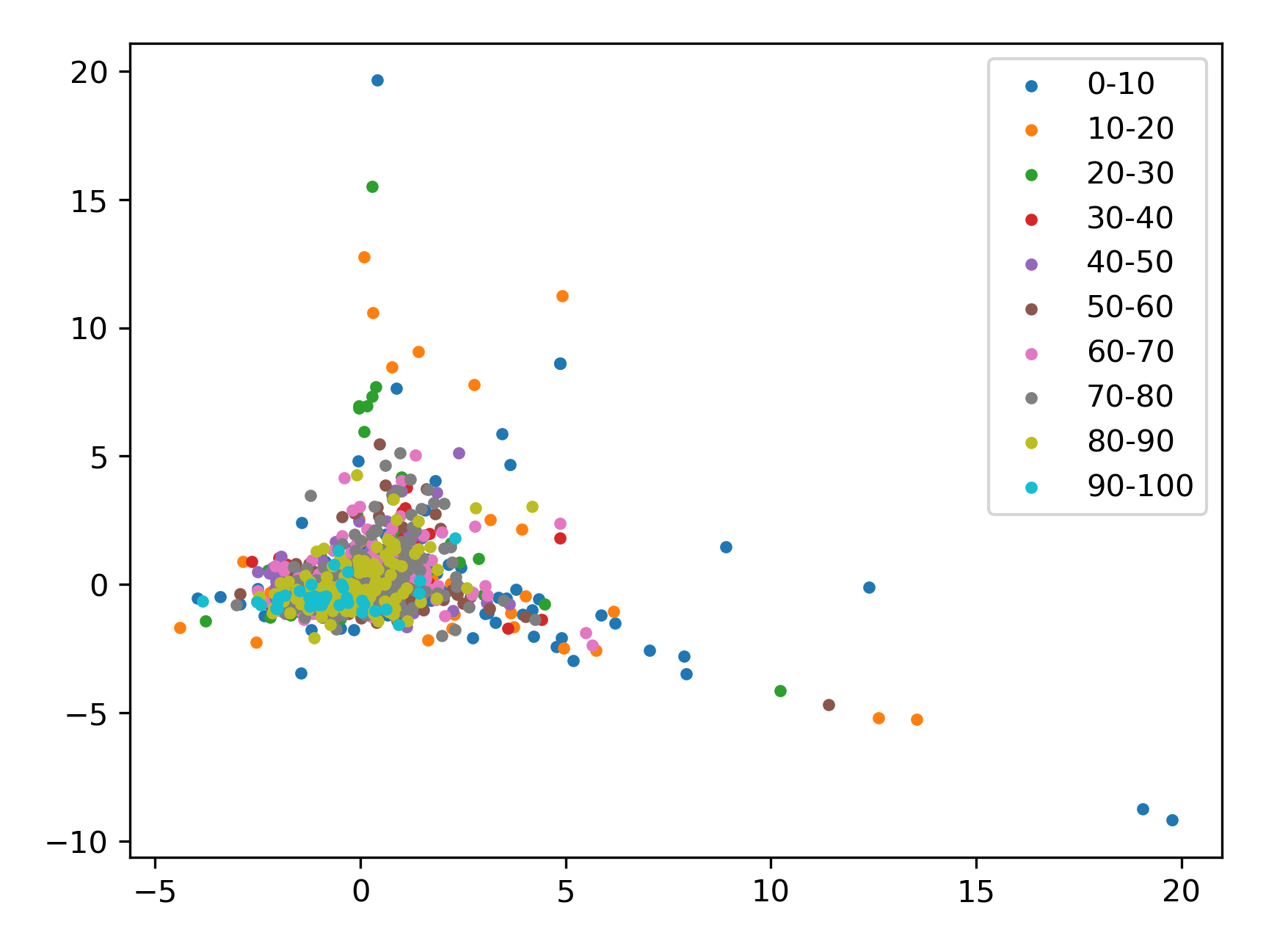}
    \caption{Visualization of user embedding vector}
    \label{fig:user_embedding_visualization}
\end{figure}

We further investigate the reason behind the observation above by visualizing the relationship among the individual users. Specifically, we use the average feature vector over all the positive items in a user as this user’s embedding vector. Then we use PCA to reduce its dimension from 25 to 2 to plot in a 2-D space, with each point labeled with the user’s group ID. The plot is shown in Figure \ref{fig:user_embedding_visualization}.

We can see that points corresponding to the most active groups (80-100) are centered near the origin, while points for the less active groups are distributed along two nearly orthogonal directions. This provides an intuitive explanation for our observations: data from groups 80-100 can boost the overall performance of most users because they roughly lie in the center of most points; but aggregating data across the less active users (0-80) degrades their own performance, because such users are extremely heterogeneous and distinct from each other.

\end{document}